\documentclass{article} 

\usepackage{comment}
\usepackage[utf8]{inputenc} 
\usepackage[T1]{fontenc}    
\usepackage{hyperref}       
\usepackage{url}            
\usepackage{booktabs}       
\usepackage{amsfonts}       
\usepackage{nicefrac}       
\usepackage{microtype}      
\usepackage[dvipsnames]{xcolor}
\usepackage{graphicx}
\usepackage{amsmath}
\usepackage{amsthm}
\usepackage{color}
\usepackage{wrapfig}
\usepackage{subcaption}
\usepackage{booktabs}
\usepackage{multirow}
\usepackage{floatrow}
\usepackage{verbatim}
\usepackage{sidecap}
\usepackage{caption}
\usepackage{blindtext}
\usepackage{subfiles}
\usepackage{xcolor}

\usepackage{iclr2021_conference,times}

\usepackage{amsmath,amsfonts,bm}

\def\eqref#1{equation~\ref{#1}}

\def\1{\bm{1}}

\DeclareMathAlphabet{\mathsfit}{\encodingdefault}{\sfdefault}{m}{sl}
\SetMathAlphabet{\mathsfit}{bold}{\encodingdefault}{\sfdefault}{bx}{n}

\newfloatcommand{capbtabbox}{table}[\captop][0.49\textwidth]

\title{Continual Learning in Recurrent Neural \\ Networks}

\author{Benjamin Ehret*, Christian Henning*, Maria R. Cervera*, Alexander Meulemans, \\
\textbf{Johannes von Oswald, Benjamin F.~Grewe} \\
*Equal contribution \\\\
  Institute of Neuroinformatics\\
  University of Zürich and ETH Zürich\\
  Zürich, Switzerland\\
  \texttt{\{behret,henningc,mariacer,ameulema,voswaldj,bgrewe\}@ethz.ch}
}

\iclrfinalcopy 
\begin{document}

\maketitle

\begin{abstract}
While a diverse collection of continual learning (CL) methods has been proposed to prevent catastrophic forgetting, a thorough investigation of their effectiveness for processing sequential data with recurrent neural networks (RNNs) is lacking. Here, we provide the first comprehensive evaluation of established CL methods on a variety of sequential data benchmarks. Specifically, we shed light on the particularities that arise when applying weight-importance methods, such as elastic weight consolidation, to RNNs. In contrast to feedforward networks, RNNs iteratively reuse a shared set of weights and require working memory to process input samples. We show that the performance of weight-importance methods is not directly affected by the length of the processed sequences, but rather by high working memory requirements, which lead to an increased need for stability at the cost of decreased plasticity for learning subsequent tasks. We additionally provide theoretical arguments supporting this interpretation by studying linear RNNs. Our study shows that established CL methods can be successfully ported to the recurrent case, and that a recent regularization approach based on hypernetworks outperforms weight-importance methods, thus emerging as a promising candidate for CL in RNNs. Overall, we provide insights on the differences between CL in feedforward networks and RNNs, while guiding towards effective solutions to tackle CL on sequential data.
\end{abstract}

\section{Introduction}
\label{sec:intro}

\vspace{-2mm}
The ability to continually learn from a non-stationary data distribution while transferring and protecting past knowledge is known as continual learning (CL).
This ability requires neural networks to be stable to prevent forgetting, but also plastic to learn novel information, which is referred to as the stability-plasticity dilemma \citep{Grossberg2007Nov, Mermillod2013Aug}. To address this dilemma, a variety of methods which tackle CL for static data with feedforward networks have been proposed (for reviews refer to \citet{Parisi2019May} and \citet{vandeVen2019Apr}).
However, CL for sequential data has only received little attention, despite recent work confirming that recurrent neural networks (RNNs) also suffer from catastrophic forgetting \citep{Schak2019Sep}.

A set of methods that holds great promise to address this problem are regularization methods, which work by constraining the update of certain parameters.
These methods can be considered more versatile than competing approaches, since they do not require rehearsal of past data, nor an increase in model capacity, but can benefit from either of the two \citep[e.g.,][]{nguyen:2017:vcl, yoon2018lifelong}. This makes regularization methods applicable to a broader variety of situations, e.g. when issues related to data privacy, storage, or limited computational resources during inference might arise.
The most well-known regularization methods are \textit{weight-importance methods}, such as elastic weight consolidation (EWC, \citet{kirkpatrick:ewc:2017}) and synaptic intelligence (SI, \citet{zenke:synaptic:intelligence}), which are based on assigning importance values to weights. 
Some of these have a direct probabilistic interpretation as prior-focused CL methods \citep{gal:bcl}, for which solutions of upcoming tasks must lie in the posterior parameter distribution of the current task (cf. Fig. \ref{fig:prior:focused:vs:hnet}), highlighting the stability-plasticity dilemma.
Whether this dilemma differently affects feedforward networks and RNNs, and whether weight-importance based methods can be used off the shelf for sequential data has remained unclear.

Here, we contribute to the development of CL approaches for sequential data in several ways.
\begin{itemize}
    \item We provide a first comprehensive comparison of CL methods applied to sequential data. For this, we port a set of established CL methods for feedforward networks to RNNs and assess their performance thoroughly and fairly in a variety of settings.
    \item We identify elements that critically affect the stability-plasticity dilemma of weight-importance methods in RNNs. 
    We empirically show that high requirements for working memory, i.e. the need to store and manipulate information when processing individual samples, lead to a saturation of weight importance values, making the RNN rigid and hindering its potential to learn new tasks. In contrast, this trade-off is not directly affected by the sheer recurrent reuse of the weights, related to the length of processed sequences.
    We complement these observations with a theoretical analysis of linear RNNs.
    \item We show that existing CL approaches can constitute strong baselines when compared in a standardized setting and if equivalent hyperparameter-optimization resources are granted. Moreover, we show that a CL regularization approach based on hypernetworks \citep{oswald:hypercl} mitigates the limitations of weight-importance methods in RNNs. 
    \item We provide a code base\footnote{Source code for all experiments (including all baselines) is available at \url{https://github.com/mariacer/cl_in_rnns}.}
    comprising all assessed methods as well as variants of four well known sequential datasets adapted to CL: the Copy Task \citep{graves2014ntm}, Sequential Stroke MNIST \citep{gulcehre2017memory}, AudioSet \citep{audioset} and multilingual Part-of-Speech tagging \citep{NIVRE16.348}. 
\end{itemize}

Taken together, our experimental and theoretical results facilitate the development of CL methods that are suited for sequential data.


\section{Related work}
\label{sec:related:work}


\vspace{-2mm}
\paragraph{Continual learning with sequential data.}
As in \citet{Parisi2019May}, we categorize CL methods for RNNs into regularization approaches, dynamic architectures and complementary memory systems.

Regularization approaches set optimization constraints on the update of certain network parameters without requiring a model of past input data.
EWC, for example, uses weight importance values to limit further updates of weights that are considered essential for solving previous tasks \citep{Kirkpatrick2017Mar}.
Throughout this work, we utilize a more mathematically sound and less memory-intensive version of this algorithm, called Online EWC \citep{huszar:ewc:note:2018, schwarz:online:ewc}.
Although a highly popular approach in feedforward networks, it has remained unclear how suitable EWC is in the context of sequential processing. Indeed, some studies report promising results in the context of natural language processing (NLP) \citep{madasu2020sequential, Thompson2019Jun}, while others find that it performs poorly \citep{asghar2018progressive, Cossu2020Apr, Li2020Compositional:Language:CL}. Here, we conduct the first thorough investigation of EWC's performance on RNNs, and find that it can often be a suitable choice.
A related CL approach that also relies on weight importance values is SI \citep{zenke:synaptic:intelligence}. Variants of SI have been used for different sequential datasets, but have not been systematically compared against other established methods \citep{yang2019task,masse:masking:pnas,Lee2017Dec}. 
Fixed expansion layers \citep{coop2012mitigation} are another method to limit the plasticity of weights and prevent forgetting, and in RNNs take the form of a sparsely activated layer between consecutive hidden states \citep{coop2013mitigation}.
Lastly, some regularization approaches rely on the use of non-overlapping and orthogonal representations to overcome catastrophic forgetting \citep{French1992Jan, French1994Aug, French1970Feb}. \citet{masse:masking:pnas}, for example, proposed the use of context-dependent random subnetworks, where weight changes are regularized by limiting plasticity to task-specific subnetworks. This eliminates forgetting for disjoint networks but leads to a reduction of available capacity per task.
In concurrent work, \citet{duncker2020organizing} introduced a learning rule which aims to optimize the use of the activity-defined subspace in RNNs learning multiple tasks. When tasks are different, catastrophic interference is avoided by forcing the use of task-specific orthogonal subspaces, whereas the reuse of dynamics is encouraged across tasks that are similar.

Dynamic architecture approaches,
which rely on the addition of neural resources to mitigate catastrophic forgetting, have also been applied to RNNs.
\citet{Cossu2020Apr} presented a combination of progressive networks \citep{rusu:progressive} and gating autoencoders \citep{aljundi2017expert}, where an RNN module is added for each new task and the reconstruction error of task-specific autoencoders is used to infer the RNN module to be used.
Arguably, the main limitation of this type of approach is the increase in the number of parameters with the number of tasks, although methods have been presented that add resources for each new task only if needed \citep{Tsuda2020Mar}.

Finally, complementary memory systems have also been applied to the retention of sequential information.
In an early work, \citet{Ans2002} proposed a secondary network that generates patterns for rehearsing previously learned information. \citet{asghar2018progressive} suggested using an external memory that is progressively increased when new information is encountered. 
\citet{sodhani2018training} combined an external memory with Net2Net \citep{Chen2015Nov}, such that the network capacity can be extended while maintaining memories. The major drawback of complementary memory systems is that they either violate CL desiderata by storing past data, or rely on the ability to learn a generative model, a task that arguably scales poorly to complex data. 
We discuss related work in a broader context in supplementary materials (SM \ref{supp:related:work}).

\vspace{-2mm}
\paragraph{Hypernetworks.}

Introduced by \citet{Ha2016Sep}, the term \textit{hypernetwork} refers to a neural network that generates the weights of another network. The idea can be traced back to \citet{schmidhuber:fast:weight:memories}, who already suggested that a recurrent hypernetwork could be used for learning to learn \citep{schmidhuber:self:referential}. 
Importantly, hypernetworks can make use of the fact that parameters in a neural network possess compressible structure \citep{denil:predicting:parameters, han2015learning}. Indeed, \citet{Ha2016Sep} showed that the number of trainable weights of feed-forward architectures can be reduced via hypernetworks. 
More recently, hypernetworks have been adapted for CL \citep{he:2019:task:agnostic:cl, oswald:hypercl}, but not for learning with sequential data.


\section{Methods}
\label{sec:methods}

\vspace{-2mm}
\paragraph{Recurrent Neural Networks.} We consider discrete-time RNNs. At timestep $t$, the network's output $\mathbf{\hat{y}}_t$ and hidden state $\mathbf{h}_t$ are given by $(\mathbf{\hat{y}}_t, \mathbf{h}_t) = f_\text{step}(\mathbf{x}_t, \mathbf{h}_{t-1}, \psi)$, where $\mathbf{x}_t$ denotes the input at time $t$ and $\psi$ the parameters of the network \citep{cho:gated:recurrent:unit, elman:network, hochreiter1997long}. In this work, we consider either vanilla RNNs (based on \citet{elman:network}), LSTMs \citep{hochreiter1997long} or BiLSTMs \citep{schuster97}.

\vspace{-2mm}
\paragraph{Naive baselines.} We consider the following naive baselines. \textbf{Fine-tuning} refers to training an RNN sequentially on all tasks without any CL protection. Each task has a different output head (multi-head), and the heads of previously learned tasks are kept fixed. \textbf{Multitask} describes the parallel training on all tasks (no CL). To keep approaches comparable, the multitask baseline uses a multi-head output. Because we focus on methods with a comparable number of parameters, we summarize approaches that allocate a different model per task in the \textbf{From-scratch} baseline, where a different model is trained separately for each task, noting that performance improvements are likely to arise in related methods (such as \citet{Cossu2020Apr}) whenever knowledge transfer is possible.

\vspace{-2mm}
\paragraph{Continual learning baselines.} We consider a diverse set of established CL methods and investigate their performance in RNNs. \textbf{Online EWC} \citep{huszar:ewc:note:2018, kirkpatrick:ewc:2017, schwarz:online:ewc} and \textbf{SI} \citep{zenke:synaptic:intelligence} are different weight-importance CL methods. A simple weighted L2 regularization ensures that the neural network is more rigid in weight directions that are considered \textit{important} for previous tasks, i.e., the loss for the $K$-th task is given by

\vspace{-3mm}
\begin{equation}
    \label{eq:weight:importance:reg}
    \mathcal{L}(\psi, \mathcal{D}_K) = \mathcal{L}_\text{task}(\psi, \mathcal{D}_K) + \lambda \sum_{i=1}^{\lvert \psi \rvert} \omega_i (\psi_i - \tilde{\psi}_i^{(K-1)} )^2
\end{equation}

\begin{figure}[t]
    \centering
    \begin{subfigure}[b]{1.9in}
        \includegraphics[width=\textwidth]{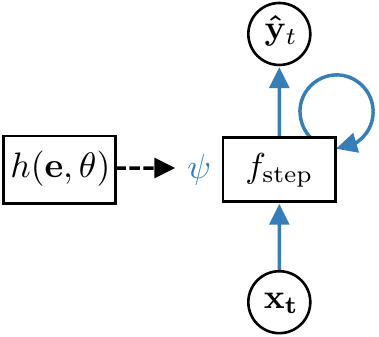}
        \caption{}
        \label{fig:hnet:schematic}
    \end{subfigure}
    \hfill
    \begin{subfigure}[b]{3.2in}
        \includegraphics[width=\textwidth]{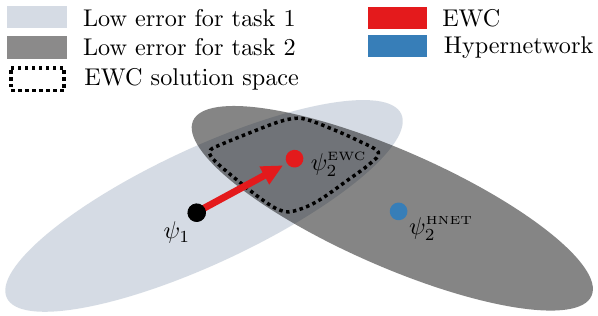}
        \caption{}
        \label{fig:prior:focused:vs:hnet}
    \end{subfigure}
    \caption{\textbf{(a)}: A hypernetwork $h(\mathbf{e}, \theta)$ produces the weights $\psi$ of a recurrent main network $f$ conditioned on $\mathbf{e}$. \textbf{(b)}: Here, we illustrate a hypernetwork-based CL approach versus a weight-importance method (such as EWC). Both methods start learning a second task from a common solution $\psi_1$. EWC is rigid along certain directions in weight space, which leads to a trade-off solution $\psi_2^{\text{EWC}}$ when seeking good optima for the upcoming task (cf. \citet{gal:bcl}). The hypernetwork-based approach (Eq.~\ref{eq:hnet:reg}) has no such trade-off built into its objective; it is only limited by the optimization algorithm and network capacity, and is capable to output the task-specific solutions $\psi_1$ and  $\psi_2^{\text{HNET}}$ (figure adapted from \citet{kirkpatrick:ewc:2017}).
    \vspace{-1em}}
    \label{fig:schematics}
\end{figure}

\vspace{-3mm}
where $\lambda$ is the regularization strength, $\omega_i$ is the \textit{importance} associated with $\psi_i$ (cf. SM \ref{supp:methods:online:ewc} and \ref{supp:methods:si}) and $\tilde{\psi}^{(K-1)}$  denotes the main network weights $\psi$ that were checkpointed after learning task $K-1$. 
We denote by \textbf{HNET} a different regularization approach based on hypernetworks that was recently proposed by \citet{oswald:hypercl}. A hypernetwork \citep{Ha2016Sep} is a neural network $\psi = h(\mathbf{e}, \theta)$ with parameters $\theta$ and input embeddings $\mathbf{e}$ that generates the weights of a main network. This method sidesteps the problem of finding a compromise between tasks with a shared model $\psi$, by generating a task-specific model $\psi^{(k)}$ from a low-dimensional embedding space via a shared hypernetwork in which the weights $\theta$ and embeddings $\mathbf{e}$ are continually learned. In contrast to \citet{oswald:hypercl}, we focus here on RNNs as main networks: $f_\text{step}(\mathbf{x}_t, \mathbf{h}_{t-1}, \psi) = f_\text{step}\big(\mathbf{x}_t, \mathbf{h}_{t-1}, h(\mathbf{e}, \theta) \big)$ (Fig. \ref{fig:hnet:schematic}). Crucially, this method has the advantage of not being noticeably affected by the recurrent nature of the main network, since CL is delegated to a feedforward meta-model, where forgetting is avoided based on a simple L2-regularization of its output.
For a fair comparison, we ensure that the number of trainable parameters is comparable to other baselines by focusing on chunked hypernetworks \citep{oswald:hypercl}, and enforcing: $\big\lvert \theta \cup \{ \mathbf{e}_k \}_{k=1}^K \big\rvert \leq \lvert \psi \rvert$.
Further details can be found in SM \ref{supp:methods:hnets}.
\textbf{Masking} (or \textit{context-dependent gating},  \citet{masse:masking:pnas}) applies a binary random mask per task for all hidden units of a multi-head network, and can be seen as a simple method for selecting a different subnetwork per task. Since catastrophic interference can occur because of the overlap between subnetworks, this method can be combined with other CL methods such as SI (\textbf{Masking+SI}).
We also consider methods based on replaying input data from previous tasks, either via a sequentially trained generative model \citep{shin:dgr:2017, van_de_ven:replay:through:feedback}, denoted \textbf{Generative Replay}, or by maintaining a small subset of previous training data \citep{rebuffi2017icarl, nguyen:2017:vcl}, denoted \textbf{Coresets-$N$}, where $N$ refers to the number of samples stored for each task. Target outputs for replayed data are obtained via a copy of the main network, stored before training on the current task (detailed baseline descriptions in SM \ref{supp:methods}).

\vspace{-2mm}
\paragraph{Task Identity.} We assume that task identity is provided to the system during training and inference, either by selecting the correct output head or by feeding the correct task embedding $\mathbf{e}_k$ into the hypernetwork, and elaborate in SM \ref{supp:remarks:task:id} on how to overcome this limitation.


\section{Analysis of weight-importance methods}
\label{sec:an:copy}

\vspace{-2mm}
Weight-importance methods have widely been used in feedforward networks, but whether or not they are suited for tackling CL in RNNs has remained unclear. Here, we investigate the particularities that weight-importance methods face when applied to RNNs. As opposed to feedforward networks, RNNs provide a natural way to process temporal sequences. Because their hidden states are a function of newly incoming inputs as well as their own activity in the previous timestep, RNNs are able to store and manipulate sample-specific information within their hidden activity, thus providing a form of working memory. Importantly, processing input sequences one sequence element at a time results in the reuse of recurrent weights for a number of times that is equal to the length of the input sequence.
In this section, we investigate whether weight importance values are directly affected by working memory requirements or by the length of processed sequences. For this, we develop some intuitions by studying linear RNNs, which we then test in non-linear RNNs using a synthetic dataset.

First, we obtain some theoretical insights by analysing how linear RNNs can learn to solve a set of tasks (for details refer to  SM \ref{supp:linear:rnns}). Whenever task-specific output heads are not rich enough to model task variabilities, RNNs trained with methods whose recurrent computation is not task-conditioned (e.g. weight-importance methods) must solve all tasks simultaneously within their hidden space. In an extreme scenario where tasks are so different that they cannot share any useful computation, it becomes clear that task interferences can be prevented if the information relevant to each task resides in task-specific orthogonal subspaces (refer to SM \ref{supp:remarks:forward:transfer} for a discussion of scenarios with task similarity). 
Maintaining this structure within the hidden space imposes certain constraints on the recurrent weights. Our theory shows these constraints increase with the number of tasks and with the dimensionality of the task-specific subspaces.
Based on these theoretical insights, we hypothesize that also in nonlinear RNNs increasing working memory requirements cause high weight rigidity (as illustrated by high importance values), whereas the recurrent reuse of weights is not a driving factor.

\begin{figure}[t]
    \includegraphics[width=0.85\textwidth]{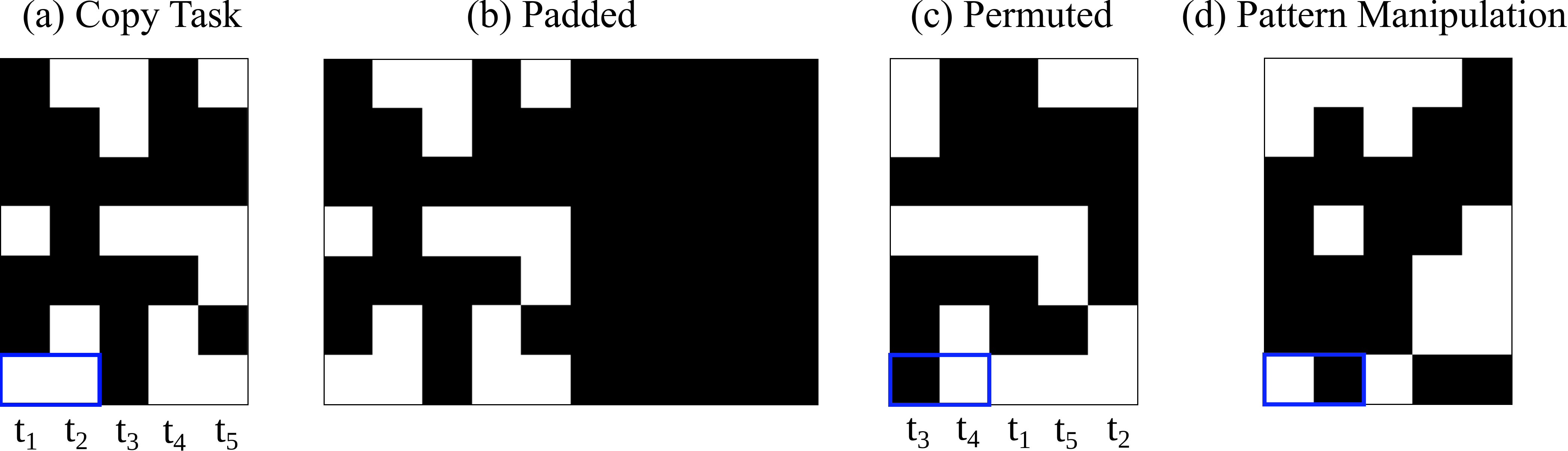}
    \caption{Variants of the Copy Task. \textbf{(a)}: Example pattern of the basic Copy Task where input and output patterns are identical. \textbf{(b)}: In Padded Copy Task, the input is padded with zeros, but the output consists only of the pattern itself, i.e. to (a).  \textbf{(c)}: In Permuted Copy, the output is a time-permuted version of the presented input in (a). \textbf{(d)}: In the Pattern Manipulation Task ($r=1)$, the output corresponds to an XOR operation between the original input (a) and a permuted version of the input (c). The blue rectangle highlights this operation for two specific bits.}
    \label{fig:copy:explained}
\end{figure}

To test this hypothesis, we explore a synthetic dataset consisting of several variations of the Copy Task \citep{graves2014ntm}, in which a random binary input pattern has to be recalled by the network after a stop bit is observed (see Fig. \ref{fig:copy:explained}, and SM \ref{supp:data:copy} for details). For all Copy Task experiments, we use vanilla RNNs combined with orthogonal regularization (see SM \ref{supp:remarks:orth:reg}).
We denote the length (number of timesteps) of the binary input pattern to be copied by $p$, and the actual number of timesteps until the stop bit by $i$ (examples can be found in SM Fig. \ref{fig:supp:copy:examples}). This distinction allows us to consider two variants, the basic Copy Task where $p=i$, and the \textit{Padded Copy Task} where $i>p$ (Fig. \ref{fig:copy:explained} a and b). In this variant, we zero-pad a binary input pattern of length $p$ for $i-p$ timesteps until the occurrence of the stop bit, resulting in an input sequence with $i$ timesteps.
Specifically, we consider a set of Copy Tasks\footnote{Note, these tasks are learned independently to isolate the effects of $i$ and $p$ on weight importance values.} with varying input lengths $i$ and, either a fixed pattern length $p=5$, or a pattern length tied to the input length ($p=i$). 
This allows us to disentangle how sequence length and memory load affect weight importance.

As in \texttt{Online EWC}, we calculate weight importance as the diagonal elements of the empirical Fisher information matrix (see SM \ref{supp:methods:online:ewc}).
To quantify memory load, we study the intrinsic dimensionality of the hidden state of the RNN using principal component analysis (PCA), 
once networks have been trained to achieve near optimal performance (above 99\%). 
We define the intrinsic dimensionality as the number of principal components that are needed to explain 75\% of the variance. 

\begin{figure}[ht!]
  \centering
  \begin{subfigure}[b]{1.65in}
    \includegraphics[width=\textwidth]{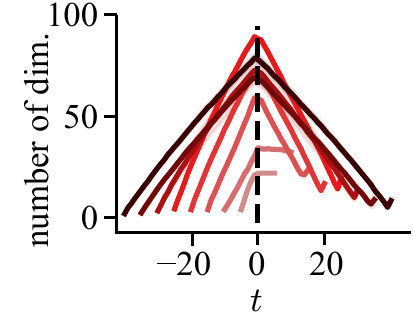}
    \caption{}
    \label{fig:one_task_b}
  \end{subfigure}
  \hfill
  \begin{subfigure}[b]{1.85in}
    \includegraphics[width=\textwidth]{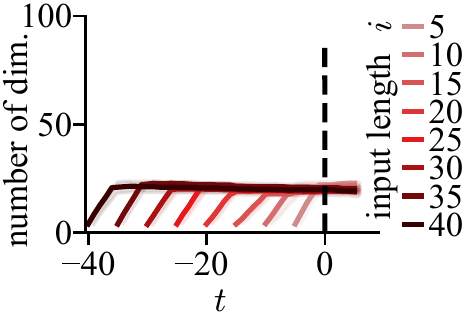}
    \caption{}
    \label{fig:one_task_c_padded}
  \end{subfigure}
  \hfill
  \begin{subfigure}[b]{1.65in}
    \includegraphics[width=\textwidth]{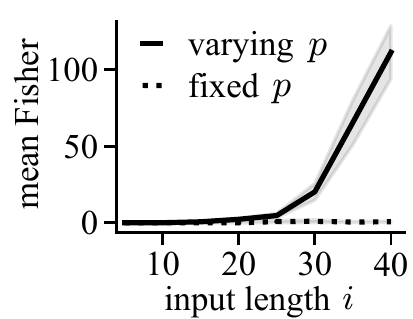}
    \caption{}
    \label{fig:one_task_a_fisher}
  \end{subfigure}
  \caption{\textbf{(a)} Intrinsic dimensionality per timestep of the 256-dimensional RNN hidden space $\mathbf{h}_t$ for the basic Copy Task, where input and pattern lengths are tied ($i$ = $p$). The stop bit (dotted black line) is shown at time $t=0$ (Mean $\pm$ SD, $n=5$). \textbf{(b)} Same as (a) for the Padded Copy Task, where the pattern length is fixed ($p=5$) but input length $i$ varies. In (a) and (b), dimensionality of the hidden state space increases only during input pattern presentation. \textbf{(c)} Mean Fisher values (weight-importance values in \texttt{Online EWC}) of recurrent weights after learning the Copy Task (solid line, $p=5,10,...40$) or the Padded Copy Task (dotted line, $p=5$) independently for an increasing set of sequence lengths $i$ (Mean $\pm$ SD, $n=5$). }
  \label{fig:copy_intuition}
\end{figure}

\vspace{-1.5mm}
As expected, the intrinsic dimensionality of the hidden space increases during input pattern presentation and peaks after $p$ timesteps, i.e. at the stop bit for tasks with $p=i$ (Fig. \ref{fig:one_task_b}) and $i-p$ timesteps before the stop bit if $p=5$ remains fixed (Fig. \ref{fig:one_task_c_padded}).\footnote{Note, Fig. \ref{fig:one_task_b} shows a decreased dimensionality at $t=0$ for $i=35$ or $i=40$ compared to $i=30$. We hypothesize that this is due to a need to non-linearly encode information into the hidden state for large $i$, and verified that the dimensionality increases with $i$ when using Kernel PCA \citep{scholkopf1997kernel:pca}, data not shown.} Weight importance values rapidly increase with memory requirements ($p$), but not sequence length (increasing $i$, fixed $p$) (Fig. \ref{fig:one_task_a_fisher}). The same trend is observed when computing weight-importance values according to \texttt{SI} (cf. SM \ref{supp:remarks:si}), and when using LSTMs instead of vanilla RNNs (cf. SM \ref{supp:remarks:copy:lstm}). We extend this analysis to a CL setting in SM \ref{supp:remarks:supervised:dim:red}.

Overall, this analysis reveals that weight-importance methods are affected by the processing and storage required by the task, but not by the sequential nature of the data, even if the same set of weights is reused over many timesteps.
This could cause weight-importance methods to suffer from a saturation of importance values when working memory requirements are high, which could in turn reduce their plasticity for learning new tasks. We explore this in the next section.


\section{Continual Learning Experiments}
\label{sec:experiments}

\vspace{-2mm}
To highlight strengths and weaknesses of different CL methods in various settings, we performed experiments on one synthetic and two real-world sequential datasets, using different types of RNNs.
We distinguish between \texttt{during} and \texttt{final} accuracies. The \texttt{during} accuracy of a CL experiment is obtained by taking the mean over the test accuracy from each task right after it has been trained on, i.e., when tasks have not yet been subject to forgetting. The \texttt{final} accuracy describes the mean test accuracy over all tasks obtained after the last task has been learned.

For all reported methods, results were obtained via an extensive hyperparameter search, where the hyperparameter configuration of the run with best \texttt{final} accuracy was selected and subsequently tested on multiple random seeds (experimental details in SM \ref{supp:exp:details}). We provide additional CL experiments on multilingual NLP data in SM \ref{supp:remarks:pos}.

\vspace{-1.5mm}
\subsection{Variations of the Copy Task}
\label{sec:exp:copy}

\vspace{-2mm}
After exposing the challenges that weight-importance methods face when dynamically processing data, we explore how these manifest in a CL scenario. We compare weight-importance methods against other CL approaches, with a particular focus on \texttt{HNET}, which can in principle bypass those challenges.
We transform the Copy Task into a set of CL tasks by applying for each task $k$ a different random time-permutation $\pi^{(k)}$. In this setting, which we denote \textit{Permuted Copy Task}, each timestep $t_i \in \{1,\dots,p\}$ from the input pattern has to be recalled at output timestep $t_o = \pi^{(k)}(t_i)$ (Fig. \ref{fig:copy:explained}c). We perform these experiments on vanilla RNNs. First, we evaluate all methods in a relatively simple scenario with five tasks using $p=i=5$ (Table \ref{tab:copy:results:p5}). \texttt{Online EWC} achieves very high performance, and  \texttt{HNET} reaches close to 100\% accuracy. The random subnetworks in \texttt{Masking} can learn individual tasks to perfection. However, weight changes within subnetworks, which result from random overlaps, cause severe performance drops and show the need to add stabilization mechanisms, e.g., \texttt{Masking+SI}. Since the input data distribution is relatively simple and identical across tasks, learning a generative model is feasible, which is illustrated by the performance of \texttt{Gen.} \texttt{Replay}.

\begin{figure}
\begin{floatrow}
\capbtabbox{%
  \begin{small}
  \resizebox{!}{59pt}{%

  \begin{tabular}{lcc} \toprule[0pt]
    & \textbf{during} & \textbf{final}\\  \midrule 
    \texttt{Multitask} & N/A & 99.87 $\pm$	0.05 \\
    \texttt{From-scratch} & N/A & 100.00 $\pm$  0.00   \\
    \texttt{Fine-tuning}  & 99.99 $\pm$  0.00 & 71.05 $\pm$  0.13  \\
    \texttt{HNET}  & 99.98 $\pm$  0.00 & 99.96 $\pm$  0.01    \\
    \texttt{Online EWC} & 99.93 $\pm$  0.01 & 98.66 $\pm$  0.14   \\
    \texttt{SI} & 98.41 $\pm$  0.06 & 94.03 $\pm$  0.24  \\
    \texttt{Masking} & 99.53 $\pm$  0.26 & 72.31 $\pm$  0.82 \\
    \texttt{Masking+SI} & 99.40 $\pm$  0.25 & 99.40 $\pm$  0.25 \\
    \texttt{Gen.} \texttt{Replay} & 100.00 $\pm$  0.00 & 100.00 $\pm$  0.00  \\
    \texttt{Coresets-$100$} & 100.00 $\pm$  0.00 & 99.94 $\pm$  0.00  \\ 
    \bottomrule[0pt]
  \end{tabular}
  }
 \vspace{-1.5em}
  \end{small}
}
{
  \caption{Mean \texttt{during} and \texttt{final} accuracies for the Permuted Copy Task with $p=i=5$  using $5$ tasks each (Mean $\pm$ SEM in \%, $n=10$).}
  \label{tab:copy:results:p5}
}
\hfill
\capbtabbox{%
  \begin{small}
  \resizebox{!}{59pt}{%
  \begin{tabular}{lcc} \toprule[0pt]
    & \textbf{during} & \textbf{final}\\  \midrule
    \multicolumn{3}{l}{\textbf{Padded Copy Task}} \\
    \texttt{HNET}  & 100.00 $\pm$  0.00 & 100.00 $\pm$  0.00 \\
    \texttt{Online EWC} & 97.94 $\pm$  0.09 & 97.89 $\pm$  0.10  \\
    \midrule
    \multicolumn{3}{l}{\textbf{Pattern Manipulation Task} $\mathbf{r=1}$} \\
    \texttt{HNET}  & 100.00 $\pm$ 0.00 & 99.84 $\pm$  0.15 \\
    \texttt{Online EWC} & 98.52 $\pm$  0.27 & 95.45 $\pm$  0.17 \\
    \midrule
    \multicolumn{3}{l}{\textbf{Pattern Manipulation Task} $\mathbf{r=5}$} \\
    \texttt{HNET} & 95.73 $\pm$  1.44 & 93.87 $\pm$  1.24 \\
    \texttt{Online EWC} & 87.40 $\pm$  4.53 & 81.80 $\pm$  3.25 \\
    \bottomrule[0pt]
  \end{tabular}
  }
 \vspace{-1.5em}
  \end{small}
}
{
  \caption{
  Detailed \texttt{Online EWC}/\texttt{HNET} comparisons (Mean $\pm$ SEM in \%, $n=5$). $r$ is the number of task-specific random permutations.}
  \label{tab:copy:results:i25_xor}
}
\end{floatrow}
\end{figure}

In the following, we focus on a comparison between \texttt{Online EWC} and \texttt{HNET} to further investigate how these methods are affected by sequence length and working memory requirements. We first test whether \texttt{Online EWC} is affected by sequence length by investigating the Permuted Copy Task at $p=5, i=25$ using 5 tasks. As Table \ref{tab:copy:results:i25_xor} shows, the performance of both methods is not markedly affected by sequence length. Interestingly, the results are slightly better for longer sequences with both methods, which can be due to an increased processing time between input presentation and recall.
Next, we compare the performance of \texttt{Online EWC} and \texttt{HNET} in a set of tasks for which working memory requirements can be easily controlled. In this setting, referred to as \textit{Pattern Manipulation Task}, difficulty is controlled by a set of $r$ task-specific random permutations along the time axis (Fig. \ref{fig:copy:explained}d). The output is computed from the input pattern by applying a binary XOR operation iteratively with all of its $r$ permutations (i.e. the result of the XOR between the input and its first permutation will then undergo a second XOR operation with the second permutation of the input, and so on).
Note that this variant substantially differs from previous Copy Task variations, since the processing of input patterns is now both input- and task-dependent. As shown in Table \ref{tab:copy:results:i25_xor}, \texttt{Online EWC} experiences a larger drop with increased task difficulty than \texttt{HNET}, confirming that it is more severely affected by working memory requirements. Finally, we investigate the difference between single-head and multi-head settings, as well as task conditional processing in SM \ref{supp:cl:experimental:setting}.  


\vspace{-1.5mm}
\subsection{Sequential Stroke MNIST}
\label{sec:ssmnist}

\vspace{-2mm}
To test whether the results from the synthetic Copy Task hold true for real world data we turned to a sequential digit recognition task where task difficulty can be directly controlled.
In the Stroke MNIST (SMNIST) dataset \citep{deJong2016Nov}, MNIST images \citep{lecun1998gradient} are represented as sequences of pen displacements, that result in the original digits when drawn in order. 
We adapt this dataset to a CL scenario by splitting it into five binary classification problems (digits 0 vs 1, 2 vs 3, etc.), reminiscent of the popular Split-MNIST experiment commonly used to benchmark CL methods on static data \citep{zenke:synaptic:intelligence}. 
Interestingly, this dataset allows exploring how the performance of different CL methods depends on the difficulty of individual tasks by generalizing the notion of Split-SMNIST to sequences of $m$ SMNIST samples \citep[cf.][]{gulcehre2017memory}, where each sequence contains only two types of digits (e.g. $2332$ or $7767$ for $m=4$). To obtain a binary decision problem, we randomly group all $2^m$ possible sequences within a task into two classes.
This ensures that despite increasing levels of task difficulty, as determined by $m$, chance level is not affected. Crucially, an increase in $m$ leads to an increase in the amount of information that needs to be stored and manipulated per input sequence, and therefore allows exploring the effect that increasing working memory requirements have on different CL methods in a real-world dataset.

\begin{figure}
\begin{floatrow}
\ffigbox[\FBwidth]{
    \includegraphics[width=0.45\textwidth]{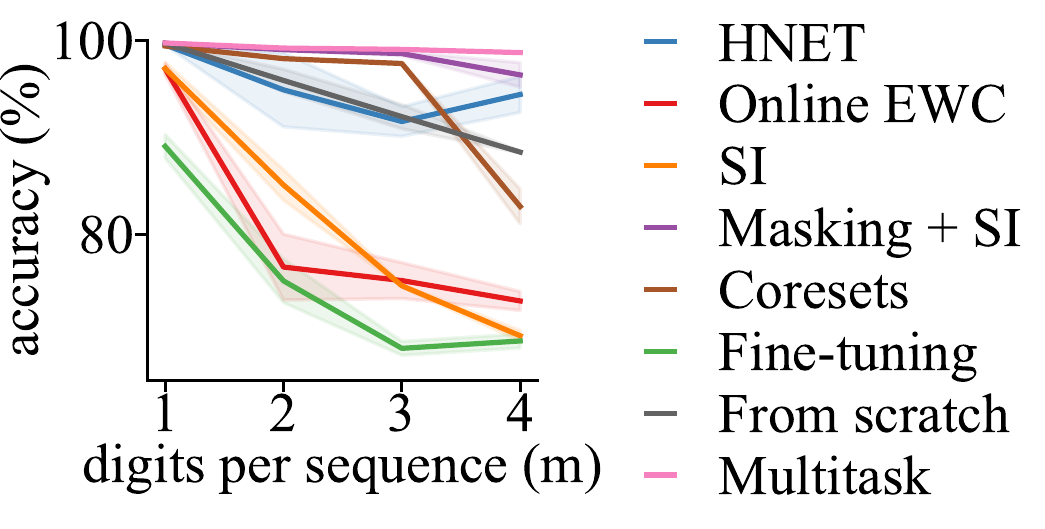}
}{
  \caption{Mean \texttt{final} accuracies for four Split Sequential-SMNIST experiments, each comprising five tasks (Mean $\pm$ SEM, $n=10$).}
  \label{fig:ssmnist_acc_vs_seq_len}
}
\ffigbox[\FBwidth]{
    \includegraphics[width=0.45\textwidth]{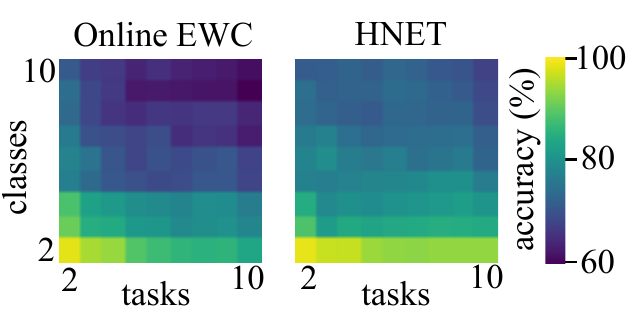}
}{
  \caption{Mean \texttt{final} accuracies for Split-AudioSet experiments with varying number of classes and tasks.}
  \label{fig:audioset_classes_vs_tasks}
}
\end{floatrow}
\end{figure}

We train LSTMs on five tasks for an increasing number of digits per sequence ($m=1,2,3,4$) and observe that methods are differently affected by the task difficulty level (see Fig. \ref{fig:ssmnist_acc_vs_seq_len}). For $m=1$ \texttt{Online EWC}, \texttt{SI} and \texttt{HNET} all achieve above $97 \%$ performance. However for $m=4$ the performance of \texttt{Online EWC} and \texttt{SI} drops to $73.16 \pm 0.98 \%$ and $69.58 \pm 0.55 \%$ respectively, while the hypernetwork approach successfully classifies $94.42 \pm 1.85 \%$ of all inputs. Thus weight-importance methods seem more strongly affected than \texttt{HNET} by an increase in task complexity and working memory requirements. Interestingly, the performance gap depends on the experimental setup as outlined in SM \ref{supp:cl:experimental:setting}.
\texttt{Coresets} perform slightly worse, especially when task-complexity increases. \texttt{Masking+SI}, which trades-off network capacity for the ability of finding solutions in a less rigid subnetwork, emerges as the preferable method for this experiment. We additionally list \texttt{during} accuracies for all methods in SM Table \ref{supp:tab:ssmnist:results} and discuss the use of replay for Split-SMNIST in SM \ref{supp:remarks:smnist:replay}.
Finally we show that, consistent with our Copy Task results, the performance of weight-importance methods is not significantly affected when sequence lengths are increased without a concomitant increase in working memory (cf. SM. \ref{supp:remarks:upsampling}). 

\vspace{-1.5mm} 
\subsection{AudioSet}
\label{sec:audioset}

\setlength\intextsep{0pt}
\begin{wraptable}{r}{.5\textwidth}
 \centering
  \caption{Mean \texttt{during} and \texttt{final} accuracies for the Split-AudioSet-10 experiments (Mean $\pm$ SEM in \%, $n=10$).}
  \begin{small}
      \begin{tabular}{lcc}
        & \textbf{during} & \textbf{final} \\ \toprule
        \texttt{Multitask} & N/A & 77.31 $\pm$ 0.10\\
        \texttt{From-scratch} & N/A & 79.06 $\pm$  0.11\\
        \texttt{Fine-tuning} & 71.95 $\pm$  0.24 & 49.02 $\pm$  1.00  \\
        \texttt{HNET} & 73.05 $\pm$ 0.45 & 71.76 $\pm$ 0.62\\
        \texttt{Online EWC} & 68.82 $\pm$  0.20 & 65.56 $\pm$  0.35\\
        \texttt{SI} & 67.66 $\pm$  0.10 & 66.92 $\pm$  0.04 \\
        \texttt{Masking} & 75.81 $\pm$  0.15 & 50.87 $\pm$  1.09 \\
        \texttt{Masking+SI} & 64.88 $\pm$  0.19 & 64.86 $\pm$  0.20  \\
        \texttt{Coresets-$100$} & 74.25 $\pm$  0.11 & 72.30 $\pm$  0.11 \\
        \texttt{Coresets-$500$} & 77.03 $\pm$  0.08 & 73.90 $\pm$  0.07 \\
      \end{tabular}
  \label{tab:audioset:CL1:results}
  \end{small}
\end{wraptable}

AudioSet \citep{audioset} is a dataset of manually annotated audio events. It consists of 10-second audio snippets which have been preprocessed by a VGG network to extract 128-dimensional feature vectors at 1 Hz. This dataset has been previously adapted for CL by \citet{kemker2018fearnet} and \citet{kemker2018measuring}, whose particular split has not been made public, and by \citet{cossu2020continual}, for which the test set size largely differed across classes.
We therefore created a new variant, which we call \textit{Split-AudioSet-10}, containing 10 tasks with 10 classes each (see SM \ref{supp:exp:details:audioset} for details). 

The results obtained in this dataset using LSTMs are listed in Table \ref{tab:audioset:CL1:results}.
\texttt{HNET} is the strongest among regularization based methods, and is only outperformed by \texttt{Coresets}, which rely on storing past data.
\texttt{Masking} \texttt{during} accuracies indicate that random subnetworks have enough capacity to learn individual tasks, but low \texttt{final} accuracies suggest that catastrophic forgetting occurs, presumably because of the overlap between subnetworks. This is partly solved in \texttt{Masking+SI} by introducing stabilization which, however, reduces plasticity for learning new tasks. In contrast to the findings of Sec. \ref{sec:ssmnist} and SM \ref{supp:remarks:pos}, \texttt{Masking+SI} performs worst among regularization approaches, indicating that trading-off capacity for complex datasets can be harmful.
The \texttt{From-scratch} baseline outperforms other methods, which is explained by the fact that it trains a separate model per task, leading to 10 times more network capacity.
Notably, we were not able to successfully train a \texttt{Generative Replay} model on this dataset despite extensive hyperparameter search. Together with the results in Sec. \ref{sec:exp:copy}, this highlights that the performance of \texttt{Generative Replay} depends on the complexity of the input data distribution, and not necessarily on the CL nature of the problem.
To further investigate the stability-plasticity trade-off, we tested \texttt{HNET} and \texttt{Online EWC} across a range of difficulty levels in individual tasks. This can be controlled by the number of classes to be learned within each task, which we varied from two to ten. For both methods we used the best hyperparameters found for Split-AudioSet-10.
Since the performance of \texttt{Online EWC} strongly depends on the regularization parameter $\lambda$, we tuned this value to achieve optimal results in each setting (cf. Fig. \ref{supp:fig:audioset_lambdas}). 
Fig. \ref{fig:audioset_classes_vs_tasks} shows the task-averaged \texttt{final} accuracies for the different task-difficulty settings. 
While \texttt{HNET} performance is primarily affected by task difficulty but not by the number of tasks, results for \texttt{Online EWC} show an interplay between task difficulty and the ability to retain good performance on many tasks. These results provide further evidence that the hypernetwork-based approach can resolve the limitations of weight-importance CL methods for sequential data.


\vspace{-1.5mm}
\section{Discussion}
\label{sec:discussion}

\vspace{-2mm}
\paragraph{The stability-plasticity dilemma with sequential data.}
Weight-importance methods address CL by progressively constraining a network's weights, directly trading plasticity for stability.
In the case of RNNs, weights are subject to additional constraints, since the same set of weights is reused across time to dynamically process an input stream of data.
We show that increased working memory requirements, resulting from more complex processing within individual tasks, lead to high weight-importance values and can hinder the ability to learn future tasks (cf. Sec. \ref{sec:an:copy}). On the contrary, we find that longer sequence lengths do not impact performance for a fixed level of task complexity (cf. Fig. \ref{fig:one_task_c_padded}, Table \ref{tab:copy:results:i25_xor}), suggesting that weight reuse doesn't interfere with the RNN's ability to retain previous knowledge. These observations are consistent with our theoretical analysis of linear RNNs (SM \ref{supp:linear:rnns}), which predicts that more challenging processing within individual tasks leads to increased interference between tasks. This aggravates the stability-plasticity dilemma in weight-importance based methods, which we confirm in a range of experiments.

\vspace{-2mm}
\paragraph{Benefits of a hypernetwork-based CL approach for sequential data.}
We propose that a hypernetwork-based approach \citep{oswald:hypercl} can alleviate the stability-plasticity dilemma when continually learning with RNNs.
Since stability is outsourced to a regularizer that does not directly limit the plasticity of main network weights (cf. SM Eq.~\ref{eq:hnet:reg}), this approach has more flexibility than weight-importance methods for finding new solutions, as shown by our experiments. Although \texttt{Coresets} and \texttt{Generative Replay} perform better, these approaches might not always be applicable; \texttt{Coresets} rely on the storage of past data (which might not always be feasible for privacy or storage reasons), and \texttt{Generative Replay} scales poorly to complex data. On the contrary, an approach based on hypernetworks has the versatility of regularization approaches and can be used in a variety of situations.

\vspace{-2mm}
\paragraph{Future avenues for CL with RNNs.} As discussed in SM \ref{supp:cl:experimental:setting}, an interesting avenue for weight-importance methods is the use of task-conditional processing in order to overcome the need of solving all learned tasks in parallel. Although hypernetworks overcome this limitation by design, they introduce additional optimization challenges, especially in conjunction with vanilla RNNs (cf. SM \ref{supp:remarks:orth:reg}), which leaves room for future improvements. 
An interesting direction is the use of a recurrent hypernetwork to generate timestep-specific weights in the main RNN \citep{Ha2016Sep, recurrent:highway:hypernets}. Although a naive application of this combination for CL using SM Eq.~\ref{eq:hnet:reg} would come at the cost of a linear increase in computation with the number of timesteps, this problem can be elegantly sidestepped by the use of a feed-forward hypernetwork that generates the weights of the recurrent hypernetwork. SM Eq.~\ref{eq:hnet:reg} can then simply be applied to the static output of this \textit{hyper-hypernetwork}, protecting a set of timestep-specific weights per task without the need to increase the regularization budget.

\vspace{-1.5mm}
\section{Conclusion}
\label{sec:conclusion}

\vspace{-2mm}
Our work advances the CL community in three ways. First, by systematically evaluating the performance of established CL methods when applied to RNNs, we provide extensive baselines that can serve as reference for future studies on CL with sequential data. 
Second, we use theoretical arguments derived from linear RNNs to hypothesize limitations of weight-importance based CL in the context of recurrent computation, and provide empirical evidence to support these statements.
Third, derived from these insights, we suggest that an approach based on hypernetworks mitigates the stability-plasticity dilemma, and show that it outperforms weight-importance methods on synthetic as well as real-world data. Finally, our work discusses several future improvements and directions of CL approaches for sequential data.

\section*{Acknowledgements}
This work was supported by the Swiss National Science Foundation (B.F.G. CRSII5-173721 and 315230\_189251), ETH project funding (B.F.G. ETH-20 19-01) and funding from the Swiss Data Science Center (B.F.G, C17-18, J. v. O. P18-03). We especially thank João Sacramento for insightful discussions, ideas and feedback. We would also like to thank Nikola Nikolov and Seijin Kobayashi for helpful advice and James Runnalls for proofreading our manuscript.

\bibliography{main}
\bibliographystyle{mynatbib}


\newpage
\setcounter{page}{1}
\setcounter{figure}{0} \renewcommand{\thefigure}{S\arabic{figure}}
\appendix

\section*{\Large{Supplementary Material:\\Continual Learning in Recurrent Neural Networks}}
\textbf{Benjamin Ehret*, Christian Henning*, Maria R. Cervera*, Alexander Meulemans, Johannes von Oswald, Benjamin F.~Grewe}

\section{Summary of notation}
\label{supp:notation}

In this section we define the mathematical notation that we consistently use throughout the paper. We consider the successive learning of $K$ datasets $\mathcal{D}_k = \big\{ (\mathbf{x}_{1:T_\text{in}^{(n)}}^{(n)}, \mathbf{y}_{1:T_\text{out}^{(n)}}^{(n)}) \big\}_{n=1}^{N_k}$. A data sample $(\mathbf{x}_{1:T_\text{in}}, \mathbf{y}_{1:T_\text{out}})$ consists of a sequence of inputs $\mathbf{x}_{1:T_\text{in}} = (\mathbf{x}_1, \dots, \mathbf{x}_{T_\text{in}})$, $\mathbf{x}_t \in \mathcal{R}^{F_\text{in}}$, and a sequence of target outputs $\mathbf{y}_{1:T_\text{out}} = (\mathbf{y}_1, \dots, \mathbf{y}_{T_\text{out}})$, $\mathbf{y}_t \in \mathcal{R}^{F_\text{out}}$, where $T_\text{in}$/$T_\text{out}$ denote the time dimension and $F_\text{in}$/$F_\text{out}$ the feature dimension, respectively. In general, the number of timesteps is sample-dependent and not constant.

The \textbf{main network}, which processes data from the datasets $\{ \mathcal{D}_k \}_{k=1}^K$, is an RNN with parameters $\psi$. With an abuse of notation, we describe it by $\mathbf{\hat{y}}_{1:T_\text{out}} = f(\mathbf{x}_{1:T_\text{in}}, \psi)$. To express the step-by-step computation of the RNN we use $(\mathbf{\hat{y}}_t, \mathbf{h}_t) = f_\text{step}(\mathbf{x}_t, \mathbf{h}_{t-1}, \psi)$. Specifically, we denote by $\psi_\text{hh}$ the hidden-to-hidden weights, which are a subset of $\psi$ and are exclusively involved in the computation from $\mathbf{h}_{t-1}$ to $\mathbf{h}_t$. The \textbf{hypernetwork} is a feedforward neural network $\psi = h(\mathbf{e}_k, \theta)$ with parameters $\theta$, that generates the parameters $\psi$ of the main network given the task embedding $\mathbf{e}_k$ of task $k$.

\section{Detailed description of all methods}
\label{supp:methods}

Here, we provide a mathematical description of all methods mentioned in Sec. \ref{sec:methods}, together with an estimate of their time and space complexity increase when compared to the naive \texttt{Fine-tuning} baseline.

The task-specific loss functions $\mathcal{L}_\text{task}(\psi, \mathcal{D}_k)$ applied across all methods are described in Sec. \ref{supp:methods:online:ewc} (cf. Eq.~\ref{supp:eq:nll:cross:entropy} and Eq.~\ref{supp:eq:nll:copy}).

\subsection{Fine-tuning}
\label{supp:methods:fine:tuning}

\texttt{Fine-tuning} \citep{li2017learning:without:forgetting} refers to sequentially optimizing the task-loss $\mathcal{L}_\text{task}(\psi, \mathcal{D}_k)$ for $k=1,\dots,K$ without any explicit protection against catastrophic forgetting. However, since each task has its own output head, the output head weights are task-specific and fixed for past tasks.

Even though \texttt{Fine-tuning} has no built-in mechanism to prevent forgetting, we selected the hyperparameter configuration based on the best \texttt{final} accuracy. This ensured consistency with other methods, and allowed directly assessing improvements when employing CL methods.

\subsection{Training from scratch}

\texttt{From-scratch} refers to the independent training of a set of network parameters $\psi^{(k)}$ per task, i.e., $K$ separate networks are trained by minimizing $\mathcal{L}_\text{task}(\psi^{(k)}, \mathcal{D}_k)$.

\paragraph{Complexity estimation.} This approach does not add time complexity, but leads to a linear increase in the memory requirements with the number of tasks.

\subsection{Multitask}

\texttt{Multitask}, or joint training \citep{li2017learning:without:forgetting}, refers to jointly training on all datasets at once: $\min_\psi \sum_{k=1}^K \mathcal{L}_\text{task}(\psi, \mathcal{D}_k)$. We performed joint training by assembling a mini-batch of size $B$ using samples equally distributed across all $K$ datasets. Note that in order to provide a fair comparison to our CL baselines, the main network is still a multi-head network with a task-specific fully-connected output layer per task. Thus, the task identity has to be provided during inference in order to select the correct output head.

\paragraph{Complexity estimation.} Even though this approach does not lead to time or memory complexity increases, it requires all data to be available at all times.


\subsection{Hypernetwork-protected models}
\label{supp:methods:hnets}

The hypernetwork-based CL approach, \texttt{HNET} \citep{oswald:hypercl}, is an L2-regularization technique that, in contrast to weight-importance methods, aims to fix certain input-output mappings of a secondary neural network, instead of directly fixing the weights of a main network (cf. Eq.~\ref{eq:hnet:reg}).
The complete loss function for learning the $K$-th task is given by:\footnote{We slightly modified the original regularizer by excluding the lookahead $\Delta \theta$ used in \citet{oswald:hypercl} and by allowing fine-tuning of previous task embeddings, which requires us to additionally checkpoint these task embeddings before learning a new task.}

\begin{equation}
    \label{eq:hnet:reg}
    \mathcal{L}(\theta, \mathbf{e}_1, \dots, \mathbf{e}_K, \mathcal{D}_K) = \mathcal{L}_\text{task}(\theta, \mathbf{e}_K, \mathcal{D}_K) + \frac{\beta}{K-1} \sum_{k=1}^{K-1} \lVert  h(\mathbf{e}_k, \theta) -  h(\mathbf{\tilde{e}}_k^{(K-1)}, \tilde{\theta}^{(K-1)})  \rVert_2^2
\end{equation}

where $\mathcal{D}_K$ is the dataset of task $K$, $\mathcal{L}_\text{task}(\cdot)$ is the loss function of the current task, $\beta$ is the regularization strength and $\tilde{\theta}^{(K-1)}, \mathbf{\tilde{e}}_1^{(K-1)}, \dots, \mathbf{\tilde{e}}_{K-1}^{(K-1)}$ denote hypernetwork weights $\theta$ and task embeddings that were checkpointed after learning task $K-1$. These checkpointed weights are fixed and needed to compute the regularization targets, which ensure that the output of the network stays constant for previously learned tasks, thus preventing forgetting.

To establish a fair comparison to other methods (in terms of number of trainable weights), we used the chunking approach described in \citet{oswald:hypercl}, who showed that in the non-parametric limit a chunked hypernetwork can realize all possible continuous mappings between embedding and weight space. This method splits the vectorized main network weights $\psi$ into equally sized chunks. Each chunk will be assigned a chunk embedding $\mathbf{c}_i$. The hypernetwork can then produce all weights $\psi$ by processing a batch of chunk embeddings (utilizing parallelization on modern GPUs): $\psi = h(\mathbf{e}_k, \theta=\tilde{\theta} \cup \{ \mathbf{c}_i \}) = \text{concat}([\dots, \tilde{h}(\mathbf{e}_k, \mathbf{c}_i, \tilde{\theta}) , \dots])$. In our implementation chunk embeddings are considered to be part of $\theta$ and are therefore shared across tasks.

This approach to chunking is agnostic to the structure that $\psi$ takes in the main network through $f$'s architectural design. Therefore, we investigated other approaches to chunking that respect the architecture of $f$. For instance, if $W_\text{hh} \in \mathbb{R}^{n_h \times n_h}$ and $W_\text{ih} \in \mathbb{R}^{n_h \times n_i}$ denote the weights of a recurrent layer, where $n_h$ and $n_i$ are the number of hidden and input units respectively, the hypernetwork can be designed to produce chunks $V_{\text{hh},i}, V_{\text{ih},i} = \tilde{h}(\mathbf{e}_k, \mathbf{c}_i, \tilde{\theta})$, with $V_{\text{hh},i} \in \mathbb{R}^{n_c \times n_h}$, $V_{\text{ih},i} \in \mathbb{R}^{n_c \times n_i}$ and $0 \equiv n_h \pmod {n_c}$. However, since we didn't observe any improvements in a set of exploratory experiments, all reported results were obtained using the approach suggested in \citet{oswald:hypercl}.

In addition, we would like to mention two properties of the hypernetwork approach that have been empirically verified \citep{oswald:hypercl}. First, the approach supports positive forward transfer, as the knowledge of previous tasks is entangled in the shared meta-model. Experiments on a low-dimensional task embedding space in \citet{oswald:hypercl} seem to indicate that the learned embedding space possesses a structure that supports transfer. Second, \citet{oswald:hypercl} noted and showed empirically that the regularizer in Eq.~\ref{eq:hnet:reg} does not have to increase linearly with the number of tasks $K$, but can instead be subsampled using a random set of $C$ tasks for each loss evaluation. 
We verified this in the Permuted Copy Task, where computing the regularizer for a single randomly chosen task ($C=1$) at each loss evaluation did not lead to a performance decrease for patterns of length $p=5$ (data not shown).

\paragraph{Complexity estimation.}  Independent of its application to CL, the use of a hypernetwork increases time complexity because weights need to be generated before being used for the forward computation of the main network. Another factor contributing to the increase in time complexity is the regularizer (Eq.~\ref{eq:hnet:reg}), which is a sum of L2 norms of the hypernetwork output (of size $\lvert \psi \rvert$) over past tasks, yielding a time complexity of $\mathcal{O}(K\lvert \psi \rvert)$ if the regularizer is applied to all previous tasks, and $\mathcal{O}(C\lvert \psi \rvert)$ otherwise.

Space complexity also increases due to two factors. First, a second network object (i.e., the hypernetwork) has to be maintained in memory. Second, the computation of the regularizer (Eq.~\ref{eq:hnet:reg}) requires storing a set of checkpointed hypernetwork weights and task embeddings when training on a new task. Since we restrict here our analyses to settings where $\big\lvert \theta \cup \{ \mathbf{e}_k \}_{k=1}^K \big\rvert \approx \lvert \psi \rvert$, we simply denote this space complexity increase by $\mathcal{O}(\lvert \psi \rvert)$.


\subsection{Elastic weight consolidation}
\label{supp:methods:online:ewc}

Here, we quickly recapitulate the basic concepts behind elastic weight consolidation (EWC, \citet{kirkpatrick:ewc:2017}). Since EWC is a prior-focused method \citep{gal:bcl}, solutions of upcoming tasks must lie inside the posterior parameter distribution of previous tasks. To achieve this, EWC approximates the posterior via a Gaussian distribution with diagonal covariance matrix. Note that this restriction does not apply to task-specific weights, which may be restricted by an arbitrary choice of the prior. However, to avoid overly cluttered notation, we explicitly ignore the multi-head setting in this section, where parameters $\psi$ can be split into task-specific (the corresponding output head's weights) and task-shared (all weights excluding the output layer) weights. 

EWC makes use of the fact that Bayes rule allows the following decomposition of the posterior parameter distribution:

\begin{equation}
    \label{supp:eq:recursive:bayes:posterior}
    p(\psi \mid \mathcal{D}_1, \dots, \mathcal{D}_K) \propto
    p(\psi \mid \mathcal{D}_1, \dots, \mathcal{D}_{K-1}) \,
    p(\mathcal{D}_K \mid \psi)
\end{equation}

where $p(\psi \mid \mathcal{D}_1, \dots, \mathcal{D}_{K-1})$ is the posterior from previous tasks and $p(\mathcal{D}_K \mid \psi)$ the likelihood of the current task. The precise derivation of the algorithm described here can be found in \citet{huszar:ewc:note:2018}, and has been termed \texttt{Online EWC} in \citet{schwarz:online:ewc}.

When learning task $K$, we aim to find a maximum a posteriori (MAP) solution of $p(\psi \mid \mathcal{D}_1, \dots, \mathcal{D}_K)$ maximizing the following loss function:

\begin{equation}
    \label{supp:eq:prior:focused:loss}
    \max_\psi \log p(\mathcal{D}_K \mid \psi) + \log p(\psi \mid \mathcal{D}_1, \dots, \mathcal{D}_{K-1})
\end{equation}

We discuss the likelihood function for sequential data below. To obtain a tractable loss function, EWC utilizes an approximate posterior $q_\zeta^{(K-1)}(\psi) \approx p(\psi \mid \mathcal{D}_1, \dots, \mathcal{D}_{K-1})$, whose parameters $\zeta$ are computed at the end of task $K-1$. Specifically, EWC first applies a Laplace approximation \cite{mackay1992laplace:approx} (using the MAP solution $\tilde{\psi}^{(K-1)}$ obtained at the end of training of task $K-1$) to obtain a Gaussian $q_\zeta^{(K-1)}(\psi)$ with mean $\tilde{\psi}^{(K-1)}$ and precision matrix $F = \sum_{k=1}^{K-1} F^{(k)}$, where $F^{(k)}$ denotes the empirical Fisher matrix.\footnote{\citet{schwarz:online:ewc} introduced an additional hyperparameter $\gamma_F \leq 1$ to explicitly promote forgetting: $F = \sum_{k=1}^{K-1} \gamma_F^{K-1-k} F^{(k)}$. We left $\gamma_F = 1$ throughout this work.} As noted in \citet{huszar:ewc:note:2018}, this version of \texttt{Online EWC} still does not carry out the Laplace approximation correctly, as the precision matrix of $q_\zeta^{(K-1)}(\psi)$ misses the prior influence and the individual terms $F^{(k)}$ are not properly scaled. However, if the prior influence on the precision matrix is ignored and dataset sizes are identical, then the proper scaling can be absorbed into the regularization strength $\lambda_\text{EWC}$. As a second approximation, EWC considers all off-diagonal elements of $F$ to be zero: $F_{i \neq j} = 0$. Taken together, while ignoring all terms independent of $\psi$, the loss described by  Eq.~\ref{supp:eq:prior:focused:loss} is approximated in \texttt{Online EWC} via (cf. Eq.~\ref{eq:weight:importance:reg}):

\begin{equation}
    \label{supp:eq:ewc:loss}
    \min_\psi - \log p(\mathcal{D}_K \mid \psi) + \lambda_\text{EWC} \sum_{i=1}^{\lvert \psi \rvert} F_{ii} (\psi_i - \tilde{\psi}_i^{(K-1)} )^2
\end{equation}

where $F_{ii}$ can be considered as weight-specific importance values and $\mathcal{L}_\text{task}(\psi, \mathcal{D}_K) \equiv - \log p(\mathcal{D}_K \mid \psi)$ describes the negative log-likelihood (NLL) detailed below.

Note that the correct deployment of Eq.~\ref{supp:eq:prior:focused:loss} requires obtaining a MAP estimate for the first task: $\tilde{\psi}_i^{(1)} = \arg\max_\psi \log p(\mathcal{D}_1 \mid \psi) + \log p(\psi)$. However, we ignored the prior influence when obtaining $\tilde{\psi}_i^{(1)}$.

\paragraph{Negative log-likelihood (NLL) for sequential data.} Finally, we discuss how to implement $\mathcal{L}_\text{task}(\psi, \mathcal{D}_K) \equiv - \log p(\mathcal{D}_K \mid \psi)$ when applied to sequential data. Note that $p(\mathcal{D}_K \mid \psi) = \prod_{n=1}^{N_K} p(\mathbf{y}_{1:T_\text{out}^{(n)}}^{(n)} \mid \psi)$ and that
$p(\mathbf{y}_{1:T_\text{out}} \mid \psi) = \prod_{t=1}^{T_\text{out}} p(\mathbf{y}_t \mid \mathbf{y}_1, \dots, \mathbf{y}_{t-1}, \psi)$. Given the autoregressive structure of an RNN, we make the following assumption: $p(\mathbf{y}_t \mid \mathbf{y}_1, \dots, \mathbf{y}_{t-1}, \psi) \approx p(\mathbf{y}_t \mid \mathbf{h}_{t-1}, \psi)$. Hence, we can decompose the NLL as follows:

\begin{equation}
    \label{supp:eq:nll:sequential}
    - \log p(\mathcal{D}_K \mid \psi) = - \sum_{n=1}^{N_K} \sum_{t=1}^{T_\text{out}^{(n)}} \log p(\mathbf{y}_t^{(n)} \mid \mathbf{h}_{t-1}^{(n)}, \psi)
\end{equation}

We first consider typical classification problems (cf. Sec. \ref{sec:ssmnist} and Sec. \ref{sec:audioset}). In this case, $\mathbf{y}_t^{(n)}$ is a one-hot encoded representation of a label $y_t^{(n)} \in \{1, \dots, F_\text{out}\}$, where $F_\text{out}$ denotes the number of classes. We consider a sofmax output $\mathbf{\hat{y}_t^{(n)}} = \text{softmax}(\tilde{\beta}_t \mathbf{\hat{z}_t^{(n)}})$, where $\tilde{\beta}_t$ denotes a timestep-specific inverse temperature that may be used to bias the loss such that it puts more emphasis on certain timesteps. For instance, setting $\tilde{\beta}_t = 0$ results in timestep $t$ being ignored for the computation of the loss. Indeed, for the experiments in Sec. \ref{sec:ssmnist} and Sec. \ref{sec:audioset}, the loss is evaluated solely based on the prediction of the last timestep $T_\text{in}^{(n)}$ of the given input sequence. Using this setting for classification problems leads to the well-known cross-entropy loss evaluated per timestep and summed over all timesteps:

\begin{equation}
    \label{supp:eq:nll:cross:entropy}
     - \sum_{n=1}^{N_K} \sum_{t=1}^{T_\text{out}^{(n)}} \log p(\mathbf{y}_t^{(n)} \mid \mathbf{h}_{t-1}^{(n)}, \psi)
    = - \sum_{n=1}^{N_K} \sum_{t=1}^{T_\text{out}^{(n)}} \sum_{c=1}^{F_\text{out}} \, [y_t^{(n)} = c] \log \big( \text{softmax}(\tilde{\beta}_t \mathbf{\hat{z}_t^{(n)}})_c \big)
\end{equation}

where $[\cdot]$ denotes the Iverson bracket and $\text{softmax}(\cdot)_c$ refers to the $c$-th entry of the softmax output vector.

Lastly, we consider the NLL for the Copy Task and its variants (cf. Sec. \ref{sec:exp:copy}), where the output has to match a binary target pattern. In this case, each pixel in the output pattern will be evaluated (independent of all other pixels) using a binary cross-entropy loss. Likelihood predictions of pixel values are obtained via a (tempered) sigmoid: $\hat{y}_{t,f}^{(n)} = \text{sigmoid}(\tilde{\beta}_{t,f} \hat{z}_{t,f}^{(n)})$, where $\hat{y}_{t,f}^{(n)}$ denotes the $f$-th entry of $\mathbf{\hat{y}}_t^{(n)}$, and $\tilde{\beta}_{t,f}$ can be interpreted as an inverse temperature that can be specified per timestep and feature. Taken together, the NLL loss for matching binary output patterns can be specified via:

\begin{equation}
    \label{supp:eq:nll:copy}
     - \sum_{n=1}^{N_K} \sum_{t=1}^{T_\text{out}^{(n)}} \log p(\mathbf{y}_t^{(n)} \mid \mathbf{h}_{t-1}^{(n)}, \psi)
    = \sum_{n=1}^{N_K} \sum_{t=1}^{T_\text{out}^{(n)}} \sum_{f=1}^{F_\text{out}} \bigg( - y_{t,f}^{(n)} \log \hat{y}_{t,f}^{(n)} - (1 - y_{t,f}^{(n)}) \log (1 - \hat{y}_{t,f}^{(n)}) \bigg)
\end{equation}

\paragraph{Conceptual differences to a hypernetwork-based approach.} An important conceptual difference between EWC (and prior-focused methods in general) and the hypernetwork-based approach (cf. Sec. \ref{supp:methods:hnets}) lies in the nature of Eq.~\ref{supp:eq:recursive:bayes:posterior}. Whereas prior-focused methods aim to find $\arg\max_\psi p(\psi \mid \mathcal{D}_1, \dots, \mathcal{D}_K)$ (which necessitates a certain compatibility across tasks), the hypernetwork-based approach allows task-specific solutions $\psi^{(k)} = \arg\max_\psi p(\psi \mid \mathcal{D}_k)$, where knowledge transfer between tasks (to exploit compatibilities) is implicitly outsourced to a meta-model (the hypernetwork).

\paragraph{Complexity estimation.}
The regularization introduced in Eq.~\ref{supp:eq:ewc:loss} leads to a time complexity increase of $\mathcal{O}(\lvert \psi \rvert)$ when computing the loss.
Additionally, the computation of Fisher values at the end of each of the $K$ tasks leads to a further increase in time complexity.
Indeed, a forward and backward computation for each sample is performed, while accumulating importance values for each entry in $\psi$. Assuming forward and backward computation only increases linearly with $\psi$, we can summarize this contribution via $\mathcal{O}(\lvert \psi \rvert \sum_k N_k)$, where $N_k$ is the number of samples in task $k$.

The increase in space complexity arises due to the storage of the diagonal Fisher elements as well as the most recent MAP solution: $\mathcal{O}(2\lvert \psi \rvert)$.


\subsection{Synaptic intelligence}
\label{supp:methods:si}

Synaptic intelligence (SI, \citet{zenke:synaptic:intelligence}) is another weight-importance method that, in contrast to EWC, computes the importance values online, i.e., during training rather than at the end of training. The method is based on a first-order Taylor approximation to estimate the loss change after an optimizer update step. This allows estimating the influence of each individual weight $\psi_i$ on the loss change. Thus, at each optimization step $s$ while training task $k$, an online importance estimate $\tilde{\omega}_i^{(k)}$ of $\psi_i$ is updated via:

\begin{equation}
    \label{supp:eq:si:online:importance:estimate}
    \tilde{\omega}_i^{(k)} \leftarrow \tilde{\omega}_i^{(k)} - \Delta \psi_i(s) \frac{\partial \mathcal{L}_\text{task}\big(\psi, \mathcal{B}(s)\big)}{\partial \psi_i}
\end{equation}

where $\Delta \psi_i(s)$ is the weight change determined by the optimizer at step $s$, and $\mathcal{B}(s) \subseteq \mathcal{D}_k$ is the $s$-th minibatch. Importantly, we compute both the optimizer update $\Delta \psi_i(s)$ and the gradient based on the task-specific loss $\mathcal{L}_\text{task}(\cdot)$ only, ignoring potential regularizers such as the SI regularizer itself. To do so, we compute the update step $\Delta \psi_i(s)$ that would be taken by the optimizer without actually taking it. Interestingly, we did not observe a noticeable difference between this variant, where importance is solely based on task-specific influences, and one where the full loss is taken into consideration.

After training of task $k$ is completed, the final importance values $\Omega_i^{(k)}$ are computed as follows:

\begin{equation}
    \label{supp:eq:si:importance:values}
    \Omega_i^{(k)} = \Omega_i^{(k-1)} + \frac{\tilde{\omega}_i^{(k)}}{\Delta \psi_i^{(k)} + \epsilon}
\end{equation}

where $\Delta \psi_i^{(k)}$ is the complete weight change (of weight $\psi_i$) between before and after training on task $k$, and $\epsilon$ ($=1e-3$) ensures numerical stability. If $\tilde{\omega}_i^{(k)} < 0$, we clamp its value to zero to avoid negative importance values. The SI loss function for training task $K$ is:

\begin{equation}
    \label{supp:eq:si:loss}
    \min_\psi \mathcal{L}_\text{task}(\psi, \mathcal{D}_K) + \lambda_\text{SI} \sum_{i=1}^{\lvert \psi \rvert} \Omega_i^{(K-1)} (\psi_i - \tilde{\psi}_i^{(K-1)} )^2
\end{equation}

\paragraph{Complexity estimation.} The increase in time complexity due to the regularization introduced in Eq.~\ref{supp:eq:si:loss} can be summarized as $\mathcal{O}(\lvert \psi \rvert)$ per loss evaluation. An additional increase arises due to the online estimation of importance values (cf. Eq.~\ref{supp:eq:si:online:importance:estimate}). The contribution is bounded by $\mathcal{O}(\lvert \psi \rvert)$ per training iteration.

The increase in space complexity arises due to the storage of $\Omega_i^{(K)}$, $\tilde{\psi}_i^{(K-1)}$, $\tilde{\omega}_i^{(K)}$, as well as a temporary copy of $\psi$ from before the current optimizer step in order to compute $\Delta \psi_i(s)$: $\mathcal{O}(4\lvert \psi \rvert)$.


\subsection{Masking}
\label{supp:methods:masking}

Context-dependent gating (or \texttt{Masking}) is a mechanism to alleviate catastrophic interference that was introduced by \citet{masse:masking:pnas}. The method stores a random binary mask per task, which is used to gate all hidden activations. For LSTM layers, this method masks the hidden state $\mathbf{h}_t$. For vanilla RNNs, which in our case are inspired by Elman networks, \texttt{Masking} affects the hidden state $\mathbf{h}_t$ as well as the RNN layer output.\footnote{Note that for LSTMs the hidden state is also the layer output, whereas a vanilla RNN layer (an Elman network) has an additional linear readout of the hidden state. If \texttt{Masking} would only affect this readout, then there would be unhampered catastrophic interference in the crucial hidden-to-hidden computation.} Throughout all experiments, we masked 80\% of the hidden activations. Due to the independent and random generation of masks, small overlaps across tasks may occur (or if activations are computed using shared weights such as in CNNs). To prevent catastrophic interference within those overlaps, one may combine \texttt{Masking} with, for instance, \texttt{SI} (cf. Sec. \ref{supp:methods:si}). If subnetworks are sufficiently task-specific, \texttt{SI} will only influence the overlaps with subnetworks of previous tasks, without introducing rigidity for the remainder of the current subnetwork.

\paragraph{Complexity estimation.}
\texttt{Masking} does not introduce an increase in time complexity. On the contrary, if efficiently implemented, it may decrease time complexity since only activations of the active subnetwork need to be computed.

Since a binary mask per task needs to be stored, there is an increase in space complexity of $\mathcal{O}(K\lvert \psi \rvert)$. However, binary masks can be stored efficiently, as only one bit per task/activation is required.
If combined with \texttt{SI}, the space and time complexity considerations mentioned in Sec. \ref{supp:methods:si} also apply.


\subsection{Coresets}
\label{supp:methods:coresets}

\texttt{Coresets} refers to CL methods that store subsets of past data that can be mixed with new data in order to prevent catastrophic interference \citep{nguyen:2017:vcl,rebuffi2017icarl}. \citet{rebuffi2017icarl} discusses strategies on how to properly select coreset samples. Here, we simply take a random subset of $N$ input samples from each previous dataset, denoted by \texttt{Coresets-$N$}, for which we aim to keep the network predictions fixed when learning new tasks. Therefore, a copy of the network $\tilde{\psi}_i^{(K-1)}$ before learning task $K$ is generated and used to create \textit{soft-targets} $\mathbf{\tilde{y}}_{1:T_\text{out}} = f(\mathbf{x}_{1:T_\text{in}}, \tilde{\psi}_i^{(K-1)})$, where $\mathbf{x}_{1:T_\text{in}}$ is a sample taken from a coreset \citep{vandeVen2019Apr,li2017learning:without:forgetting}. The soft-targets $\mathbf{\tilde{y}}_{1:T_\text{out}}$ are distilled \cite{hinton2015distilling} into the network while training on the current task. This can be viewed as a form of regularization that incorporates past data. In addition to the current mini-batch $\mathcal{B}(s) \subseteq \mathcal{D}_K$, an additional mini-batch $\mathcal{\tilde{B}}(s)$ is assembled from inputs randomly distributed across all $K-1$ coresets together with their corresponding soft-targets. We chose to always assume that both of these mini-batches have the same size. The total loss for task $K$ can then be described as follows:

\begin{equation}
    \label{supp:eq:distill:loss}
    \min_\psi \mathcal{L}_\text{task}(\psi, \mathcal{B}(s)) + \lambda_\text{distill} \mathcal{L}_\text{distill}(\psi, \mathcal{\tilde{B}}(s))
\end{equation}

where $\lambda_\text{distill}$ is a hyperparameter and $\mathcal{L}_\text{distill}(\cdot)$ denotes the distillation loss \citep{hinton2015distilling}.

\paragraph{Complexity estimation.} The time complexity of the loss evaluation roughly doubles (the time complexities of $\mathcal{L}_\text{task}(\cdot)$ and $\mathcal{L}_\text{distill}(\cdot)$ are comparable).

Storage increases by $\mathcal{O}(\lvert \psi \rvert)$ due to the network copy $\tilde{\psi}_i^{(K-1)}$. However, the critical storage increase is due to the storage of past data, which can be summarized by $\mathcal{O}(K N F_\text{in} T_\text{in})$, assuming all samples within coresets have the same temporal dimension $T_\text{out}$.


\subsection{Generative replay}
\label{supp:methods:generative:replay}

Conceptually, \texttt{Generative Replay} \citep{shin:dgr:2017,van_de_ven:replay:through:feedback} is similar to \texttt{Coresets} (cf. Sec. \ref{supp:methods:coresets}), i.e., it is based on the rehearsal of past input data whose soft-targets are subsequently distilled into the network (cf. Eq.~\ref{supp:eq:distill:loss}). The major difference is that \texttt{Coresets} directly store past data, while \texttt{Generative Replay} relies on the ability to learn a generative model of past input data. In this study, we consider Variational Autoencoders (VAE,  \citet{kingmaW13:vae, rezende14:vae}) as generative models. We first recap the workings of a VAE on sequential data in Sec. \ref{supp:methods:generative:replay:svae} before explaining in Sec. \ref{supp:methods:generative:replay:cl:with:svae} how catastrophic interference can be mitigated in a VAE when learning a set of tasks sequentially.

\subsection{Sequential variational autoencoder}
\label{supp:methods:generative:replay:svae}

The traditional VAE (for static data) defines a generative model via marginalization of a hidden variable model: $p_\nu(\mathbf{x}) = \int_\mathcal{Z} p_\nu(\mathbf{x} \mid \mathbf{z}) p(\mathbf{z}) d\mathbf{z}$. Here, $\mathbf{z} \in \mathcal{Z}$ denotes a latent variable (or hidden cause), $p(\mathbf{z})$ is the prior and $p_\nu(\mathbf{x} \mid \mathbf{z})$ is a likelihood function defined via a decoder network whose parameters are denoted by $\nu$. To learn the parameters $\nu$ given a dataset $\mathcal{D} = \{\mathbf{x}_n\}_{n=1}^N$, the corresponding hidden causes $\mathbf{z}_n$ have to inferred from the posterior $p_\nu(\mathbf{z} \mid \mathbf{x}) \propto p_\nu(\mathbf{x} \mid \mathbf{z}) p(\mathbf{z})$. However, the precise value of the posterior is in general intractable. Therefore, VAEs resort to variational inference (VI) to approximate the posterior using $q_\psi(\mathbf{z} \mid \mathbf{x}) \approx p_\nu(\mathbf{z} \mid \mathbf{x})$, where $q_\psi(\mathbf{z} \mid \mathbf{x})$ is realized through an encoder network with parameters $\psi$. VI utilizes the following inequality (cf. \cite{kingmaW13:vae} for a derivation):

\begin{equation}
    \label{supp:eq:elbo}
    \log p_\nu(\mathbf{x}) \geq - KL \big(q_\psi(\mathbf{z} \mid \mathbf{x}) \,||\, p(\mathbf{z}) \big) + \mathbb{E}_{q_\psi(\mathbf{z} \mid \mathbf{x})} \big[ \log p_\nu(\mathbf{x} \mid \mathbf{z}) \big]
\end{equation}

where the right-hand side is commonly known as evidence lower bound (ELBO). VAE training proceeds by maximizing the ELBO or equivalently by minimizing the negative ELBO which decomposes into a \textit{prior-matching term} $KL \big(q_\psi(\mathbf{z} \mid \mathbf{x}) \,||\, p(\mathbf{z}) \big)$ and a \textit{negative log-likelihood} (NLL) term $-\mathbb{E}_{q_\psi(\mathbf{z} \mid \mathbf{x})} \big[ \log p_\nu(\mathbf{x} \mid \mathbf{z}) \big]$.

Next, we discuss how to extend this framework to sequential data (also cf. \cite{bengio:sequential:vae, storn:sequential:vae}).
We use an independence assumption when defining a prior for a sequence of hidden causes:

\begin{equation}
    p(\mathbf{z}_{1:T}) = \prod_t p(\mathbf{z}_t)
\end{equation}

In addition, we consider the following decomposition of the likelihood function:

\begin{equation}
    p_\nu(\mathbf{x}_{1:T} \mid \mathbf{z}_{1:T}) = \prod_t p_\nu(\mathbf{x}_t \mid \mathbf{x}_{<t}, \mathbf{z}_{\leq t})
\end{equation}

The decoder network is an RNN defined via
$[\varphi_t, \mathbf{h}^\text{dec}_t] = f_\text{dec,step}(\mathbf{z}_t, \mathbf{h}^\text{dec}_{t-1}, \nu)$, where $\mathbf{h}^\text{dec}_t$ denotes the hidden state of the decoder network and $\varphi_t \in \Phi$ denotes the parameters of a parametric distribution (e.g., a Gaussian), which can be used to tractably compute densities $p_\nu(\mathbf{x}_t \mid \mathbf{x}_{<t}, \mathbf{z}_{\leq t})$ conditioned on $\mathbf{z}_{1:T}$.

As a last ingredient, we have to define the recognition model $q_\psi(\mathbf{z}_{1:T} \mid \mathbf{x}_{1:T})$. If the prior and likelihood defined above are inserted into Bayes rule, there is no obvious way to simplify the dependency structure of the true posterior such that the autoregressive nature of an RNN recognition model is not violated. We therefore apply an additional assumption when defining the decomposition applied to our recognition model:

\begin{equation}
    \label{supp:eq:seq:rec:model}
    q_\psi(\mathbf{z}_{1:T} \mid \mathbf{x}_{1:T}) \stackrel{\text{chain rule of prob.}}{=}
    \prod_t q_\psi(\mathbf{z}_t \mid \mathbf{z}_{<t}, \mathbf{x}_{1:T})
    \stackrel{\text{filtering assumption}}{\approx}
    \prod_t q_\psi(\mathbf{z}_t \mid \mathbf{z}_{<t}, \mathbf{x}_{\leq t})
\end{equation}

Analogously to the likelihood, the components $q_\psi(\mathbf{z}_t \mid \mathbf{z}_{<t}, \mathbf{x}_{\leq t})$ of the approximate posterior are represented by an RNN encoder network 
$[\xi_t, \mathbf{h}^\text{enc}_t] = f_\text{enc,step}(\mathbf{x}_t, \mathbf{h}^\text{enc}_{t-1}, \psi)$, where $\xi_t \in \Xi$ are the parameters of a distribution over the latent space $\mathcal{Z}$.

At this point, we have all ingredients of the ELBO (cf. Eq.~\ref{supp:eq:elbo}) defined and can now focus our discussion on how to tractably evaluate the ELBO for the case of sequential data. We will start with decomposing the prior-matching term:

\begin{align}
   & KL \big( q_\psi(\mathbf{z}_{1:T} \mid \mathbf{x}_{1:T}) \,||\, p(\mathbf{z}_{1:T}) \big) \nonumber \\
   =& \int_{\mathbf{z}_{1:T}} \prod_{t'} q_\psi(\mathbf{z}_{t'} \mid \mathbf{z}_{<t'}, \mathbf{x}_{\leq t'}) \sum_t \log \frac{q_\psi(\mathbf{z}_t \mid \mathbf{z}_{<t}, \mathbf{x}_{\leq t})}{ p(\mathbf{z}_t)} \, d\mathbf{z}_{1:T} \nonumber \\
   =& \sum_t \int_{\mathbf{z}_{1:T}} \prod_{t'} q_\psi(\mathbf{z}_{t'} \mid \mathbf{z}_{<t'}, \mathbf{x}_{\leq t'}) \log \frac{q_\psi(\mathbf{z}_t \mid \mathbf{z}_{<t}, \mathbf{x}_{\leq t})}{ p(\mathbf{z}_t)} \, d\mathbf{z}_{1:T} \nonumber \\
   =& \sum_t \int_{\mathbf{z}_{1:t}} \prod_{t' \leq t} q_\psi(\mathbf{z}_{t'} \mid \mathbf{z}_{<t'}, \mathbf{x}_{\leq t'}) \log \frac{q_\psi(\mathbf{z}_t \mid \mathbf{z}_{<t}, \mathbf{x}_{\leq t})}{ p(\mathbf{z}_t)} \, d\mathbf{z}_{1:t}
\end{align}

Note that the last manipulation is possible since the log-ratio does not depend on $\mathbf{z}_{t'}$ when $t' > t$ and, therefore, the log-ratio can be moved outside the respective integrals which evaluate to 1. We can further simplify the expression as follows: 

\begin{align}
   & KL \big( q_\psi(\mathbf{z}_{1:T} \mid \mathbf{x}_{1:T}) \,||\, p(\mathbf{z}_{1:T}) \big) \nonumber \\
   =& \sum_t \int_{\mathbf{z}_{1:t-1}} \prod_{t' < t} q_\psi(\mathbf{z}_{t'} \mid \mathbf{z}_{<t'}, \mathbf{x}_{\leq t'}) \int_{\mathbf{z}_t} q_\psi(\mathbf{z}_{t} \mid \mathbf{z}_{<t}, \mathbf{x}_{\leq t}) \log \frac{q_\psi(\mathbf{z}_t \mid \mathbf{z}_{<t}, \mathbf{x}_{\leq t})}{ p(\mathbf{z}_t)} \, d\mathbf{z}_t \, d\mathbf{z}_{1:t-1} \nonumber \\
   =& \sum_t \int_{\mathbf{z}_{1:t-1}} \prod_{t' < t} q_\psi(\mathbf{z}_{t'} \mid \mathbf{z}_{<t'}, \mathbf{x}_{\leq t'})  KL \big( q_\psi(\mathbf{z}_t \mid \mathbf{z}_{<t}, \mathbf{x}_{\leq t}) \,||\, p(\mathbf{z}_t) \big) \, d\mathbf{z}_{1:t-1} \nonumber \\
   =& \sum_t \int_{\mathbf{z}_1} q_\psi(\mathbf{z}_{1} \mid \mathbf{x}_1) \int_{\mathbf{z}_2} \dots  \int_{\mathbf{z}_{t-1}} q_\psi(\mathbf{z}_{t-1} \mid \mathbf{z}_{<t-1}, \mathbf{x}_{\leq t-1})  KL (\dots) \, d\mathbf{z}_{t-1} \dots d\mathbf{z}_1
   \label{supp:eq:pm:term:sequential:data}
\end{align}

Note that the KL divergence term $KL \big( q_\psi(\mathbf{z}_t \mid \mathbf{z}_{<t}, \mathbf{x}_{\leq t}) \,||\, p(\mathbf{z}_t) \big)$ in Eq.~\ref{supp:eq:pm:term:sequential:data} is analytically solvable based on a proper choice of prior and likelihood. The surrounding integrals can be estimated via Monte-Carlo (MC) sampling. In the simplest case, they are estimated by taking one sample per integral, i.e., given an input sequence $\mathbf{x}_{1:T}$, we use the recognition model $[\xi_t, \mathbf{h}^\text{enc}_t] = f_\text{enc,step}(\mathbf{x}_t, \mathbf{h}^\text{enc}_{t-1}, \psi)$ to compute a latent sequence $\mathbf{z}_{1:T}$ via $\mathbf{z}_t \sim q_\psi(\mathbf{z}_t \mid \mathbf{z}_{<t}, \mathbf{x}_{\leq t}) \Leftrightarrow \mathbf{z}_t \sim p_{\xi_t}(\mathbf{z}_t)$, where $p_{\xi_t}(\cdot)$ is an explicit parametric distribution that we chose for the latent space (typically Gaussian), to evaluate the KL term. Note, $\xi_t$ depends on $\mathbf{x}_t$ and $\mathbf{h}^\text{enc}_{t-1}$. However, in the implementation that we chose for this study, $\mathbf{h}^\text{enc}_{t-1}$ does not explicitly depend on $\mathbf{z}_{<t}$ (only implicitly through its distribution determined by $\xi_t$) even though $q_\psi(\mathbf{z}_t \mid \mathbf{z}_{<t}, \mathbf{x}_{\leq t})$ requires an explicit dependency.\footnote{This limitation could be overcome if the RNN definition would be slightly adapted. For instance, if the definition of the encoder would change to $[\xi_t, \mathbf{h}^\text{enc}_t] = f_\text{enc,step}(\mathbf{x}_t, \mathbf{z}_{t-1}, \mathbf{h}^\text{enc}_{t-1}, \psi)$ with $\mathbf{z}_{t-1} \sim p_{\xi_{t-1}}(\mathbf{z}_{t-1})$.}

Taken together, we approximate the prior-matching term as follows:

\begin{equation}
    KL \big( q_\psi(\mathbf{z}_{1:T} \mid \mathbf{x}_{1:T}) \,||\, p(\mathbf{z}_{1:T}) \big)
   \approx \sum_t KL \big( p_{\xi_t}(\mathbf{z}_t) \,||\, p(\mathbf{z}_t) \big)
   \label{supp:eq:vae:pm:approx}
\end{equation}

Similarly, we can handle the negative log-likelihood (\textbf{NLL}):

\begin{align}
    \text{NLL} &= - \mathbb{E}_{q_\psi(\mathbf{z}_{1:T} \mid \mathbf{x}_{1:T})} \big[ \log p_\nu(\mathbf{x}_{1:T} \mid \mathbf{z}_{1:T}) \big]  \nonumber \\
    &= - \int_{\mathbf{z}_{1:T}} \prod_{t'} q_\psi(\mathbf{z}_{t'} \mid \mathbf{z}_{<t'}, \mathbf{x}_{\leq t'}) \sum_t \log p_\nu(\mathbf{x}_t \mid \mathbf{x}_{<t}, \mathbf{z}_{\leq t}) \, d\mathbf{z}_{1:T} \nonumber \\
    &= - \sum_t \int_{\mathbf{z}_{1:t}} \prod_{t' \leq t} q_\psi(\mathbf{z}_{t'} \mid \mathbf{z}_{<t'}, \mathbf{x}_{\leq t'})  \log p_\nu(\mathbf{x}_t \mid \mathbf{x}_{<t}, \mathbf{z}_{\leq t}) \, d\mathbf{z}_{1:t} \nonumber \\
    &= - \sum_t \int_{\mathbf{z}_1} q_\psi(\mathbf{z}_{1} \mid \mathbf{x}_1) \dots \int_{\mathbf{z}_{t}} q_\psi(\mathbf{z}_{t} \mid \mathbf{z}_{<t}, \mathbf{x}_{\leq t})  \log p_\nu(\mathbf{x}_t \mid \mathbf{x}_{<t}, \mathbf{z}_{\leq t}) \, d\mathbf{z}_{t} \dots d\mathbf{z}_1 \nonumber \\
    &\stackrel{\text{MC sample size of 1}}{\approx} - \sum_t \log p_\nu(\mathbf{x}_t \mid \mathbf{x}_{<t}, \mathbf{z}_{\leq t}) \label{supp:eq:vae:nll:approx}
\end{align}

If $p_\nu(\mathbf{x}_t \mid \mathbf{x}_{<t}, \mathbf{z}_{\leq t})$ is a Gaussian distribution (which we assume for the SMNIST and AudioSet experiments), Eq.~\ref{supp:eq:vae:nll:approx} becomes a sum over mean-squared error (MSE) losses (after dropping constant terms and assuming the covariance matrix to be a scaled identity matrix $\tau^{-1}I$). Thus, we assume the output $\varphi_t$ of the decoder $[\varphi_t, \mathbf{h}^\text{dec}_t] = f_\text{dec,step}(\mathbf{z}_t, \mathbf{h}^\text{dec}_{t-1}, \nu)$ is the mean of a Gaussian distribution $\mathcal{N}(\mathbf{x}_t; \varphi_t, \tau^{-1}I)$, therefore $\mathcal{X} \equiv \Phi$. One could sample reconstructions from this distribution using the reparametrization trick \citep{kingmaW13:vae}. However, at this level we do not introduce additional noise and instead aim to match encoder input $\mathbf{x}_t$ and decoder output $\varphi_t$ directly:\footnote{Note, in contrast to the approximate posterior distribution $q_\psi(\mathbf{z}_t \mid \mathbf{z}_{<t}, \mathbf{x}_{\leq t})$ (which we crucially require to replay samples of prior tasks), we only require a sensible mean of the likelihood $p_\nu(\mathbf{x}_t \mid \mathbf{x}_{<t}, \mathbf{z}_{\leq t})$ to represent reconstructions.}

\begin{equation}
    \text{NLL} \approx \sum_{t=1}^{T_\text{in}} \frac{\tau}{2} \lVert \mathbf{x}_t - \varphi_t \rVert^2
\end{equation}

In case of the Copy Task (and its variants), it makes sense to choose $p_\nu(\mathbf{x}_t \mid \mathbf{x}_{<t}, \mathbf{z}_{\leq t})$ to be a Bernoulli distribution (assuming the raw decoder output $\varphi_t$ has been squeezed through a sigmoid):

\begin{equation}
    \text{NLL} \approx \sum_{t=1}^{T_\text{in}} \sum_{f=1}^{F_\text{in}} - x_{t,f} \log \varphi_{t,f} - (1-x_{t,f}) \log (1-\varphi_{t,f})
\end{equation}

\subsection{Generative replay using a sequential VAE}
\label{supp:methods:generative:replay:cl:with:svae}

Above, we describe how we train a VAE on sequential data. In order to use it as a generative model for CL, we have to employ strategies that mitigate catastrophic interference when training consecutively on multiple tasks. We therefore explore two strategies inspired by related work on static data, \texttt{RtF} \citep{van_de_ven:replay:through:feedback} (referred to as \texttt{Generative Replay} in the main text) and \texttt{HNET+R} \citep{oswald:hypercl}. In both cases, we use the main model simultaneously as classifier and VAE encoder $[\mathbf{\hat{y}}_t, \xi_t, \mathbf{h}_t] = f_\text{step}(\mathbf{x}_t, \mathbf{h}_{t-1}, \psi)$, where $\mathbf{\hat{y}}_t$ remains the model's prediction (cf. Sec. \ref{supp:notation}) and $\xi_t$ encodes a mean and diagonal covariance matrix of a Gaussian distribution used to sample latent representations of the VAE $\mathbf{z}_t \sim p_{\xi_t}(\mathbf{z}_t)$.

\texttt{RtF} \citep{van_de_ven:replay:through:feedback} refers to training the VAE on data from all tasks seen so far. However, in a CL setting, data from previous tasks is not available. Therefore, a checkpointed decoder $\tilde{\nu}_i^{(K-1)}$ is used to replay data from tasks $1$ to $K-1$ while training on task $K$. In summary, similar to \texttt{Coresets} (cf. Sec. \ref{supp:methods:coresets}), a mini-batch $\mathcal{B}(s)$ with data from task $K$ and a mini-batch $\mathcal{\tilde{B}}(s)$ with replayed data (using $\tilde{\nu}_i^{(K-1)}$) from tasks $1$ to $K-1$ is assembled. In addition to the distillation loss (cf. Eq.~\ref{supp:eq:distill:loss}; using hyperparameter $\lambda_\text{distill}$), that only affects the encoder $f_\text{step}$, the reconstruction loss $\mathcal{L}_\text{rec}(\cdot)$ (cf. Eq.~\ref{supp:eq:vae:nll:approx}) and prior-matching loss $\mathcal{L}_\text{pm}(\cdot)$ (cf. Eq.~\ref{supp:eq:vae:pm:approx}) are evaluated on $\mathcal{B}(s)$ and $\mathcal{\tilde{B}}(s)$:

\begin{align}
    \min_{\psi,\nu} \, &\mathcal{L}_\text{task}(\psi, \mathcal{B}(s)) + \lambda_\text{distill} \mathcal{L}_\text{distill}(\psi, \mathcal{\tilde{B}}(s)) +
    \lambda_\text{rec} \mathcal{L}_\text{rec}\big(\psi,\nu, \mathcal{B}(s) \cup \mathcal{\tilde{B}}(s)\big) \nonumber \\
    &+ \lambda_\text{pm} \mathcal{L}_\text{pm}\big(\psi,\nu, \mathcal{B}(s) \cup \mathcal{\tilde{B}}(s)\big)
    \label{supp:eq:rtf:loss}
\end{align}

where $lambda_\text{rec}$ and $\lambda_\text{pm}$ denote two new hyperparameters. Note, in order to train a multi-head main network $f_\text{step}$ with replayed data, the output head (task identity) of replayed data has to be known. To achieve this, task identity has to be provided as a one-hot encoding to the decoder in addition to the latent variable $\mathbf{z}_t$.

The generative model used by \texttt{RtF} is therefore continuously retrained on its own replayed data. Hence, distributional shifts and mismatches accumulate over time, leading to a decrease in quality of replayed samples \citep{oswald:hypercl}. The method \texttt{HNET+R} \citep{oswald:hypercl} circumvents this problem of \texttt{RtF} by training a task-specific decoder, where decoders of previous tasks are protected by a hypernetwork (cf. Eq.~\ref{eq:hnet:reg}) and only the current task's decoder is trained on actual data. To do so, a hypernetwork is introduced for the decoder (and not the main network) $\nu = h_\text{dec}(\mathbf{e}_k^\text{dec}, \theta^\text{dec})$. The loss in this case becomes (cf. Eq.~\ref{eq:hnet:reg} ad Eq.~\ref{supp:eq:rtf:loss}):

\begin{align}
    \min_{\psi,\theta^\text{dec}} \, &\mathcal{L}_\text{task}(\psi, \mathcal{B}(s)) + \lambda_\text{distill} \mathcal{L}_\text{distill}(\psi, \mathcal{\tilde{B}}(s)) +
    \lambda_\text{rec} \mathcal{L}_\text{rec}\big(\psi,\theta^\text{dec}, \mathcal{B}(s)\big) \nonumber \\
    &+ \lambda_\text{pm} \mathcal{L}_\text{pm}\big(\psi,\theta^\text{dec}, \mathcal{B}(s)\big) \nonumber \\
    &+ \frac{\beta_\text{dec}}{K-1} \sum_{k=1}^{K-1} \lVert  h_\text{dec}(\mathbf{e}_k^\text{dec}, \theta^\text{dec}) - 
    h_\text{dec}(\mathbf{\tilde{e}}_k^{(\text{dec}, K-1)}, \tilde{\theta}^{(\text{dec}, K-1)}) \rVert_2^2
    \label{supp:eq:hnet:replay:loss}
\end{align}

where $\beta_\text{dec}$, $\mathbf{\tilde{e}}_k^{(\text{dec}, K-1)}$, $\tilde{\theta}^{(\text{dec}, K-1)}$ are defined for the decoder hypernetwork $h_\text{dec}$ analogously as described for the main network's hypernetwork in Sec.~\ref{sec:methods}.

\paragraph{Complexity estimation.} \texttt{RtF} and \texttt{HNET+R} are affected from the same complexity considerations as \texttt{Coresets} (cf. Sec. \ref{supp:methods:coresets}) except for storing past data (which are instead replayed from a checkpointed decoder $f_\text{step,dec}$ resp. decoder-hypernetwork $h_\text{dec}$). Method \texttt{HNET+R} has the additional complexity increases mentioned in Sec. \ref{supp:methods:hnets}. Both methods require maintaining an additional decoder network. Both also require the evaluation of two extra loss terms, $\mathcal{L}_\text{rec}(\cdot)$ and $\mathcal{L}_\text{pm}(\cdot)$, whereas this cost is doubled for \texttt{RtF} as it always evaluates these terms on current and replayed data.

\section{A theoretical view on CL in linear RNNs}
\label{supp:linear:rnns}

In this section, we provide theoretical insights on why  high working memory requirements might be problematic when weight-importance methods such as EWC are applied to RNNs.

We empirically showed in Fig. \ref{fig:copy_intuition} that increasing pattern lengths $p$ of Copy Task inputs lead to increasing weight-importance values as calculated by EWC. We also observed that higher working memory requirements (resulting from increasing pattern length) force the networks to utilize more of their capacity, which leads to a higher intrinsic dimensionality of the hidden state. These observations led us to the prediction that high working memory requirements can lead to a saturation of weight-importance values, thus decreasing performance of methods such as EWC when sequentially learning multiple tasks.

Here, we examine these statements from a theoretical perspective for the case of linear RNNs. More specifically, we explore why using a shared set of recurrent weights for several tasks can be problematic when the intrinsic dimensionality of the hidden state increases. Note that this framework is therefore applicable to any method that uses a single set of recurrent weights for several tasks, no matter whether these are learned sequentially or not (which includes weight-importance methods but also, for instance, replay methods and the multitask setting).

\paragraph{Model.}
We consider a linear RNN with one recurrent hidden layer $\mathbf{h}_t$ of dimension $n_h$. The dynamics of the network are defined as follows:
\begin{align}
    \mathbf{h}_t &= W_{hh} \mathbf{h}_{t-1} + W_{xh}\mathbf{x}_t 
    \label{supp:eq:linear:rnn}\\
    \hat{\mathbf{y}}_t &= W_{hy} \mathbf{h}_t
\end{align}
with $W_{hh}$, $W_{xh}$ and $W_{hy}$ weight matrices.
We consider a setting in which the network has to learn $K$ different tasks using the shared weights $W_{hh}$ and $W_{xh}$, and a set of task-specific output heads $W_{hy}^{(k)}$.
We denote by $\mathbf{h}_t^{(k)} \in \mathbb{R}^{n_h}$ the content of $\mathbf{h}_t$ that is utilized for the task-specific processing of task $k$.

\paragraph{Task.}
We consider a variant of the Copy Task, in which at timesteps $t=1:p$ the network needs to output a manipulated copy of the network inputs at timesteps $t=-p:-1$. The input $\mathbf{x}_t$ is zero for $t>0$, and the specific manipulation of the input is different for all $K$ tasks.

\paragraph{Simplifying assumptions.}
To make the analysis as clear as possible we make the following simplifying assumptions:
\begin{enumerate}
    \item Task-specific recurrent processing on $\mathbf{h}_t$ via $W_{hh}$ is still required for $t>0$ in order to solve task $k$ (i.e. the task-specific output heads $W_{hy}^{(k)}$ are not rich enough to model all task variabilities).
    \item Each task $k$ needs a completely distinct processing mechanism from other tasks. There is thus no possibility of transfer-learning across tasks, and if the processing of $\mathbf{h}_t^{(k)}$ by $W_{hh}$ overlaps with the processing of $\mathbf{h}_t^{(l \neq k)}$, the two tasks will interfere with each other, leading to a drop in performance.
\end{enumerate}

\paragraph{Theoretical analysis of the linear toy problem.}
Our PCA analyses show that the hidden state $\mathbf{h}_t$ is embedded in a lower-dimensional linear subspace of $\mathbb{R}^{n_h}$.
Based on the above simplifying assumptions, the only way for a linear RNN to ensure a task-specific processing is that $\mathbf{h}_t^{(k)}$, the information within $\mathbf{h}_t$ relevant for solving task $k$, populates distinct and non-overlapping linear subspaces of $\mathbb{R}^{n_h}$ for each task across all $t>0$: 
\begin{align}
    \mathbf{h}_t^{(k)} &\in \mathcal{S}_{k}\\
    \mathcal{S}_{k} \cap \mathcal{S}_{l\neq k} &= \{\mathbf{0}\}
\end{align}

If this wasn't the case and $\mathcal{S}_k$ overlapped with other subspaces, $\mathbf{h}^{(k)}_{t}$ could have components in $\mathcal{S}_{l \neq k}$, which would be influenced by the task-specific processing of other tasks. 

Because $W_{hh}$ is sequentially applied to $\mathbf{h}_{t}$, it must perform a subspace-retaining operation on $\mathbf{h}_t^{(k)}  \in \mathcal{S}_k$ such that:
\begin{align}
    \mathbf{h}^{(k)}_{t+1} &= W_{hh} \mathbf{h}^{(k)}_{t}\\
    \mathbf{h}^{(k)}_{t+1} &\in \mathcal{S}_k
\end{align}

The task-specific output head $W_{hy}^{(k)}$ can then select a linear subspace $S_k$ of $\mathbf{h}_t$ that serves as output for task $k$. Hence, as long as the $W_{hh}$ can represent a task-specific and subspace-retaining operation on $\mathbf{h}_t^{(k)}$, it is possible for the RNN to represent $K$ different tasks that do not interfere with each other, and the task-specific output head $W_{hy}^{(k)}$ can select the appropriate subspace of $\mathbf{h}_t$ to present at the output.

In the following, we show that it is possible to have such task specific processing of the hidden-state vectors by $W_{hh}$ if the subspaces $\mathcal{S}_k$ are orthogonal to each other and if the intrinsic dimensionality of task-relevant information within the hidden space is sufficiently small across tasks.
We use this finding as an intuition to justify why increasing intrinsic dimensionality of the hidden state can lead to interference across tasks when a single matrix $W_{hh}$ is used.

Let's represent each subspace $\mathcal{S}_k$ by the column space of a matrix $U_k$ with orthonormal columns. Because $\mathbf{h}_t^{(k)}$ only has components in $\mathcal{S}_k$, it can be written as:
\begin{align}
    \mathbf{h}_t^{(k)} = U_k \mathbf{c}_t^{(k)}
\end{align}
with $\mathbf{c}_t^{(k)} \in \mathbb{R}^{p_k}$ the coordinates of $\mathbf{h}_t^{(k)}$ in the basis $U_k$.
If subspaces are orthogonal and $\sum_k p_k \leq n_h$, we can state that: $U_k$ is orthogonal to all other $U_{l \neq k}$, and that $\bar{U} = \left[ U_1 ... U_K \tilde{U} \right]^{n_h \times n_h}$ is an orthogonal basis for $\mathbb{R}^{n_h}$, with $\tilde{U}$ orthogonal to all $U_k$.

Now we can define $Q = \bar{U}^T W_{hh} \bar{U}$, a change of basis of $W_{hh}$ under $\bar{U}$. $Q$ can be structured in the following blocks:
\begin{align}
    Q = \begin{bmatrix}
Q_{11}  & Q_{12}      &  \hdots     &Q_{1 \sim}   \\
Q_{21}   & \ddots &      & \vdots\\
\vdots &  & \ddots & Q_{K\sim}    \\
Q_{\sim 1}  & \hdots  & Q_{\sim K}       & Q_{\sim \sim} 
\end{bmatrix}
\end{align}

where $Q_{ij}$ corresponds to the computation within $W_{hh}$ that leads a subspace transformation from $S_j$ to $S_i$.
Then, $\mathbf{h}_{t+1}^{(k)}$ is given by
\begin{align}
    \mathbf{h}_{t+1}^{(k)} &= W_{hh} \mathbf{h}_t^{(k)} \\
    &= W_{hh} U_k \mathbf{c}_t^{(k)} \\
    &= \bar{U} Q \bar{U}^T U_k \mathbf{c}_t^{(k)} \\
    &= \sum_{l=1}^K U_l Q_{lk}\mathbf{c}_t^{(k)} + \tilde{U} Q_{\sim k}\mathbf{c}_t^{(k)}
\end{align}

We can easily see that, if $Q_{ij}=\mathbf{0}$ for $i \neq j$, we obtain:
\begin{align}
    \mathbf{h}_{t+1}^{(k)} =  U_k Q_{kk}\mathbf{c}_t^{(k)}
\end{align}

Therefore, one can easily design $Q$ in such a way that $W_{hh}$ performs a subspace-retaining transformation on $\mathbf{h}_t^{(k)}$, i.e. $Q$ needs to have a block diagonal structure.
Otherwise, $U_l Q_{lk}\mathbf{c}_t^{(k)}$ for $l \neq k$ is non-zero, and $\mathbf{h}_{t+1}^{(k)}$ will contain components in $\mathcal{S}_{l \neq k}$. 

To summarize, we see that it is possible for the RNN to have a task-specific processing of the hidden-state vector for each task, without interfering with the other tasks, as long as $\sum_k p_k \leq n_h$. If $\sum_k p_k > n_h$, it is not possible anymore to have $K$ orthogonal linear subspaces $\mathcal{S}_k$, which can lead to interference between tasks and a resulting drop in performance.

\paragraph{Implications for CL.} We showed that it is possible to build a linear RNN that doesn't suffer from interference across tasks despite using a single set of recurrent weights, as long as the intrinsic dimensionality of the hidden space is not too large. This observation has clear implications for weight-importance methods in CL, which progressively restrict the plasticity of  a single set of recurrent weights when sequentially learning different tasks.
Theoretically, weight-importance methods can encourage task-relevant information of the hidden state to be encoded in orthogonal subspaces, such that the learning of new tasks does not interfere with the previously learned tasks. However, if the subspace dimensionality $p_k$ increases (e.g., for increasing pattern lengths in the Copy Task) or if the number of tasks is too large, leading to $\sum_k p_k > n_h$, the various tasks will start interfering with each other, and the performance will drop. Even though we consider a simplified scenario where the recurrent processing cannot be shared across tasks, most commonly tasks will benefit from some form of shared processing. Crucially, whenever the subspaces associated with individual tasks may overlap, the overall dimensionality of the used hidden space can be less than the sum of dimensionalities of individual task-related subspaces. This frees up capacity in the recurrent weights to learn new tasks. Therefore, together with the working memory of individual tasks and the number of tasks, task similarity will also play a role in the effectiveness of weight-importance methods for RNNs.
Interestingly, assuming it translates to nonlinear RNNs, one can use this intuition to design CL methods that avoid interference between tasks, but use shared computation whenever possible, as demonstrated by concurrent work \citep{duncker2020organizing}.

\paragraph{Theoretical benefits of hypernetworks for CL.} 
Following the above analysis, we conclude that hypernetworks provide a theoretical advantage over weight-importance methods. With hypernetworks, a task-specific $W_{hh}^{(k)}$ can be generated for every new task, without forgetting $W_{hh}^{(l)}$ of previous tasks $l < k$. Hence, because $W_{hh}$ does not need to represent subspace-retaining operations, a hypernetwork-based CL approach exhibits more flexibility for mitigating the stability-plasticity dilemma.

\section{Further Discussion of Related Work}
\label{supp:related:work}

The literature contains a number of studies that touch upon the problem of retaining and transferring past knowledge when dealing with sequential data. Some of these studies do not directly address the problem of catastrophic forgetting in the form of retaining performance on old tasks while learning new tasks, or they are not applicable to the experimental settings investigated in this study. The purpose of this section is to provide a brief overview over these studies and contrast them to the subject of this work.

Nowadays, real-world applications of RNNs are almost exclusively trained using backpropagation-through-time (BPTT) as underlying optimization algorithm. Even though most CL approaches are by design agnostic to this choice, they are typically tested and benchmarked against each other using BPTT. The study at hand is no exception in that regard.
In contrast, recent work by \citet{ororbia2020continual} has developed a biologically-inspired alternative to BPTT that enables zero-shot adaptation in a range of sequential generative modelling tasks. They perform \texttt{Fine-Tuning} experiments for a set of optimization techniques without utilizing explicit mechanisms supporting CL and show that the choice of optimization algorithm can lead to slight differences in terms of vulnerability to catastrophic forgetting. However, the applicability of this method outside the domain of token-level generative modelling remains unclear since, for example, it is not straightforward to adapt to classification problems.

\citet{Li2020Compositional:Language:CL} consider a specific sequential CL setting where it is assumed that the type of recurrent computation to be processed across tasks is identical. In their particular case, the recurrent component is frozen after the first task, i.e., recurrent computation is shared across tasks, and therefore catastrophic forgetting only needs to be prevented in the feedforward component of the architecture using an existing approach such as EWC \cite{kirkpatrick:ewc:2017}.

The work from \citet{philps2019making:good:unfulfilled:promise} focuses on financial time-series data. Their approach to CL is memory-based, where tuples of a representation of the training data and the model parameters are stored. Predictions can then be obtained via a weighted average of existing models, where a model is given more weight based on how similar its data representation is to the currently observed data. Such an approach is arguably expensive in terms of memory but also in terms of computation during inference, as final predictions are obtained by a weighted average of all models in the memory bank.

Learning without task-boundaries is another interesting challenge of CL and has been explored in the context of NLP tasks by \citet{kruszewski2020class:agnostic:cl}. They provide datasets consisting of either four or five distinct tasks, which are randomly repeated such that the network observes 100 fragments, each consisting of samples from a single task. Their proposed method consists of an ensemble of models whose predictions are averaged using learned weighting factors. The updates of those weighting factors are faster than those of the ensemble members, implicitly allowing a mechanism for fast remembering (cf. \citet{he:2019:task:agnostic:cl}) and therewith avoiding that ensemble members are completely overwritten upon task switches, which mitigates catastrophic interference. Note, the CL methods studied in our work can be adapted to an experimental setting where task-boundaries are not provided during training by either monitoring outliers in the loss or the model's predictive uncertainty (assuming tasks with non-overlapping input data distributions).

An approach for efficient knowledge transfer using ideas from CL for sentiment classification has been proposed by \citet{lv2019sentiment:classification}. This approach contains two subnetworks, one that has an explicit mechanism against catastrophic forgetting (using a soft-masking approach) and a network that has full flexibility to adapt to the task at hand. The predictions of these two subnetworks are combined when forming a final decision.

Several studies have used methodologies from the CL literature for solving NLP problems \citep{wolf2018hierarchical:multiscale, madasu2020sequential, Thompson2019Jun}. For instance, \citet{wolf2018hierarchical:multiscale} consider language modelling and employ a meta-model that can update the parameters of the language model such that it can focus on the local context. However, this approach is vulnerable to loosing the overall, pretrained language modelling skills, and therefore benefits from EWC to constrain the updates made by the meta-model.

\section{Datasets and tasks}
\label{supp:data}

Here we provide details on the datasets and tasks used in this study. All details on preprocessing or generating data, as well as links for downloading the precise datasets can also be found in the accompanied code repository.

\vspace{3mm} 
\begin{table*}[ht]
 \centering
  \caption{Summary of the data used to train and evaluate one subtask for each of the four datasets. $i$ and $p$ refer to the input sequence and pattern lengths of the Copy Task, $m$ refers to the number of digits in a SMNIST sequence. For the PoS dataset we report mean and standard deviation over tasks; the input feature size is given by the size of the word embeddings.}
  \begin{small}
  \begin{tabular}{lcccc} 
    & \textbf{Copy Task Variants} & \textbf{SMNIST} & \textbf{AudioSet} & \textbf{PoS} \\ \toprule
    \textbf{Classes} & N/A  & 2 & 10 & 17 \\ 
    \textbf{Training samples} & 100000  & 2 * 6000 & 10 * 750 & 9582 $\pm$ 4962 \\ 
    \textbf{Validation samples} & 1000  & 2 * 500 & 10 * 50 & 1492 $\pm$ 1875\\ 
    \textbf{Test samples} & 1000  & 2 * 1000 & 10 * 200 & 1467 $\pm$ 2066\\ 
    \textbf{Input feature size} & 8  & 4 & 128 & 64\\ 
    \textbf{Number of timesteps} & $i+1+p$ & 117*$m$ & 10 & 20 $\pm$ 6 \\ 
  \end{tabular}
  \label{supp:tab:datasets}
  \end{small}
\end{table*}


\subsection{Variations of the Copy Task}
\label{supp:data:copy}

The Copy Task \citep{graves2014ntm} is a synthetic dataset that we use to investigate different aspects of CL with sequential data.
In this section, we first explain the basic Copy Task, and subsequently give details about the different manipulations we introduced to create variations of this task.
For all variants, we used the training / validation / testing scheme described in Table \ref{supp:tab:datasets}.

\subsubsection{Basic Copy Task}
\label{supp:data:copy:basic}

\begin{figure}[ht]
    \centering
    \includegraphics[width=\textwidth]{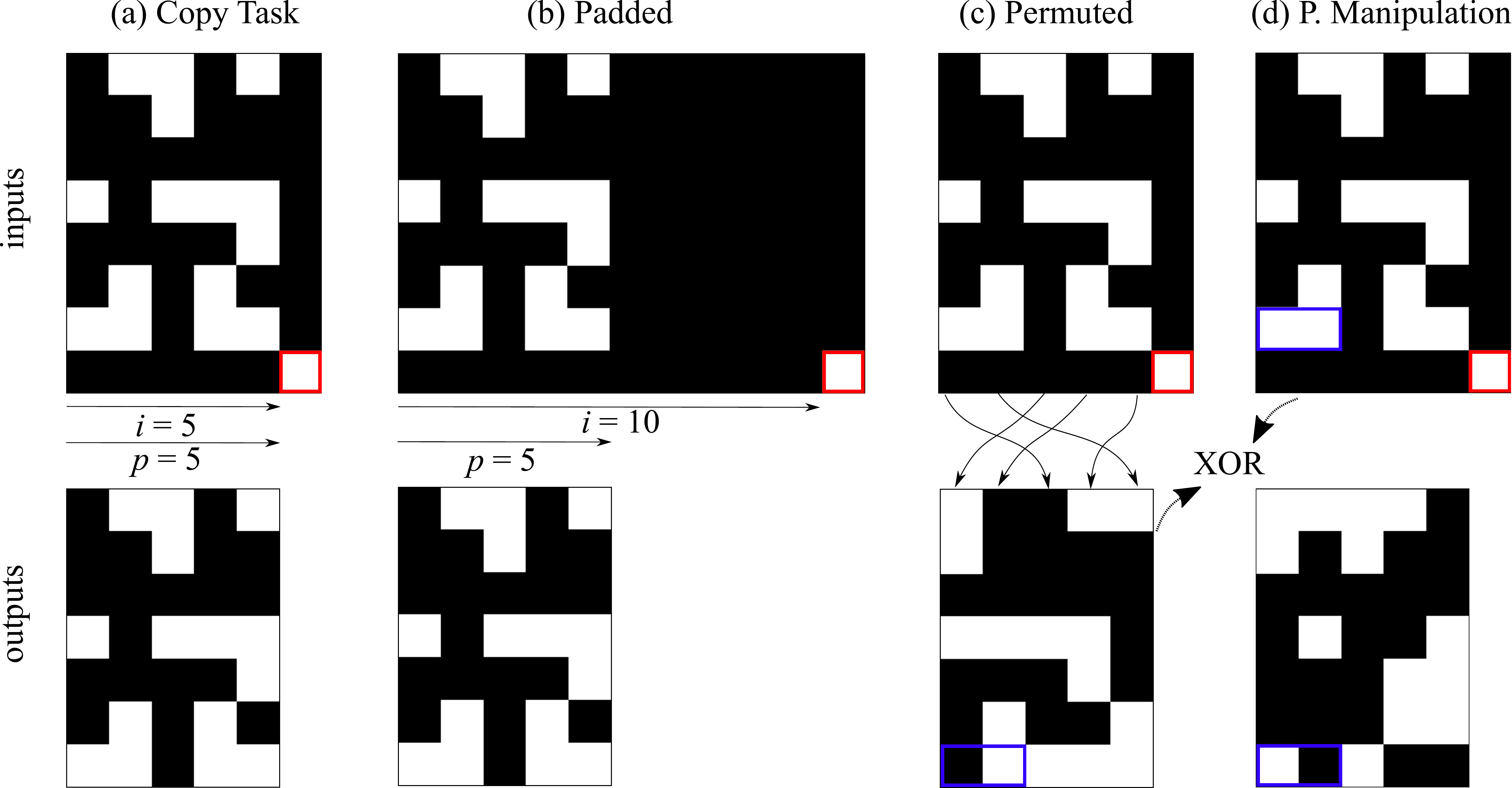}
    \caption{Variants of the Copy Task. The upper and lower rows show example input and output patterns, respectively. \textbf{(a)} In the basic Copy Task, the input sequence length $i$ is equal to the actual pattern length $p$ ($i=p=5$ in this example), and the output is identical to the input except for the fact that the stop bit (signaled in red) is omitted. \textbf{(b)} In the Padded Copy Task, the pattern of length $p$ to be copied is padded with zeros, yielding an input sequence length $i>p$. \textbf{(c)} In the Permuted Copy Task, the output corresponds to a time-permuted version of the input (indicated by the arrows). \textbf{(d)} In the Pattern Manipulation Task, the output is computed from the input pattern by applying a binary XOR operation iteratively with all of its $r$ permutations, where $r$ allows to control task difficulty. Here we illustrate the case where $r=1$, and the XOR is computed between the input and its permuted version shown in (c). The blue rectangles highlight this operation for two specific bits.}
    \label{fig:supp:copy:examples}
\end{figure}

In the basic version of the Copy Task, networks are trained to memorize and reproduce random sequences, whose input sequence length $i$ is equal to the length of the pattern $p$ to be copied ($i=p$, cf. Fig. \ref{fig:supp:copy:examples}). 
An input sample $\mathbf{x}_{1:T}$ (with $T = i+1+p$) consists of a random binary pattern at timesteps $t=1,\dots, p$, where only feature dimensions $1$ to $F_\text{in}-1$ are used for the binary pattern, while feature dimension $F_\text{in}$ is reserved for the stop bit. It contains zeroes at timesteps $t=i-p,\dots,i$, a stop flag at timestep $t=i+1$ and zeroes at timesteps $t=i+2,\dots,i+1+p$.
The target output sequence $\mathbf{y}_{1:T}$ has no feature dimension reserved for the stop bit ($F_\text{out} = F_\text{in}-1$). It consists of zeroes up to timestep $i+1$ and contains a copy of the random input pattern at timesteps $t=i+2,\dots,i+1+p$
(cf. Fig. \ref{fig:supp:copy:examples}).

\subsubsection{Padded Copy Task}

The Padded Copy Task is a simple extension of the basic version described above where $i>p$. This variant allows us to assess the effects of increasing sequence length, realized through increasing $i$ while keeping the complexity of the underlying task constant (i.e., keeping $p$ fixed).  

\subsubsection{Permuted Copy Task}

To adapt the Copy Task to a CL setting, we introduce the Permuted Copy Task. Here, the output sequence $\mathbf{y}_{1:T}$ corresponding to an input sequence $\mathbf{x}_{1:T}$ is obtained by permuting the random input pattern $\mathbf{x}_{1:p}$ along the time dimension before assigning it as target to $\mathbf{y}_{i+2:i+1+p}$ (cf. Sec. \ref{supp:data:copy:basic}). In our CL experiments, the subtasks differ in the random permutation which is used to generate these input-output mappings.

\subsubsection{Pattern Manipulation Task}

The main challenge of the Copy Task is the memorization and recall of the presented input sequences. However, we additionally wanted to test how CL methods are affected by data processing requirements that go beyond simple memorization and which are different across tasks. The Pattern Manipulation Task offers a way to gradually increase the difficulty of this processing. 
Here we exclusively consider the case where $p=i$. Target patterns $\mathbf{y}_{i+2:i+1+p}$ are generated from input patterns $\mathbf{x}_{1:p}$ by iterating the following procedure $r$ times (where $r$ determines the task difficulty). We start by assigning $\mathbf{y}_{i+2:i+1+p} \leftarrow \mathbf{x}_{1:p}$  and then iterate for $r' = 1 \dots r$
\begin{enumerate}
    \item Permute $\mathbf{x}_{1:p}$ along the time dimension using the $r'$-th permutation to generate a pattern $\mathbf{x}_{1:p}^{(r')}$.
    \item Update $\mathbf{y}_{i+2:i+1+p}$ by computing the logical XOR operation between the current $\mathbf{y}_{i+2:i+1+p}$ and $\mathbf{x}_{1:p}^{(r')}$. 
\end{enumerate}


\subsection{Sequential Stroke MNIST}
\label{supp:data:smnist}

Stroke MNIST (SMNIST, \citet{deJong2016Nov})  represents MNIST images as a sequence of quadruples $ \big\{ dx_i, dy_i, eos_i, eod_i \big\}^T_{i=1}$. 
The length $T$ of the sequence corresponds to the number of pen displacements needed to define the digit, $(dx_i, dy_i)$ correspond to the relative offset from the previous pen position, $eos_i$ is a binary feature denoting the end of a stroke, and $eod_i$ denotes the end of a digit.
We downloaded the dataset\footnote{ \url{https://github.com/edwin-de-jong/mnist-digits-stroke-sequence-data/}} and split the 70000 sample digits into training, validation and test sets (50000, 10000 and 10000 samples respectively).
Since samples have different sequence lengths $T$, we zero padded the samples to obtain a uniform input length of 117 (maximal $T$).
The result of this procedure is available for download.
\footnote{\url{https://www.dropbox.com/s/sadzc8qvjvexdtx/ss_mnist_data?dl=1}}
For our Split Sequential SMNIST experiments, we generated training, validation and test sample sequences from the corresponding digit sets. 
For experiments with $m$ digits per sequence, we generated the same number of samples for all of the possible $2^m$ binary sequences (e.g. 22, 23, 32 and 33 for $m = 2$ in the split containing only 2s and 3s).
Finally we randomly assigned the $2^m$ possible sequences to two classes to create a binary decision problem. 
      

\subsection{AudioSet}
\label{supp:data:audioset}

AudioSet \citep{audioset} consists of more than two million 10-second audio samples, that are hierarchically ordered into 632 classes.
To generate a set of classification tasks for CL, we selected and preprocessed a subset of the available data.\footnote{\url{https://research.google.com/audioset/download.html}}
Following \citet{kemker2018measuring}, we selected classes and samples according to the following criteria.
We only considered classes that have 
(1) no restrictions according to the AudioSet ontology, 
(2) no parent-child relationship with any of the other classes and 
(3) a quality estimate provided by human annotators of $\geq$ 70\%.
Samples were excluded if they did not contain data for the entire 10 seconds, or if they belonged to multiple of the considered classes.
This procedure yielded a set of 189 classes, out of which 106 had a number of samples $\geq$ 1000. 
To generate a balanced dataset, we randomly selected 1000 samples from each of the 100 classes with the highest number of samples. Finally, we split the 1000 samples per class into 800 samples for training and 200 samples for testing.  
The result of this procedure is available for download.\footnote{\url{https://www.dropbox.com/s/07dfeeuf5aq4w1h/audioset_data_balanced?dl=1}}
For our Split-AudioSet-10 experiments, we randomly grouped the 100 classes into 10 subtasks with 10 classes each. Validation samples were randomly selected from the training data, while maintaining the balance between classes.


\subsection{Multilingual Part-of-Speech Tagging}
\label{supp:data:pos}

The Universal Dependencies dataset \citep{NIVRE16.348} consists of grammar annotation treebanks from 92 different languages (version 2.6). For our multilingual Part-of-Speech (PoS) Tagging experiments we chose treebanks from 20 frequently used languages \citep[cf.][]{plank-etal-2016-multilingual}. If multiple treebanks are available for a given language, we choose treebanks according to the available number of samples and whether they use the universal tagset (17 tags). We use the provided splits for training, test and validation samples. The data we use in our experiments is available for download\footnote{\url{https://www.dropbox.com/s/9xjrtprc2mfxcla/mud_data_2_6.pickle?dl=1}}, and the exact choice of treebanks as well as any other preprocessing steps can be reproduced with the code that accompanies this paper. Lastly, we use pretrained polyglot word embeddings. \footnote{\texttt{https://sites.google.com/site/rmyeid/projects/polyglot}}

\section{Experimental details}
\label{supp:exp:details}

Here we give further details on the results provided in the main text, and describe the procedures that we used to obtained these results.


\subsection{Copy Task}
\label{supp:exp:details:copy}

The analyses on the intrinsic dimensionality of the RNN's hidden space were performed when learning a single task of the basic Copy Task setting (cf. Sec. \ref{supp:data:copy:basic}), where outputs are a copy of the inputs. We computed the hidden state activations $\mathbf{h}_{1:T}$ on the test set after learning the task. Then, we performed principal component analysis (PCA) on these activations, independently for each timestep. Specifically, for each timestep we performed PCA on a matrix of size $\mathbb{R}^{N \times n_h}$ , where $N$ is the number of samples in the test set, and $n_h$ is the number of hidden neurons. We then defined the intrinsic dimensionality of the hidden space as the number of principal components needed to explain 75\% of the variance. Qualitatively similar results can be obtained obtained irrespective of the value of this threshold (we tested 20\%, 30\% ... 90\%).


\subsection{Sequential Stroke MNIST}
\label{supp:exp:details:smnist}

We use LSTM main networks with 256 hidden units and a fully connected output head per task for all SMNIST experiments. 
Further parameter choices and hyperparameter searches are detailed in Sec. \ref{supp:exp:details:hpsearches}. 
Table \ref{supp:tab:ssmnist:results} shows all \texttt{during} and \texttt{final} accuracies of the SMNIST experiment described in Sec. \ref{sec:ssmnist}.

\vspace{2mm}
\begin{table*}[h]
    \renewcommand{\arraystretch}{1.15}
 \centering
  \caption{Task averaged \texttt{during} and \texttt{final} test accuracies for the SMNIST experiments (Mean $\pm$ SEM in \%, $n=10$).}
  \begin{small}
  \begin{tabular}{lccccc} 
    & \textbf{accuracy} & $\mathbf{m=1}$ & $\mathbf{m=2}$ & $\mathbf{m=3}$ & $\mathbf{m=4}$ \\ \toprule

    \multirow{2}{*}{\texttt{Online EWC}} & during & 98.52 $\pm$ 0.11 & 91.86 $\pm$ 1.13 & 83.62 $\pm$ 1.10 & 92.33 $\pm$ 2.49 \\ 
    & final & 97.05 $\pm$ 0.59 & 76.65 $\pm$ 3.39 & 75.26 $\pm$ 1.85 & 73.16 $\pm$ 0.98 \\ 
    \hline
    \multirow{2}{*}{\texttt{HNET}} & during & 99.54 $\pm$ 0.02 & 97.87 $\pm$ 0.99 & 91.63 $\pm$ 1.54 & 95.49 $\pm$ 1.13 \\ 
    & final & 99.52 $\pm$ 0.02 & 94.89 $\pm$ 3.81 & 91.63 $\pm$ 1.53 & 94.42 $\pm$ 1.85 \\ 
    \hline
    \multirow{2}{*}{\texttt{Fine-tuning}} & during & 99.67 $\pm$ 0.03 & 99.30 $\pm$ 0.04 & 99.14 $\pm$ 0.04 & 98.96 $\pm$ 0.03 \\ 
    & final & 89.07 $\pm$ 1.22 & 75.23 $\pm$ 2.25 & 68.28 $\pm$ 0.72 & 69.04 $\pm$ 0.73 \\ 
    \hline
    \multirow{2}{*}{\texttt{Masking}} & during & 99.63 $\pm$ 0.02 & 99.25 $\pm$  0.02 & 96.04 $\pm$  1.35 & 87.78 $\pm$  0.92 \\ 
    & final &  99.23 $\pm$ 0.14 & 93.96 $\pm$  1.29 & 86.20 $\pm$  0.95 & 76.75 $\pm$  1.32 \\ 
    \hline
    \multirow{2}{*}{\texttt{Masking + SI}} & during & 99.68 $\pm$ 0.02 & 99.02 $\pm$ 0.04 & 98.61 $\pm$  0.22 & 96.42 $\pm$  1.27 \\ 
    & final & 99.68 $\pm$ 0.02 & 99.02 $\pm$ 0.04 & 98.62 $\pm$  0.22 & 96.43 $\pm$  1.26 \\ 
    \hline
    \multirow{2}{*}{\texttt{SI}} & during & 99.21 $\pm$ 0.04 & 89.78 $\pm$ 1.09 & 74.74 $\pm$ 0.14 & 69.58 $\pm$ 0.55 \\ 
    & final & 97.08 $\pm$ 0.66 & 85.10 $\pm$ 1.57 & 74.72 $\pm$ 0.14 & 69.58 $\pm$ 0.55 \\ 
    \hline
    \multirow{2}{*}{\texttt{From scratch}} & during & 99.70 $\pm$ 0.02 & 95.86 $\pm$ 1.16 & 92.13 $\pm$ 1.25 & 88.48 $\pm$ 0.09 \\ 
    & final & 99.70 $\pm$ 0.02 & 95.86 $\pm$ 1.16 & 92.13 $\pm$ 1.25 & 88.48 $\pm$ 0.09 \\ 
    \hline
    \multirow{2}{*}{\texttt{Coresets}} & during & 99.61 $\pm$ 0.01 & 99.05 $\pm$ 0.03 & 98.70 $\pm$ 0.05 & 85.46 $\pm$ 1.60 \\ 
    & final & 99.40 $\pm$ 0.03 & 98.10 $\pm$ 0.07 & 97.60 $\pm$ 0.08 & 82.89 $\pm$ 1.80 \\ 
    \hline
    \multirow{2}{*}{\texttt{Multitask}} & during & 99.72 $\pm$ 0.02 & 99.19 $\pm$ 0.04 & 99.06 $\pm$ 0.04 & 98.72 $\pm$ 0.06 \\ 
    & final & 99.72 $\pm$ 0.02 & 99.19 $\pm$ 0.04 & 99.06 $\pm$ 0.04 & 98.72 $\pm$ 0.06 \\ 
  \end{tabular}
  \label{supp:tab:ssmnist:results}
  \end{small}
\end{table*}


\subsection{Split-AudioSet-10}
\label{supp:exp:details:audioset}

The experiments are performed using a main network with one LSTM layer with 32 units and a fully-connected output head per task.
We initially used larger LSTM layers but observed extensive overfitting. Therefore, we ran a \texttt{fine-tuning} hyperparameter search for LSTM layer sizes: 8, 16, 32, 64, 128 and 256 and chose 32 as it resulted in the least amount of overfitting, while not leading to significant drops in maximum \texttt{during} accuracy.
We also increased the hyperparameter search grid of the \texttt{Multitask} baseline compared to other reported results, incorporating larger batch sizes since all tasks are trained at once.

\begin{figure}[ht!]
  \centering
  \includegraphics[width=0.7\textwidth]{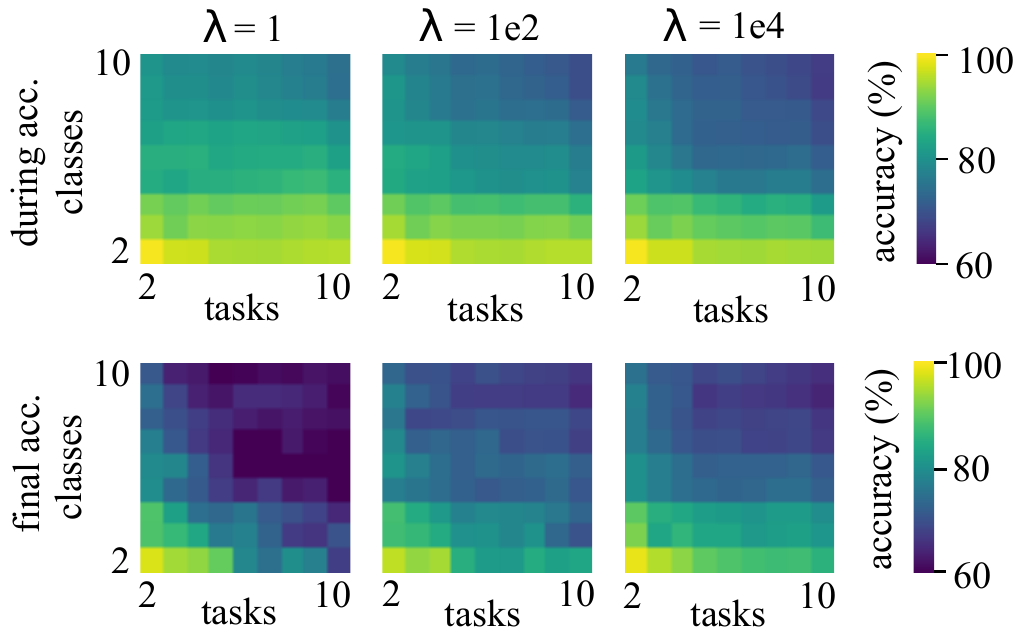}
  \caption{Task-averaged \texttt{during} and \texttt{final} test accuracies for \texttt{Online EWC} AudioSet experiments with varying numbers of classes per task, performed with different values for $\lambda_{EWC}$ (cf. Fig. \ref{fig:audioset_classes_vs_tasks}).}
  \label{supp:fig:audioset_lambdas}
\end{figure}

For our AudioSet experiments with varying levels of difficulty, we used the best hyperparameter configurations from our Split-AudioSet-10 experiments (cf. Table \ref{tab:audioset:CL1:results}). For \texttt{Online EWC}, we ran each experiment with multiple $\lambda$ values, as this parameter directly controls the trade-off between stability and plasticity. Fig. \ref{supp:fig:audioset_lambdas} shows \texttt{during} and \texttt{final} accuracies in the different settings for three $\lambda$ values. For the lowest $\lambda$ value, \texttt{during} accuracies are highest because few restrictions apply when solving individual tasks, but the performance drops when testing after all tasks are learned. For higher $\lambda$ values, \texttt{final} accuracies get closer to the \texttt{during} performance. This, however, comes at the cost of decreased \texttt{during} accuracies due to the restrictions imposed by the strong regularization controlled by $\lambda$.


\subsection{Pos Tagging}

We use bidirectional LSTM main networks with 32 units and a fully connected output head per task. We decided to not fine-tune word embeddings. Further parameter choices and hyperparameter searches are detailed in Sec. \ref{supp:exp:details:hpsearches}. 


\subsection{Hyperparameter searches}
\label{supp:exp:details:hpsearches}

We performed extensive hyperparameter searches for all methods in all experiments. Because of computational reasons, we limited the number of explored configurations to 100 per method and experiment (taking a random subset of all possible combinations defined by the search grid).
By default, we tested the run with the best \texttt{final} accuracy on multiple random seeds. If however, the best run did not prove to be random seed robust, we additionally evaluated the second and third best runs on multiple random seeds and selected the configuration with the best results across a set of random seeds. 
For the \texttt{HNET}, we only searched feedforward fully-connected architectures that yielded a compression ratio of approximately 1, meaning that the number of weights in the hypernetwork is approximately equal to the number of weights of the main RNN.
All experiments where conducted using the Adam optimizer.
For all results, exact command line calls are provided in the README files of the published code base. 

All experiments were performed with access to 32 GPUs of type NVIDIA RTX 2080 TI and NVIDIA QUADRO RTX 6000.

\subsubsection{Basic Copy Task}

For basic Copy Task experiments of a single task, used for the analyses on the intrinsic dimensionality of the RNN's hidden space, the hyperparameters searches are described in Table \ref{supp:tab:bcp:hpsearch}.

\vspace{3mm}
\begin{table*}[ht]
    \renewcommand{\arraystretch}{1.15}
 \centering
  \caption{Hyperparameter search for the Basic Copy Task}
  \begin{small}
  \begin{tabular}{ll} 
    \textbf{Hyperparameter} & \textbf{Searched values}\\ \toprule
    number of iterations & 20000\\ 
    number of hidden units of the main network & 256\\
    main network activation function & \textit{tanh}\\
    batch size & 68, 128\\ 
    learning rate & 5e-4, 1e-3, 5e-3, 1e-2 \\ 
    clip gradient norm & None, 1, 100 \\ 
    orthogonal initialization & True, False \\ 
    orthogonal regularization strength & 0, 1 \\ 
  \end{tabular}
  \label{supp:tab:bcp:hpsearch}
  \end{small}
\end{table*}

\subsubsection{Permuted Copy Task}

For Permuted Copy Task experiments with five tasks and $p=i=5$, the hyperparameter searches are described in Table \ref{supp:tab:pmct:hpsearch}.

\begin{table*}[ht]
    \renewcommand{\arraystretch}{1.15}
 \centering
  \caption{Hyperparameter search for the Permuted Copy Task}
  \begin{small}
  \begin{tabular}{lll} 
   \textbf{Method} & \textbf{Hyperparameter} & \textbf{Searched values}\\  \toprule
    \multirow{8}{*}{All} & number of iterations & 20000\\ 
    & batch size & 128\\ 
    & number of hidden units of the main network & 256\\ 
    & main network activation function & \textit{tanh}\\ 
    & learning rate & 5e-4, 1e-3, 5e-3, 1e-2 \\ 
    & clip gradient norm & None, 1, 100 \\ 
    & orthogonal initialization & True, False \\ 
    & orthogonal regularization strength & 0, 0.01, 1 \\ 
    \midrule
    \texttt{Online EWC} & $\lambda_{EWC}$ & 1e2, 1e3, \ldots, 1e10 \\ 
    \midrule
    \texttt{SI} & $\lambda_{SI}$ & 1e-3, 1e-2, 1e-1, 1, 1e2, 1e3 \\
    \midrule
    \texttt{Masking} & masked fraction & 0.2, 0.4, 0.6, 0.8 \\
    \midrule
    \multirow{2}{*}{\texttt{Masking + SI}} & $\lambda_{SI}$ &  1e-3, 1e-2, 1e-1, 1, 1e2, 1e3 \\
    & masked fraction & 0.2, 0.4, 0.6, 0.8 \\
    \midrule
    \multirow{4}{*}{\texttt{Generative Replay}} & strength of the prior-matching term ($\lambda_{pm}$) & 1, 10 \\
    & strength of the reconstruction term ($\lambda_{rec}$) & 1, 10 \\
    & strength of the soft-target distillation loss ($\lambda_{distill}$) & 1, 10 \\
    & dimensionality of the VAE latent space & 8, 100 \\
    \midrule
    \multirow{5}{*}{\texttt{HNET}} & $\beta$ &  1e-2, 1e-1, 1, 1e1, 1e2 \\ 
    & SD for the initialization of the task embeddings & .1, 1 \\ 
    & SD for the initialization of the chunk embeddings & .1, 1 \\ 
    & \texttt{HNET} hidden layers & "", "25,25", "50,50" \\ 
    & \texttt{HNET} output size & 2000, 5000 \\ 
    & chunk embedding size & 16, 32 \\ 
  \end{tabular}
  \label{supp:tab:pmct:hpsearch}
  \end{small}
\end{table*}

\subsubsection{Padded Copy Task}

For Padded Copy Task experiments with five tasks and $p=5$, $i=25$, the hyperparameter searches for the different methods are specified in Table \ref{supp:tab:pct:hpsearch}.

\vspace{3mm}
\begin{table*}[ht]
    \renewcommand{\arraystretch}{1.15}
 \centering
  \caption{Hyperparameter search for the Padded Copy Task}
  \begin{small}
  \begin{tabular}{lll} 
    \textbf{Method} & \textbf{Hyperparameter} & \textbf{Searched values}\\ \toprule
    \multirow{8}{*}{All} & number of iterations & 20000\\ 
    & batch size & 128\\ 
    & number of hidden units of the main network & 256\\ 
    & main network activation function & \textit{tanh}\\ 
    & learning rate & 5e-4, 1e-3, 5e-3, 1e-2 \\ 
    & clip gradient norm & None, 1, 100 \\ 
    & orthogonal initialization & True, False \\ 
    & orthogonal regularization strength & 0, 1 \\ 
    \midrule
    \texttt{Online EWC} & $\lambda_{EWC}$ & 1e2, 1e3, 1e4, 1e5, 1e6, 1e7, 1e8, 1e9, 1e10 \\ 
    \midrule
     \multirow{5}{*}{\texttt{HNET}} & $\beta$ &  5, 10, 50 \\ 
    & \texttt{HNET} hidden layers & "60,60,30" \\ 
    & \texttt{HNET} output size & 4000 \\ 
    & task embedding size & 16, 32 \\ 
    & chunk embedding size & 16, 32 \\ 
  \end{tabular}
  \label{supp:tab:pct:hpsearch}
  \end{small}
\end{table*}

\subsubsection{Pattern Manipulation Task}

For Pattern Manipulation Task experiments, the hyperparameter searches are described in Table \ref{supp:tab:pmt:hpsearch}.

\begin{table*}[ht]
    \renewcommand{\arraystretch}{1.15}
 \centering
  \caption{Hyperparameter search for the Pattern Manipulation Task}
  \begin{small}
  \begin{tabular}{lll} 
    \textbf{Method} & \textbf{Hyperparameter} & \textbf{Searched values}\\ \toprule
    \multirow{6}{*}{All} & number of iterations & 20000\\ 
    & batch size & 128\\ 
    & number of hidden units of the main network & 256\\ 
    & main network activation function & \textit{tanh}\\ 
    & learning rate & 5e-3, 1e-3, 5e-4 \\
    & clip gradient norm & None, 1, 100 \\ 
    \midrule
    \multirow{2}{*}{\texttt{Online EWC}} & $\lambda_{EWC}$ & 1e-2, 1e-1, 1e0, 1e1, 1e2, 1e3 \\ 
    & orthogonal regularization strength & 1, 10 \\ 
    \midrule
    \multirow{5}{*}{\texttt{HNET}} & $\beta$ & 1e-2, 1e-1, 1e0, 1e1, 1e2, 1e3 \\ 
    & \texttt{HNET} hidden layers & "64,64,64", "64,64,32" \\ 
    & \texttt{HNET} output size & 2000, 4000 \\ 
    & chunk embedding size &  32 \\ 
    & task embedding size & 32 \\ 
  \end{tabular}
  \label{supp:tab:pmt:hpsearch}
  \end{small}
\end{table*}

\subsubsection{Sequential Stroke MNIST}

The hyperparameter searches for Sequential Stroke MNIST experiments are described in Table \ref{supp:tab:smnist:hpsearch}.
The number of iterations was set according to the the number of digits in the sequences used in a given SMNIST experiment.

\vspace{3mm}
\begin{table*}[ht]
    \renewcommand{\arraystretch}{1.15}
 \centering
  \caption{Hyperparameter search for Sequential Stroke MNIST}
  \begin{small}
  \begin{tabular}{lll}
    \textbf{Method} & \textbf{Hyperparameter} & \textbf{Searched values}\\  \toprule
    \multirow{9}{*}{All} & batch size & 64, 128\\ 
    & number of hidden units of the main network & 256\\ 
    & main network activation function & \textit{tanh}\\ 
    & learning rate & 1e-3, 5e-3, 1e-4 \\
    & clip gradient norm & None, 1, 100 \\
    & number of iterations for $m=1$ & 2000, 3000, 4000\\ 
    & number of iterations for $m=2$ & 6000, 8000, 10000\\ 
    & number of iterations for $m=3$ & 8000, 12000, 16000\\ 
    & number of iterations for $m=4$ & 15000, 20000, 25000\\ 

    \midrule
    \texttt{Online EWC} & $\lambda_{EWC}$ &  1e1, 1e2, \ldots, 1e10 \\ 
    \midrule
    \texttt{SI} & $\lambda_{SI}$ &  1e-3, 1e-2, 1e-1, 1e0, 1e1, 1e2, 1e3 \\
    \midrule
    \texttt{Masking} & masked fraction & 0.2, 0.4, 0.6, 0.8 \\
    \midrule
    \multirow{2}{*}{\texttt{Masking + SI}} & $\lambda_{SI}$ &  1e-3, 1e-2, 1e-1, 1e0, 1e1, 1e2, 1e3 \\
    & masked fraction & 0.2, 0.4, 0.6, 0.8 \\
    \midrule
    \multirow{6}{*}{\texttt{HNET}} & $\beta$ & 1e-1, 1e0, 1e1 \\ 
    & \texttt{HNET} hidden layers & "32,32","32,16","64,32,16","32,32,32" \\ 
    & \texttt{HNET} output size & 8000, 16000 \\ 
    & chunk embedding size &  32, 64 \\ 
    & task embedding size & 32, 64 \\ 
    \midrule
    \texttt{Coreset} & $\lambda_{distill}$ & 1e-1, 1e0, 1e1 \\
    \midrule
    \texttt{Multitask} & batch size & 64, 128, 256, 512 \\
  \end{tabular}
  \label{supp:tab:smnist:hpsearch}
  \end{small}
\end{table*}

\subsubsection{Audioset}

The hyperparameter searches for Audioset experiments are described in Table \ref{supp:tab:audioset:hpsearch}.

\begin{table*}[ht]
    \renewcommand{\arraystretch}{1.15}
 \centering
  \caption{Hyperparameter search for Audioset}
  \begin{small}
  \begin{tabular}{lll}
    \textbf{Method} & \textbf{Hyperparameter} & \textbf{Searched values}\\  \toprule
    \multirow{8}{*}{All}& number of hidden units of the main network & 32\\ 
    & main network activation function & \textit{tanh}\\ 
    & batch size & 64, 128\\ 
    & number of iterations & 10000, 15000, 25000, 50000 \\
    & learning rate & 1e-3, 1e-4, 1e-5 \\
    & clip gradient norm & None, 1 \\
    & orthogonal initialization & False, True \\
    & orthogonal regularization strength & 0, .1 \\
    \midrule
    \texttt{Online EWC} & $\lambda_{EWC}$ &  1e-1, 1e0, \ldots, 1e10 \\ 
    \midrule
    \texttt{SI} & $\lambda_{SI}$ &  1e-4, 1e-3, 1e-2, 1e-1, 1, 1e2, 1e3, 1e4 \\
    \midrule
    \texttt{Masking} & masked fraction & 0.2, 0.4, 0.6, 0.8 \\
    \midrule
    \multirow{2}{*}{\texttt{Masking + SI}} & $\lambda_{SI}$ &  1e-4, 1e-3, 1e-2, 1e-1, 1, 1e2, 1e3, 1e4 \\
    & masked fraction & 0.2, 0.4, 0.6, 0.8 \\
    \midrule
    \multirow{7}{*}{\texttt{HNET}} & $\beta$ & 1e-2, 1e-1, 1, 1e1 \\ 
    & SD for the initialization of the task embeddings & .1, 1 \\
    & SD for the initialization of the chunk embeddings & .1, 1 \\
    & \texttt{HNET} hidden layers & "10,10", "20,20" \\ 
    & \texttt{HNET} output size & 1000, 2000 \\ 
    & chunk embedding size &  32 \\ 
    & task embedding size & 32 \\ 
    \midrule
    \texttt{Coreset} & $\lambda_{distill}$ & 1e-1, 1e0, 1e1 \\
    \midrule
    \texttt{Multitask} & batch size & 64, 128, 256 \\
  \end{tabular}
  \label{supp:tab:audioset:hpsearch}
  \end{small}
\end{table*}

\subsubsection{PoS Tagging}

The hyperparameter searches for the PoS-Tagging experiments are described in Table \ref{supp:tab:pos:hpsearch}.

\begin{table*}[ht]
    \renewcommand{\arraystretch}{1.15}
 \centering
  \caption{Hyperparameter search for PoS Tagging}
  \begin{small}
  \begin{tabular}{lll}
    \textbf{Method} & \textbf{Hyperparameter} & \textbf{Searched values}\\ \toprule
    \multirow{7}{*}{All}& number of hidden units of the main network & 32\\ 
    & main network activation function & \textit{tanh}\\ 
    & batch size & 64\\ 
    & number of iterations & 2500, 5000\\
    & learning rate & 5e-3, 1e-3, 1e-4 \\
    & clip gradient norm & None, 100 \\
    & orthogonal regularization strength & 0, 1 \\
    \midrule
    \texttt{Online EWC} & $\lambda_{EWC}$ &  1e1, 1e2, \ldots, 1e10 \\ 
    \midrule
    \texttt{SI} & $\lambda_{SI}$ &  1e-3, 1e-2, 1e-1, 1e0, 1e1, 1e2, 1e3 \\
    \midrule
    \multirow{5}{*}{\texttt{HNET}} & $\beta$ & 5e-2, 5e-1, 5e0, 5e1 \\ 
    & \texttt{HNET} hidden layers & "10,10", "25,25,25", "75,125" \\ 
    & \texttt{HNET} output size & 190, 1600, 4000 \\ 
    & \texttt{HNET} activation function & \textit{sigmoid}, \textit{relu} \\
    & chunk embedding size &  8, 32 \\ 
    & task embedding size & 8, 32 \\
    \midrule
    \texttt{Coreset} & $\lambda_{distill}$ & 1e-1, 1e0, 1e1 \\
    \midrule
    \texttt{Masking} & masked fraction & 0.2, 0.4, 0.6, 0.8 \\
    \midrule
    \multirow{2}{*}{\texttt{Masking + SI}} & $\lambda_{SI}$ &  1e-3, 1e-2, 1e-1, 1e0, 1e1, 1e2, 1e3 \\
    & masked fraction & 0.2, 0.4, 0.6, 0.8 \\
  \end{tabular}
  \label{supp:tab:pos:hpsearch}
  \end{small}
\end{table*}

\section{Supplementary experiments and further remarks}


\subsection{Processing sequential data with RNNs}

Although recent results suggest that feedforward networks, which have parallelization and optimization benefits during training \citep{pixelrnn}, can successfully process sequential data \citep{devlin-etal-2019-bert, oord2016wavenet, radford2019gpt-2}, RNNs still have theoretical benefits compared to their feedforward alternatives \citep{oord2016wavenet, vaswani2017transformer}, including an unlimited receptive field in time, and a linear time complexity in sequence length. We therefore consider research on RNNs as vital and hope that future works utilizes the insights and baselines provided in this study to develop CL algorithms tailored to RNNs.


\subsection{Notes on optimization for the Copy Task}
\label{supp:remarks:orth:reg}

We observed better empirical results with vanilla RNNs than with LSTMs in the variants of the Copy Task.
We also observed that throughout all CL methods, the Copy Task with vanilla RNNs can only be solved when using orthogonal regularization \citep{vorontsov2017orth:reg} for all hidden-to-hidden weight matrices, whereas orthogonal initialization did not seem to play an important role. 

The requirement of using orthogonal regularization poses a particular problem in combination with hypernetworks. In contrast to all other methods, orthogonal regularization will regularize the output of a neural network and not the weight matrix itself. We consistently observed that the orthogonal regularization loss is harder to optimize and usually plateaus at higher values when used in combination with hypernetworks. We unsuccessfully experimented with several potential resolutions to overcome this problem, but did not use any of them for the results reported in this paper.

We first tried an annealing schedule for the orthogonal regularization strength, starting at very high values putting the emphasis of the optimizer on producing orthogonal hidden-to-hidden matrices via the hypernetwork. This can be also viewed as a pretraining phase, where the hypernetwork is pretrained to produce orthogonal matrices (to sidestep the limitation that we cannot initialize hidden-to-hidden weights orthogonally when using a hypernetwork).

In another attempt, we periodically measured the highest singular value of the hypernetwork-produced hidden-to-hidden matrix, and divided the outputted matrix by it (inspired by spectral normalization,  \cite{miyato2018spectral:norm}). The purpose of this approach is to mitigate exploding activations/gradients and therefore to avoid the saturation of the tanh nonlinearity, which would lead to vanishing gradients.

However, we did not see consistent improvements using any of the aforementioned approaches and therefore neglected them for all our experiments.


\subsection{Analysis of Synaptic Intelligence}
\label{supp:remarks:si}
 
\begin{figure}
  \begin{center}
    \includegraphics[width=0.3\textwidth]{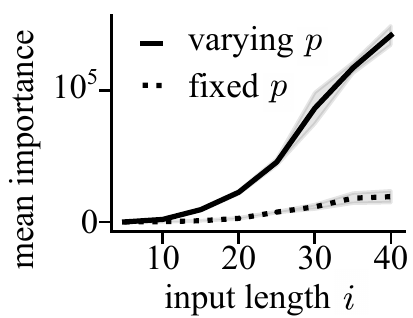}
  \end{center}
  \caption{Mean importance values (as computed in \texttt{SI}) of hidden-to-hidden weights after learning the Copy Task (solid line, $p=5,10,...40$) or the Padded Copy Task (dotted line, $p=5$) independently for an increasing set of sequence lengths $i$ (Mean $\pm$ SD, $n=5$). Cf. Fig. \ref{fig:copy_intuition}.}
  \label{supp:fig:importance_si}
\end{figure}

We repeated the analysis of weight-importance values for the Copy Task (cf. \ref{sec:an:copy}), but this time computing the weight-importance values as prescribed by \texttt{SI}. We obtained qualitatively similar results to \texttt{Online EWC} (Fig. \ref{supp:fig:importance_si}). Specifically, we observe that weight importance values noticeably increase when the length of the sequence to be recalled increases (varying $p$), whereas the increase is comparatively negligible when the pattern length remains constant (varying $i$ with constant $p$).
Thus, we were able to empirically validate the hypotheses derived from our linear analysis (cf. SM \ref{supp:linear:rnns}) in nonlinear RNNs for two different weight-importance methods, indicating that the described mechanisms might in general influence the performance of weight-importance methods in RNNs.


\subsection{Analysis of Weight-Importance Methods in LSTMs}
\label{supp:remarks:copy:lstm}

To test whether our results from the Copy Task (cf. Sec. \ref{sec:an:copy}) hold for different types of RNNs, we extend our analysis by using LSTMs instead of vanilla RNNs. Specifically, we use networks with an LSTM layer of 256 units, and a fully connected layer of 128 units before the actual output layer. 
Because we found it difficult to train LSTMs (cf. Sec. \ref{supp:remarks:orth:reg}) to recall long sequences in the Copy Task, we limit our analysis to inputs of length $i=5,10,..25$. All other details are identical to those explained in Sec. \ref{sec:an:copy}.

\begin{figure}
  \centering
  \begin{subfigure}[b]{1.65in}
    \includegraphics[width=\textwidth]{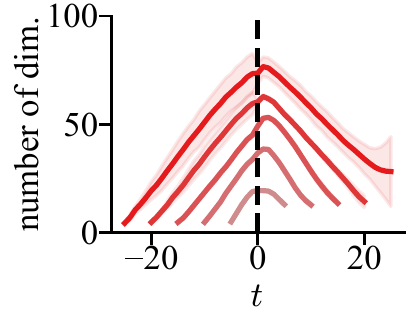}
    \caption{}
  \end{subfigure}
  \hfill
  \begin{subfigure}[b]{1.85in}
    \includegraphics[width=\textwidth]{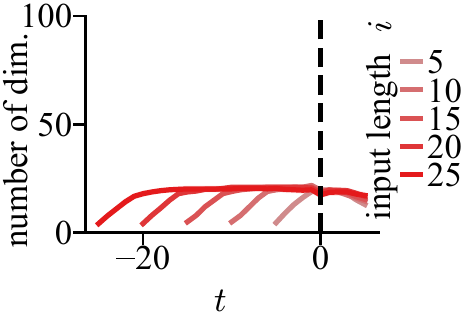}
    \caption{}
  \end{subfigure}
  \hfill
  \begin{subfigure}[b]{1.65in}
    \includegraphics[width=\textwidth]{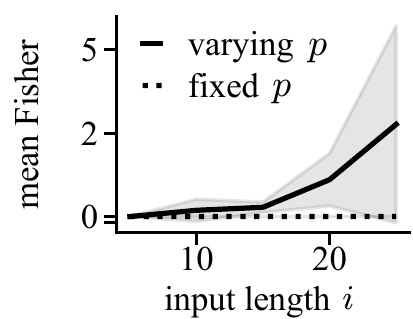}
    \caption{}
  \end{subfigure}
  \caption{Same analysis as in Fig. \ref{fig:copy_intuition} but for LSTM instead of vanilla RNNs. \textbf{(a)} Intrinsic dimensionality per timestep of the 256-dimensional LSTM hidden space $\mathbf{h}_t$ for the basic Copy Task, where input and pattern lengths are tied ($i$ = $p$). The stop bit (dotted black line) is shown at time $t=0$ (Mean $\pm$ SD, $n=5$). \textbf{(b)} Same as (a) for the Padded Copy Task, where the pattern length is fixed ($p=5$) but input length $i$ varies. In (a) and (b), dimensionality of the hidden state space increases only during input pattern presentation. \textbf{(c)} Mean Fisher values (weight-importance values in \texttt{Online EWC}) of recurrent weights after learning the Copy Task (solid line, $p=5,10,...25$) or the Padded Copy Task (dotted line, $p=5$) independently for an increasing set of sequence lengths $i$ (Mean $\pm$ SD, $n=5$). }
  \label{supp:fig:copy:lstm}
\end{figure}

The results of this analysis are shown in Fig. \ref{supp:fig:copy:lstm}.
We observe that the intrinsic dimensionality of the LSTM hidden space increases with the length $p$ of the pattern to be copied (Fig. \ref{supp:fig:copy:lstm}a) but not necessarily with the input length $i$ (Fig. \ref{supp:fig:copy:lstm}). Furthermore, we observe that an increase in the intrinsic dimensionality of the hidden space relates to an increase in mean importance values as calculated with \texttt{Online EWC}. 
Therefore, we observe with LSTMs the same trends we observed for vanilla RNNs. This validates the observation that working memory requirements, and not necessarily sequence length, lead to an increase in importance values as computed by weight-importance methods when sequentially learning a set of tasks with a recurrent network.


\subsection{Determining task-relevant intrinsic dimensionality}
\label{supp:remarks:supervised:dim:red}

The experiments in Sec. \ref{sec:an:copy} and SM \ref{supp:remarks:si} to analyse the intrinsic task-specific dimensionality of the hidden-state for varying task complexities relied on unsupervised dimensionality reduction methods. As such, these methods cannot differentiate between task-unrelated (e.g., random) information encoded in the hidden state and information extracted by the output heads from those hidden states in order to perform inference. To overcome this limitation we perform a simple type of supervised dimensionality reduction, which allows us determine the subspace of the hidden state that contains the necessary information to carry out predictions. In particular, we are interested to perform this type of analysis in a CL setting, i.e., studying the evolution of the  intrinsic hidden-state dimensionality when successively learning tasks with comparable complexity.

\begin{figure}[ht!]
  \centering
  \includegraphics[width=0.48\textwidth]{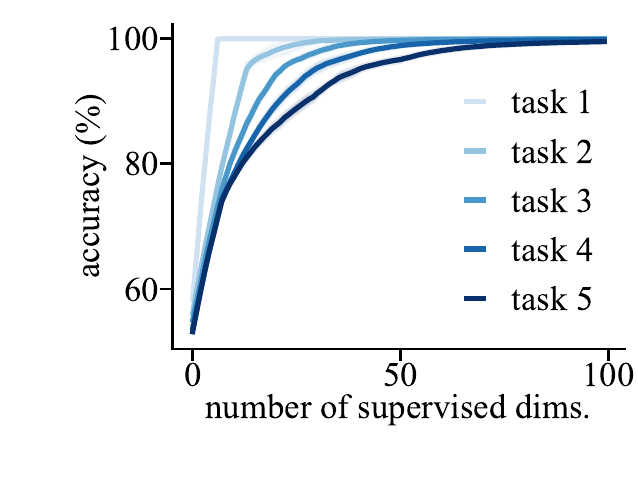}
  \caption{To complement our analysis of intrinsic hidden-state  dimensionality for tasks with varying complexity from Sec. \ref{sec:an:copy} and SM \ref{supp:remarks:si}, we provide here the evolution of the estimated hidden-state dimensionality when continually learning five subtasks of the \textit{Permuted Copy Task} using \texttt{Online EWC} (cf. \ref{sec:exp:copy}). The plot shows, for any given dimensionality, what predictive performance can be achieved when choosing the respective hidden-state subspace that maximizes this performance (cf. SM \ref{supp:remarks:supervised:dim:red} for details). The task value indicates the number of tasks that have been learned up to that point (Mean $\pm$ SD in \%, $n=5$).}
  \label{supp:fig:supervised_dim_red}
\end{figure}

In the following, we briefly sketch how we perform supervised dimensionality reduction. We require the checkpointed RNN model parameters $\tilde{\psi}^{(k)}$ obtained after training on task $k$. We can use these models to collect the hidden states for samples of all tasks trained so far (i.e., from $\mathcal{D}_1, \dots, \mathcal{D}_k$). We gather those hidden states into the rows of a matrix $H^{(k)}$ and perform a reconstruction from an $n_\text{red} \leq n_h$ dimensional subspace using an approximately orthonormal matrix $U^{(k)} \in \mathbb{R}^{n_h \times n_h}$ via $\tilde{H}^{(k)} = H^{(k)} U^{(k)}_{:n_\text{red}} (U_{:n_\text{red}}^{(k)})^T$, where $U_{:n_\text{red}}^{(k)} \in \mathbb{R}^{n_h \times n_\text{red}}$ denotes the matrix obtained by taking the first $n_\text{red}$ columns from $U^{(k)}$, and $n_h$ is the number of hidden neurons. We can then continue obtaining predictions from the RNN with parameters $\tilde{\psi}^{(k)}$ using the reconstructions $\tilde{H}^{(k)}$ and compare these to the ground-truth targets in order to judge the quality of the subspace.

We now describe how we obtain $U^{(k)}$ for each task $k$. As this process is identical for every $k$, we omit the superscript $k$ from our notation. The matrix $U$ is retrieved by an iterative procedure (column-by-column), where every added column is optimized to extract a new subspace dimension that maximizes predictive performance. Let $\mathbf{\hat{y}}_{1:T} = f_\text{out}(\mathbf{h}_{1:T}, \tilde{\psi})$ denote the part of the RNN $f$ that generates predictions from a given hidden state sequence $\mathbf{h}_{1:T}$ (note, $f_\text{out}$ only performs time-independent feed-forward computation, thus processing hidden states independently).
In general, different timesteps may use different subspaces, so the matrix $U$ should be specific to a predefined timestep $s$, such that reconstructed hidden states can be obtained via $\mathbf{\tilde{h}}_s^T = \mathbf{h}_s^T U_{:n_\text{red}} U_{:n_\text{red}}^T$, $\mathbf{\hat{y}}_s = f_\text{out}(\mathbf{\tilde{h}}_s, \tilde{\psi})$ and $(\mathbf{\hat{y}}_t, \mathbf{\tilde{h}}_t) = f_\text{step}(\mathbf{x}_t, \mathbf{\tilde{h}}_{t-1}, \tilde{\psi})$ for $t=s+1, \dots, T$. The network predictions $\mathbf{\hat{y}}_{s:T}$ from the \textit{reconstructed} hidden states $\mathbf{\tilde{h}}_{s:T}$ can then be compared to the ground-truth $\mathbf{y}_{s:T}$ via some loss $\mathcal{L}_\text{task}(\mathbf{\hat{y}}_{s:T}, \mathbf{y}_{s:T})$ and $U$ can be optimized to minimize this loss. More precisely, we minimize the following loss function iteratively for $n=1, \dots, n_h$:

\begin{equation}
    \min_{\mathbf{u}_n} \, \lVert U_{:n}^T U_{:n} - I_{:n} \lVert^2 + \gamma \mathcal{L}_\text{task}(\tilde{\mathcal{Y}}, \mathcal{Y})
\end{equation}

where $\gamma$ is a hyperparameter, $\mathbf{u}_n$ denotes the $n$-th column of U and $\tilde{\mathcal{Y}}$ consists of sequences $\mathbf{\hat{y}}_{s:T}$ computed from reconstructed hidden states using $U$ with input samples taken from the training sets of all previously learned tasks, while $\mathcal{Y}$ contains the corresponding ground-truth sequences $\mathbf{y}_{s:T}$.

As this process is computationally quite demanding and it is not a priori clear which timesteps to focus on, we decided to take a simpler approach and learn a shared matrix $U$ for all timesteps, such that for every $t$ we compute a reconstruction using $\mathbf{\tilde{h}}_t^T = \mathbf{h}_t^T U_{:n_\text{red}} U_{:n_\text{red}}^T$ as well as the corresponding predictions $\mathbf{\hat{y}}_{1:T} = f_\text{out}(\mathbf{\tilde{h}}_{1:T}, \tilde{\psi})$. With this procedure, we might overestimate the actual intrinsic dimensionality required for solving all tasks seen so far, but we can still assess relative differences, e.g., an increase in intrinsic dimensionality if more tasks are learned. 

The results of this analysis on an instance of the \textit{Permuted Copy Task} are depicted in Fig. \ref{supp:fig:supervised_dim_red}.
We observe that the number of dimensions required to achieve high accuracy levels increases as more tasks are being learned. In other words, the task-relevant dimensionality of the hidden space increases with the number of tasks that has been learned so far, which aligns with the predictions from SM \ref{supp:linear:rnns}. 

It is worth noting that if $f_\text{out}$ is a linear operation, one could directly obtain the dimensionality of task-specific subspaces read by each output head by applying singular value decomposition to the hidden-to-output weight matrix, and thresholding the singular values. This procedure allows one to obtain a task-specific basis $U^{(k)}$, which can be compared to other tasks $k'$ by computing the subspace similarity $\text{sim}(k, k') = \lVert U^{(k)} (U^{(k')})^T \rVert_F$ and can therefore serve as a proxy for task similarity.


\subsection{Increased difficulty of the Permuted Copy Task}
\label{supp:remarks:perm:copy:difficulty}

We empirically observed that the Permuted Copy Task (cf. \ref{sec:exp:copy}) is harder to solve (for both vanilla RNNs and LSTMs, data not shown). Intuitively, such increase in difficulty can already be anticipated by analyzing a linear RNN (cf. Eq.~\ref{supp:eq:linear:rnn}). The basic Copy Task can be manually implemented as linear RNN by realizing a queue-like mechanism (i.e., the input-to-hidden weights write inputs into a subspace of the hidden space, while the hidden-to-hidden weights shift these chunks consecutively through subspaces until they reach an output subspace which is read out by the hidden-to-output weights). This specific implementation cannot be trivially extended to the time-permuted case (where the order in the queue needs to change before elements are shifted to the output subspace), which indicates why an increase in difficulty may occur. 

We hypothesize that the increase in difficulty can also be linked to optimization, and more specifically to the large variation in backpropagation through time (BPTT) path lengths from each output timestep to its corresponding input timestep. Note that the mean BPTT path length is the same for the permuted and unpermuted case, but the standard deviation is zero for the unpermuted case. We observed that this variability in BPTT path lengths creates an optimization bias towards pairs of input/output timesteps that lie closer together in time (data not shown). Furthermore, previous work suggested in similar sets of experiments that the order of recall matters \cite[e.g., ][]{zaremba2014learning:to:execute:copy:task:order}, providing more evidence that there are indeed intrinsic differences between solving the basic and Permuted Copy Task.


\subsection{Replay for Split-SMNIST experiments} 
\label{supp:remarks:smnist:replay}

\begin{wraptable}{r}{.5\textwidth}
 \centering
  \caption{Mean \texttt{during} and \texttt{final} accuracies for Split-SMNIST rehearsal experiments (Mean $\pm$ SEM in \%, $n=10$). Method \texttt{RtF} was denoted \texttt{Generative Replay} in the main text. Both, \texttt{RtF} and \texttt{HNET+R}, are introduced in Sec. \ref{supp:methods:generative:replay:cl:with:svae}. Methods denoted with a * use a decoder architecture that has an additional fully-connected layer of size 256 before and after the LSTM layer.}
  \begin{small}
      \begin{tabular}{lcc}
        & \textbf{during} & \textbf{final} \\ \toprule
        \texttt{Multitask} & N/A & 99.18 $\pm$ 0.05\\
        \texttt{Coresets-10} &  99.64 $\pm$  0.02 & 96.44 $\pm$  0.25\\
        \texttt{Coresets-100} &  99.51 $\pm$  0.01 & 98.85 $\pm$  0.05\\
        \texttt{RtF} &  98.95 $\pm$  0.08 & 95.01 $\pm$  0.88\\
        \texttt{RtF}$^*$ &  99.51 $\pm$  0.02 & 98.41 $\pm$  0.22\\
        \texttt{HNET+R} &  99.67 $\pm$  0.01 & 99.34 $\pm$  0.04\\
        \texttt{HNET+R}$^*$ &  99.44 $\pm$  0.03 & 99.10 $\pm$  0.13\\
      \end{tabular}
  \label{tab:smnist:replay}
  \end{small}
\end{wraptable}

To complement our investigations of the Split-SMNIST experiments in Sec. \ref{sec:ssmnist}, we provide further experimental results on rehearsal methods in this section. As the training of generative models is challenging on real-world data, we restrict our exploration in this section to the original Split-SMNIST experiment, i.e., difficulty $m=1$ (cf. Sec. \ref{sec:ssmnist}).

The results are shown in Table \ref{tab:smnist:replay}. As can be seen, hypernetwork-protected replay \texttt{HNET+R} outperforms other rehearsal approaches and performs on par with \texttt{Multitask} training.
However, when analysing results obtained from methods based on generative replay, namely \texttt{RtF} and \texttt{HNET+R}, we realized that even though reconstruction is feasible, rehearsal via samples obtained from the prior did not lead to visually meaningful digits.\footnote{Note that a sequence of pen-strokes (SMNIST sample) can easily be converted into an image.} Aside from the difficulty of training a generative model, we hypothesize that this behavior is due to the coarse approximations made in Sec. \ref{supp:methods:generative:replay:cl:with:svae}. Interestingly, we did not observe these difficulties for the Copy Task, where input samples are sequences without direct temporal dependencies (aside from the correct placement of the stop bit).


\subsection{Sequence length and working memory requirements in real world tasks}
\label{supp:remarks:upsampling}

Analyzing how sequence length and working memory requirements influence CL in RNNs requires a setting in which these two factors can be clearly disentangled. 
The synthetic Copy Task provides the flexibility to independently manipulate the two aspects and thus isolate their effects.
However, it is unclear whether the observations made on this simple task generalize to more complex, real-world scenarios.
Here we show further experiments that support our Copy Task results in a more practically relevant scenario related to the concept of temporal resolution.

Sequential data often contains redundant information, e.g. when the sampling process is much faster than the data generation process. 
Thus, we can manipulate the length of a sequence without changing the information it contains by temporal upsampling.
In the same way, we can change the sequence length of a task's input, without noticeably affecting the working memory required to solve it.

In our sequential SMNIST experiments (cf. Sec \ref{sec:ssmnist}), input sequence length and working memory requirements both increase with increasing digit sequence length $m$. 
Here, we contrast this setting with Split-SMNIST experiments with fixed working memory requirements ($m=1$), but increasing input sequence length (as dictated by an upsampling factor $u \in \{2,3,4\}$).

We use \texttt{Online EWC} to train networks on five tasks for three different upsampling factors and display the results in Table \ref{supp:tab:upsampling}.
Increasing the upsampling factor $u$ does not yield a drop in performance. 
This is contrasted by the sequential SMNIST results, where an increase in the number of digits $m$ leads to substantially worse results. 
This difference indicates that the drop in performance observed for increasing $m$ is caused by higher task difficulty, and is independent of the length of input sequences. 
Taken together, these results support our conclusions from the Copy Task experiments and show that they hold in less artificial settings. 
 
\vspace{3mm}
\begin{table*}[ht]
 \centering
  \caption{Mean \texttt{final} accuracies for Split-SMNIST experiments under different task difficulty scenarios. $l$ is the factor by which input sequence length is increased w.r.t the original sequence composed of one digit. This factor can be increased by either increasing the number of digits $m$ (top row, copied from sequential SMNIST experiments), or by increasing the upsampling factor $u$ (bottom row). (Mean $\pm$ SEM in \%, $n=10$)}
  \begin{small}
      \begin{tabular}{lcccc}
        \textbf{Setting} & $\mathbf{l=1}$ & $\mathbf{l=2}$ & $\mathbf{l=3}$ & $\mathbf{l=4}$ \\ \toprule
        Varying $m$ &  97.05 $\pm$ 0.59 & 76.65 $\pm$ 3.39 & 75.26 $\pm$ 1.85 & 73.16 $\pm$ 0.98 \\ 
        Varying $u$ &  97.05 $\pm$ 0.59 & 95.35 $\pm$  1.06 & 94.65 $\pm$  0.85 & 96.27 $\pm$  0.66 \\ 
      \end{tabular}
  \label{supp:tab:upsampling}
  \end{small}
\end{table*}


\subsection{Part-of-Speech Tagging}
\label{supp:remarks:pos}

Part-of-Speech Tagging is a classical NLP task in which words in a sentence are tagged according to their grammatical properties (e.g. nouns, verbs, etc.). The Universal Dependencies dataset \citep{NIVRE16.348} contains large annotated text corpora from a wide range of languages and has previously been used in a multitask setting \citep{plank-etal-2016-multilingual,heinzerling-strube-2019-sequence}. We cast this dataset to a CL setting, by sequentially training RNNs one language at a time. 
The results of training bidirectional LSTMs (BiLSTM) on 20 languages are displayed in Table \ref{tab:pos:results}. 

\begin{table*}[b]
 \centering
  \caption{Mean \texttt{during} and \texttt{final} accuracies for the PoS experiments (Mean $\pm$ SEM in \%, $n=10$).}
  \begin{small}
      \begin{tabular}{lcc}
        & \textbf{during} & \textbf{final} \\ \toprule
        \texttt{Multitask} & N/A & 92.52 $\pm$ 0.02 \\
        \texttt{From-scratch} & N/A & 95.04 $\pm$  0.01 \\
        \texttt{Fine-tuning} & 91.63 $\pm$  0.01 & 48.62 $\pm$  0.58 \\
        \texttt{HNET} & 89.83 $\pm$  0.09 & 89.30 $\pm$  0.09  \\
        \texttt{Online EWC} & 87.49 $\pm$  0.04 & 86.89 $\pm$  0.03  \\
        \texttt{SI} & 85.66 $\pm$  0.06 & 84.62 $\pm$  0.08 \\
        \texttt{Masking} & 91.29 $\pm$  0.02 & 46.76 $\pm$  0.81 \\
        \texttt{Masking+SI} & 82.60 $\pm$  0.09 & 82.34 $\pm$  0.09 \\
        \texttt{Coresets-$100$} & 91.64 $\pm$  0.02 & 90.05 $\pm$  0.03 \\ 
        \texttt{Coresets-$500$} & 91.60 $\pm$  0.03 & 90.63 $\pm$  0.03  \\
      \end{tabular}
  \label{tab:pos:results}
  \end{small}
\end{table*}

We can observe a similar trend to other experiments, where \texttt{HNET} outperforms other regularization methods. The performance of \texttt{Masking+SI} indicates that no good trade-off has been found between small, distinct but plastic subnetworks and large, overlapping and thus rigid subnetworks. Again the simple method \texttt{Coresets} exhibits the strongest performance among CL approaches, making it the preferable choice when data storage is feasible.


\subsection{Task Similarity and Forward Transfer}
\label{supp:remarks:forward:transfer}

Our analysis on linear RNNs learning multiple tasks relies on the assumption that the tasks are so different that they cannot share any recurrent computation. This is, however, an extreme scenario and in practice some level of similarity will exist between the tasks. CL approaches can in theory exploit this similarity by using some form of shared processing across tasks, and therefore facilitating knowledge transfer between tasks. 


In this section, we provide additional analyses and insights regarding task similarity and transfer of knowledge. For this, we consider a Permuted Copy Task experiment with ten tasks, where we present the same set of five tasks to the network twice. A thorough comparison of how each method might benefit from knowledge transfer is beyond the scope of our work. However, we provide here some pointers on how weight-importance methods and the hypernetwork approach might reuse previously acquired knowledge, noting that the two approaches transfer knowledge in a fundamentally different way. 

\begin{figure}
  \centering
  \begin{subfigure}[b]{1.7in}
    \includegraphics[width=\textwidth]{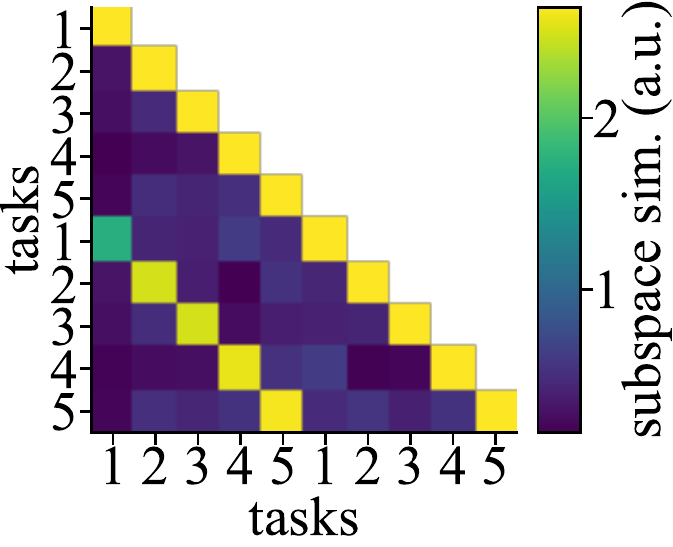}
    \caption{}
    \label{fig:transfer:ewc}
  \end{subfigure}
  \hfill
  \begin{subfigure}[b]{1.7in}
    \includegraphics[width=\textwidth]{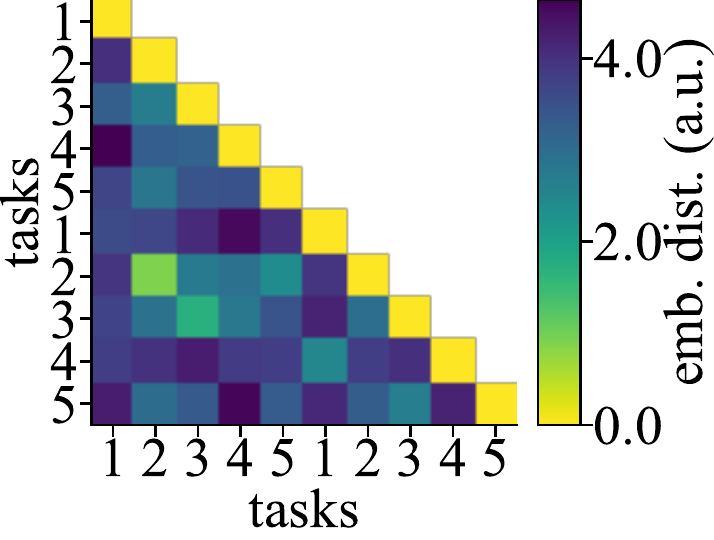}
    \caption{}
    \label{fig:transfer:hnet:emb}
  \end{subfigure}
  \hfill
  \begin{subfigure}[b]{1.7in}
    \includegraphics[width=\textwidth]{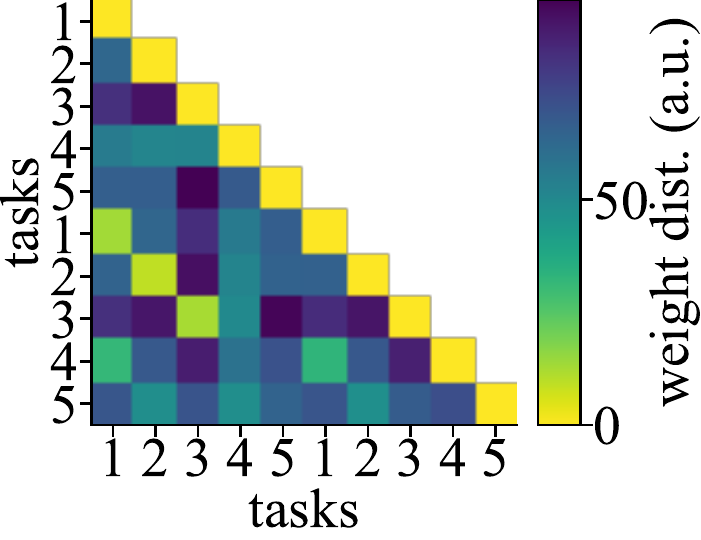}
    \caption{}
    \label{fig:transfer:hnet:weights}
  \end{subfigure}
  \caption{Mechanisms for knowledge transfer in a Permuted Copy Task setting with similarity across tasks (tasks 1 to 5 are identical to tasks 6 to 10). \textbf{(a)} Subspace similarity between tasks for \texttt{Online EWC}. \textbf{(b)} Euclidean distance between task-embeddings for \texttt{HNET}. \textbf{(c)} Euclidean distance in task-specific \texttt{HNET} outputs (i.e. main network weights).}
\end{figure}

\paragraph{Weight-importance methods.}
Since weight-importance methods use the same set of weights for all tasks, computation relevant for multiple tasks can directly be reused. In the experiment mentioned above, \texttt{Online EWC} achieves a mean final accuracy of 99.01\%. To compare the subspaces read out by the task-specific linear output heads, we compute subspace similarities between pairs of tasks by performing SVD on the head-specific weight matrices, as described at the end of Sec. \ref{supp:remarks:supervised:dim:red}. Fig. \ref{fig:transfer:ewc} shows that the subspaces used to solve the first set of tasks are being reused when the network solves the tasks for the second time, demonstrating that subspace reuse is a possible mechanism for transfer in weight-importance methods. 

\paragraph{Hypernetworks.}

\citet{oswald:hypercl} showed that hypernetworks can benefit from forward transfer between tasks. Since in this approach a new set of weights is generated for every task, transfer has to occur in the mapping from task embeddings to main network weights. Indeed, \citet{oswald:hypercl} also showed that this can be mediated by a suitably structured embedding space. For the Permuted Copy Task version described above, where the hypernetwork approach achieves a mean final accuracy of 98.74\%, transfer could occur by simply reusing the embeddings of the first set of tasks. However, the pairwise distances between task embedding vectors, as well as between the generated main network weights do not strongly support this (i.e. only the embeddings and weights of the tasks 2 and 3 being learned for the second time show some similarity to the original solutions). This could indicate that, in this setting, \texttt{HNET} finds a different solution when solving a task for the second time. Alternatively, this can be explained by the fact that Euclidean distances are not an appropriate measure to capture transfer in this scenario, since proximity in the embedding and weight spaces does not necessarily reflect similarity in the functional space.

One way to incentivize the reuse of previously found solutions, could be to selectively increase the learning rate of the embedding optimization when starting to train on a new task. This could allow for fast exploration of the space of existing solutions and opens interesting avenues for future work. 

\subsection{Discussion on task-specific solutions in CL}
\label{supp:remarks:hnet:benefits}

In this section, we further discuss differences between CL regularization approaches, specifically between weight-importance methods and \texttt{HNET}.

We start by reiterating our main motivation for proposing the use of \texttt{HNET} for RNNs. Because RNNs, just like feedforward networks, have a static set of weights, the regularization employed by \texttt{HNET} is not affected by the recurrent processing occurring in the main network (cf. Eq.~\ref{eq:hnet:reg}). Therefore, the CL regularization is independent of the choice between an RNN or feedforward network as main network. 
Note however, that individual tasks are initially learned through the recurrent main network while the data is available (thus before CL regularization is required); therefore the sequential nature of individual tasks does influence hypernetwork optimization (Sec. \ref{supp:remarks:orth:reg}).

Another benefit of the method \texttt{HNET} becomes apparent when considering the non-parametric limit. If the hypernetwork is considered a universal function approximator, the generation of a new set of weights per task can in theory allow CL without any forgetting. In contrast, weight-importance methods cannot provide the same guarantees even if the main network is considered a universal function approximator. For instance, it is not guaranteed that the posterior mode found via \texttt{Online EWC} for the first task contains viable solutions for all upcoming tasks.

In this work, we show that these intuitions transfer to practical applications and that \texttt{HNET} often exhibits strong performance advantages over weight-importance methods. To guide future research in this area, we devote the rest of the section to the main conceptual difference between these methods, namely that \texttt{HNET} allows task-specific solutions whereas weight-importance methods aim for a trade-off solution (cf. Fig. \ref{fig:prior:focused:vs:hnet}). We draw intuition from Bayesian inference and therefore confine ourselves to a subclass of weight-importance methods that can be interpreted as prior-focused methods (cf. \cite{gal:bcl}, e.g., \texttt{Online EWC}). In addition, we ignore knowledge transfer through the shared meta-model and consider \texttt{HNET} as an approximation to the \texttt{From-Scratch} baseline. In this simplified setting, \texttt{From-Scratch}/\texttt{HNET} have the ability to gather a sample from a task-specific posterior (e.g., the maximum-a-posteriori (MAP)), i.e., for tasks $k = 1, \dots, K$:

\begin{equation}
    \label{supp:eq:remarks:from:scratch:map}
    \psi_\text{MAP}^{(k)} = \arg\max_\psi p(\psi \mid \mathcal{D}_k) = \arg\min_\psi - \big( \log p(\mathcal{D}_k \mid \psi) + \log p(\psi) \big)
\end{equation}

Note, that the right-hand side of the equation above often matches the optimization criterion applied to the current task, where $-\log p(\mathcal{D}_k \mid \psi)$ matches the negative log-likelihood (cf. Eq.~\ref{supp:eq:nll:sequential}) and the prior influence can be realized via weight-decay if $p(\psi)$ is assumed to be an isotropic Gaussian distribution.

In contrast, a prior-focused method aims to find a single shared solution:

\begin{equation}
    \label{supp:eq:remarks:ewc:map}
    \psi_\text{MAP}^{(1:K)} = \arg\max_\psi p(\psi \mid \mathcal{D}_1, \dots, \mathcal{D}_K) = \arg\min_\psi - \big( \log p(\mathcal{D}_K \mid \psi) + \log p(\psi \mid \mathcal{D}_1, \dots, \mathcal{D}_{K-1}) \big)
\end{equation}

For completeness, if the main network weights $\psi$ are split into a shared body $\psi_\text{shared}$ and task-specific output-head weights $\psi_\text{specific}^{(1)}, \dots, \psi_\text{specific}^{(K)}$, the equation above becomes

\begin{align}
    \psi_\text{MAP}^{(1:K)} &= \arg\max_\psi p(\psi \mid \mathcal{D}_1, \dots, \mathcal{D}_K) \\
    &= \arg\min_\psi - \bigg( \log p(\mathcal{D}_K \mid \psi_\text{shared}, \psi_\text{specific}^{(K)}) \\
    & \quad + \log p(\psi_\text{shared}, \psi_\text{specific}^{(1)}, \dots, \psi_\text{specific}^{(K-1)} \mid \mathcal{D}_1, \dots, \mathcal{D}_{K-1}) + \log p(\psi_\text{specific}^{(K)}) \bigg)
\end{align}

However, for the sake of readability, we ignore the multi-head setting in the remainder of the section. 

As a way to analyze the stability of the MAP estimates, we perform a Laplace approximation of the posterior parameter distributions using the solutions obtained in Eq.~\ref{supp:eq:remarks:from:scratch:map} and Eq.~\ref{supp:eq:remarks:ewc:map} (cf. Sec. \ref{supp:methods:online:ewc} and \cite{huszar:ewc:note:2018}), yielding:

\begin{equation}
    p(\psi \mid \mathcal{D}_k) \approx \mathcal{N}\big(\psi_\text{MAP}^{(k)}, (\Omega^{(k)})^{-1} \big) \quad \text{and} \quad
    p(\psi \mid \mathcal{D}_1, \dots, \mathcal{D}_K) \approx \mathcal{N}\big(\psi_\text{MAP}^{(1:K)}, (\Omega^{(1:K)})^{-1} \big)
\end{equation}

where the precision matrices are given by:

\begin{equation}
    \Omega^{(k)} = \frac{1}{\sigma_\text{prior}^2} I + N_k F^{(k)}
\end{equation}

\begin{equation}
    \Omega^{(1:K)} = \frac{1}{\sigma_\text{prior}^2} I + \sum_{k=1}^K N_k F^{(k)}
\end{equation}

assuming an isotropic Gaussian prior $p(\psi) = \mathcal{N}(0, \sigma_\text{prior}^2 I)$. In the equations above, $N_k$ denotes the dataset size of task $k$ and $F^{(k)}$ the Fisher information matrix estimated on the model after learning task $k$.

Since the Gaussian posterior approximation contains viable solutions for the data to be explained, high entries in the covariance matrix indicate a flat solution around the MAP estimate. Analogously, the Fisher information matrix is connected to the expected Hessian of the log-likelihood. Therefore, low Fisher values correspond approximately to low curvature of the likelihood loss landscape and are thus indicative of flat minima, which are often considered desirable and have been linked to better generalization \citep{hochreiter_flat_1997, chaudhari_entropy-sgd_2019}. Incidentally, flat minima can also be considered desirable for the \texttt{HNET} approach. Because flatness indicates robustness of the found solutions against parameter perturbations, it loosens the pressure on the L2 regularization applied in Eq.~\ref{eq:hnet:reg}, and therefore decreases sensitivity to the regularization strength $\beta$.

\begin{figure}[t]
  \begin{center}
    \includegraphics[width=0.6\textwidth]{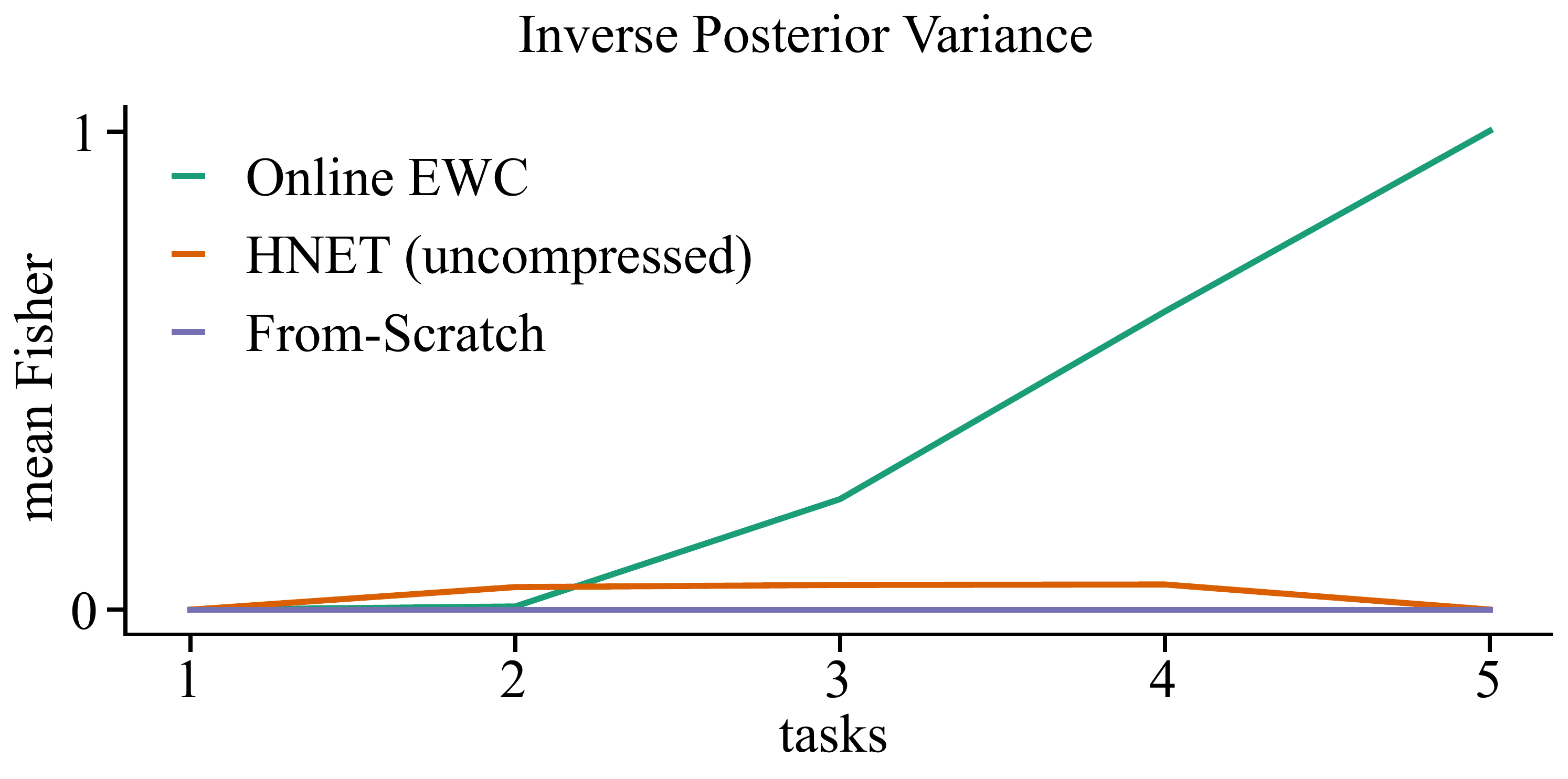}
  \end{center}
  \caption{Inverse posterior variance for some CL methods. The plot depicts the mean of the diagonal Fisher matrix as obtained from the \texttt{during} solutions found by different methods. For methods using task-specific solutions, such as \texttt{From-Scratch} and \texttt{HNET}, robustness only has to be measured with respect to the current task and therefore the Fisher has been evaluated on the corresponding task only. For \texttt{Online EWC}, the solution always has to be robust with respect to all tasks seen so far.}
  \vspace{-1.5em}
  \label{supp:fig:fisher:methods}
\end{figure}

To test these insights empirically, we studied the mean of the diagonal Fisher matrix obtained in the \texttt{during} models for the Permuted Copy task with $p=i=5$ (cf. Sec. \ref{sec:exp:copy}). The results are plotted in Fig. \ref{supp:fig:fisher:methods}. The \texttt{From-Scratch} baseline consists of independently trained models and therefore its solution only has to be stable with respect to the current task. Since the hypernetwork approach aspires to this behavior while allowing transfer, we treat it identically. Thus, for both methods we characterize the flatness of the solution using the mean of the corresponding task-specific diagonal Fisher matrix $F^{(k)}$. As expected, the \texttt{From-Scratch} baseline exhibits low Fisher values that are similar across tasks, illustrating the flatness of the separately found solutions. For the compressed hypernetworks used in this work, we do not see a clear trend. Instead, the mean Fisher values are highly variable and depend on the hyperparameter setting. Therefore, we instead considered an uncompressed version, where the hypernetwork is allowed to be five times as big as the corresponding main model (i.e., it consumes roughly the same amount of model parameters as the \texttt{From-Scratch} baseline). In this case, and as shown in Fig. \ref{supp:fig:fisher:methods}, the mean Fisher values are similar across tasks and the flatness of solutions is comparable to those found by \texttt{From-Scratch}. This observation is interesting as it opens up possible avenues for future work. As indicated above, it would be desirable to obtain flat, task-specific solutions with the \texttt{HNET} approach. Since this does not seem to occur automatically (at least in the compressed regime), one could enforce flat solutions explicitly. One could use, for instance, methods such as Entropy-SGD \citep{chaudhari_entropy-sgd_2019} that are applicable to the hypernetwork approach.

Lastly, the solution found by \texttt{Online EWC} has to be robust with respect to all previously seen tasks and therefore its curvature for task $k$ can be characterized via the sum over all tasks seen so far $\sum_{k' < k} N_{k'} F^{(k')}$. As expected, mean Fisher values are progressively increasing (Fig. \ref{supp:fig:fisher:methods}). This has the well known side effect of increasing rigidity and harming flexibility for learning new tasks. Furthermore, and connecting back to the literature on flat minima, it can be argued that the final solution found by \texttt{Online EWC} is less desirable as it resides in a sharper minimum of the loss landscape and empirical evidence suggests that such minima are harmful to generalization.


\subsection{Inferring task identity at test time}
\label{supp:remarks:task:id}

In this study, we only consider the case where task identity is known to the system during test time. A more challenging but arguably also more interesting CL scenario overcomes this constraint by inferring task identity based on the input sequence.\footnote{This kind of CL scenario was termed \texttt{CL3} in \citet{vandeVen2019Apr} and \citet{oswald:hypercl}.} However, this is only possible for task sets where the data input distributions are sufficiently dissimilar to allow discrimination. For instance, the Copy Task and its variants would not be applicable to this scenario, as all tasks share the same input data distribution. Thus, inferring the task identity from the input alone is impossible in such a case.\footnote{It is however always possible to design an auxiliary system that infers task identity from a given and appropriately chosen context \citep{oswald:hypercl, he:2019:task:agnostic:cl}}

One possible way to achieve this is by sequentially turning the CL problem into a multitask problem via replay. For classification problems, the multi-head output could be replaced by a growing softmax \citep{van_de_ven:replay:through:feedback} that is trained analogously as described in Sec. \ref{supp:methods:coresets} and \ref{supp:methods:generative:replay}.\footnote{Distillation targets have to be zero-padded as the softmax dimension is growing with each task.} However, this solution relies on successfully training generative models or on storing a sufficient amount of past data. It also successively turns the CL problem into a multitask problem leading to an undesirable increase of computational demands.

An alternative approach suggested in \citet{oswald:hypercl} relies on outlier detection via predictive uncertainty. For instance, in a multi-head setting, one could choose the output head with the lowest predictive uncertainty for classification, as the input sample can be considered "in-distribution" for this head. Even though proper out-of-distribution detection is a challenging and in itself still unresolved problem of machine learning \citep{snoek2019can:you:trust:your:models:uncertainty}, it would be an interesting direction for future work to investigate this approach for RNNs.

Another alternative, utilized in \citet{Cossu2020Apr, cossu2020continual} on sequential data, is the use of a different autoencoder per task. The autoencoder with the lowest reconstruction error for a given input sample will determine the task identity.\footnote{Note that regularized autoencoders have been shown to elicit properties of the data-generating density function \citep{bengio:denoising:aes}. Hence, this method of task inference can be loosely linked to proper out-of-distribution detection.} Such an approach also relies on the ability to successfully train generative models. In addition, the naive implementation requires one autoencoder per task. However, this last problem can be sidestepped using a hypernetwork-protected autoencoder (cf. method \texttt{HNET+R} in Sec. \ref{supp:methods:generative:replay:cl:with:svae}).


\subsection{The effect of modifying the experimental setup}
\label{supp:cl:experimental:setting}

To further highlight the importance of comparing CL methods within a clearly defined experimental setting, we perform a few controls where we vary this setting and study the impact on performance, focusing on \texttt{Online EWC}.

Despite an even distribution of hyperparameter-optimization resources among methods, a comparison might still be inconclusive if certain experimental variables are misaligned, such as the number of output heads, the number of model parameters or the availability of task identity information. Such factors can positively or negatively influence the performance of each method and therefore need to be carefully controlled for.

\paragraph{Using a single output head.} Throughout this study all methods are trained in a multi-head setting, where a different set of output weights is used for each task, consistent with the first CL scenario described by \citet{vandeVen2019Apr}. Another possibility is to use a single head, i.e. a common set of output weights across all tasks.
One can expect that a method producing task-conditional weights, such as \texttt{HNET}, might be better suited for a single-head setting than a method that has to progressively adapt its output weights, especially if the output is normalized (e.g., by using a softmax, since individual weight changes without co-adaptation of other weights in the output layer can drastically alter all predictions on prior tasks).

To empirically assess whether \texttt{HNET} and \texttt{Online EWC} are differently affected by the type of output layer used, we rerun the hyperparameter searches for the \textit{Permuted Copy Task} experiment $i=p=5$ from Sec. \ref{sec:exp:copy} using a single shared output head for all tasks. For fair comparison (cf. paragraph below), task identity in the form of a one-hot vector is provided as an additional input to the networks trained with \texttt{Online EWC}.

While the performance for \texttt{HNET} is only slightly affected ($95.78 \pm  2.13$ mean \texttt{final} accuracy), \texttt{Online EWC} drops to  $68.41 \pm 1.45$ \% mean \texttt{final} accuracy, which drastically differs from the multi-head result obtained in Sec. \ref{sec:exp:copy}. 
This comparison highlights the importance of having task-specific output weights in a CL setting with dissimilar tasks.

\paragraph{Weight-importance methods with task-conditional computation.} This study focuses on comparing methods in a CL scenario where task identity is known during inference (cf. SM \ref{supp:remarks:task:id}). Traditionally, weight-importance methods solely use task identity to select the correct output head in a multi-head setting. This implies that computation within the RNN is shared among all tasks up to the output layer, and that all tasks need to be solved in parallel if task identity cannot be inferred from the provided inputs.
To overcome this constraint and allow for task-conditional computation, task identity can be provided to the network as an additional input  (e.g., in the form of a one-hot encoding).

\vspace{3mm}
\begin{table*}[ht]
  \centering
  \caption{Mean \texttt{during} and \texttt{final} accuracies across several experiments for \texttt{Online EWC} either without providing the task identity as additional input (left column) or by explicitly enabling task-conditional computation by giving the task identity as additional input (right column). (Mean $\pm$ SEM in \%, $n=10$).}
  \begin{small}
      \begin{tabular}{lcc|cc}
        & \multicolumn{2}{c}{w/o task identity} & \multicolumn{2}{c}{with task identity} \\
        & \textbf{during} & \multicolumn{1}{c}{\textbf{final}} & \textbf{during} & \textbf{final} \\ \toprule
        \texttt{Permuted Copy} $p=i=5$ & 99.93 $\pm$  0.01 & 98.66 $\pm$  0.14 & 97.61 $\pm$  0.31 & 97.54 $\pm$  0.30 \\
        \texttt{Sequential Split-SMNIST} $m=1$ & 98.52 $\pm$ 0.11 & 97.05 $\pm$ 0.59 & 99.39 $\pm$  0.08 & 98.36 $\pm$  0.37  \\
        \texttt{Sequential Split-SMNIST} $m=2$ & 91.86 $\pm$ 1.13 & 76.65 $\pm$ 3.39 & 95.71 $\pm$  0.53 & 88.51 $\pm$  1.26 \\
        \texttt{Sequential Split-SMNIST} $m=3$ & 83.62 $\pm$ 1.10 & 75.26 $\pm$ 1.85 & 96.63 $\pm$  0.89 & 93.14 $\pm$  1.24  \\
        \texttt{Sequential Split-SMNIST} $m=4$ & 92.33 $\pm$ 2.49 & 73.16 $\pm$ 0.98 & 90.65 $\pm$  1.28 & 88.94 $\pm$  1.32 \\
        \texttt{Audioset} & 68.82 $\pm$  0.20 & 65.56 $\pm$  0.35 & 71.74 $\pm$  0.20 & 66.35 $\pm$  0.36  \\
        \texttt{PoS tagging} & 87.49 $\pm$  0.04 & 86.89 $\pm$  0.03 & 89.78 $\pm$  0.03 & 89.67 $\pm$  0.04 \\
      \end{tabular}
  \label{supp:tab:ewc:task:identity}
  \end{small}
\end{table*}

\vspace{2mm}
To evaluate whether a weight-importance method (namely \texttt{Online EWC}) can benefit from such task-conditioning, we rerun the hyperparameter-searches for some of our main experiments while providing the task-identity as an additional input. The results are depicted in Table \ref{supp:tab:ewc:task:identity}. We observe that in some experiments where the main network is an LSTM or BiLSTM, performance can be greatly improved by introducing task-conditioning. Performance gains are particularly striking in the Sequential Split-SMNIST experiment. This is somewhat surprising since in this dataset task identity can be inferred from the inputs alone (even though this might require observing a significant portion of the input sequence first), which should enable task-conditional processing without the need to explicitly provide the task identity.

When moving into more difficult CL scenarios, task-identity is not explicitly given to the system and has to be first inferred from the inputs to allow task-conditional processing. SM \ref{supp:remarks:task:id} details several approaches for how an auxiliary system can provide a task-identity signal. However, a more natural approach for adapting task-conditional EWC to these CL scenarios would be to use predictive uncertainty (which was suggested by \cite{oswald:hypercl} in the context of CL with hypernetworks). Therefore, one could use the approximate parameter posterior distribution, that has been obtained during training, also during inference to quantify uncertainty of output heads when considering all possible task-identity inputs. Note, however, that this increases computational efforts significantly as (1) an MC estimate of the posterior predictive distribution needs to be obtained and (2) this process needs to be repeated for every task identity.

In conclusion, this unorthodox but justified use of \texttt{Online EWC} reveals great potential of weight-importance methods for sequential processing.
Because of its performance, ease of implementation and elegant mathematical derivation, \texttt{Online EWC} might be preferable in simple use cases over more elaborate methods such as \texttt{HNET}, even though \texttt{HNET} remains the better performing option in most situations. Therefore, an important contribution of our work is to identify \texttt{Online EWC} as a strong contender for sequential processing, despite having been questioned by existing work \citep{asghar2018progressive, Cossu2020Apr, Li2020Compositional:Language:CL} and despite the shortcomings revealed by our own study.

\end{document}